\documentclass[10pt,twocolumn,letterpaper]{article}

\usepackage{iccv}
\usepackage{times}
\usepackage{epsfig}
\usepackage{graphicx}
\usepackage{amsmath}
\usepackage{amssymb}

\usepackage[accsupp]{axessibility}

\usepackage{amsthm}
\newtheorem{thm}{Theorem}
\newtheorem{prop}[thm]{Proposition}
\newtheorem{cor}[thm]{Corollary}

\theoremstyle{definition}
\newtheorem*{remk*}{Remark}
\usepackage{pifont}
\newcommand{\cmark}{\ding{51}}%
\usepackage{multirow}
\usepackage{boldline,makecell}
\usepackage{booktabs}
\usepackage{xcolor,colortbl}
\definecolor{finesky}{HTML}{E6F5F0}
\definecolor{lightgray}{HTML}{ECECEC}
\definecolor{lightcyan}{rgb}{0.88,1,1}
\definecolor{weakblue}{HTML}{0071bc}
\DeclareMathOperator{\sign}{sign}
\DeclareMathOperator{\diag}{diag}
\usepackage{subcaption}
\newcommand\blfootnote[1]{%
\begingroup
\renewcommand\thefootnote{}\footnote{#1}%
\addtocounter{footnote}{-1}%
\endgroup
}

\usepackage[pagebackref=true,breaklinks=true,letterpaper=true,colorlinks,bookmarks=false]{hyperref}

\iccvfinalcopy % *** Uncomment this line for the final submission

\def\iccvPaperID{6034} % *** Enter the ICCV Paper ID here

\ificcvfinal\fi

\begin{document}

%%%%%%%%% TITLE
\title{Understanding the Feature Norm for Out-of-Distribution Detection}

%% omit the following lines up until the closing ``}''.
%% Additional authors and addresses can be added with ``\and'',
%% just like the second author.
%% To save space, use either the email address or home page, not both
%\and

\author{
Jaewoo Park$^{1,2}$ \quad Jacky Chen Long Chai$^{1}$ \quad Jaeho Yoon$^{1}$ \quad Andrew Beng Jin Teoh$^{1\dag}$ \\
$^1$Yonsei University \quad $^2$AiV Co. 
%\\
%{\tt\small \{julypraise, bjteoh\}@yonsei.ac.kr, jung.yoo@northeastern.edu}
}

\maketitle
\ificcvfinal\thispagestyle{empty}\fi

%%%%%%%%% ABSTRACT
\begin{abstract}
A neural network trained on a classification dataset often exhibits a higher vector norm of hidden layer features for in-distribution (ID) samples, while producing relatively lower norm values on unseen instances from out-of-distribution (OOD). Despite this intriguing phenomenon being utilized in many applications, the underlying cause has not been thoroughly investigated. In this study, we demystify this very phenomenon by scrutinizing the discriminative structures concealed in the intermediate layers of a neural network. Our analysis leads to the following discoveries: (1) The feature norm is a confidence value of a classifier hidden in the network layer, specifically its maximum logit. Hence, the feature norm distinguishes OOD from ID in the same manner that a classifier confidence does. (2) The feature norm is class-agnostic, thus it can detect OOD samples across diverse discriminative models. (3) The conventional feature norm fails to capture the deactivation tendency of hidden layer neurons, which may lead to misidentification of ID samples as OOD instances. To resolve this drawback, we propose a novel negative-aware norm (NAN) that can capture both the activation and deactivation tendencies of hidden layer neurons. We conduct extensive experiments on NAN, demonstrating its efficacy  and compatibility with existing OOD detectors, as well as its capability in label-free environments.
\end{abstract}
%%%%%%%%% BODY TEXT

\blfootnote{\noindent $^{\dag}$ Corresponding author: Andrew Beng Jin Teoh}

\section{Introduction}
Deep learning-based models are increasingly used for safety-critical applications such as autonomous driving and medical diagnosis. Despite the effectiveness of deep models in closed-set environments where all test queries are sampled from the same distribution of train data, the deep models are reported fairly vulnerable \cite{nalisnick2018deep,hendrycks2016baseline} to  outliers from out-of-distribution \cite{hendrycks2021natural,yang2022openood} and  make highly confident but invalid predictions thereon \cite{nguyen2015deep}. As it is critical to prevent such malfunction in deploying deep models for open environment applications, the out-of-distribution (OOD) detection problem has attracted massive attention in recent years \cite{yang2021generalized}.

Despite the importance of this field, only a handful of works have been devoted to understanding
how the deep network becomes aware of OOD \cite{fang2022out,fang2021learning,dietterich2022familiarity,ming2022impact,morteza2022provable}. 
One particular under-studied signal in OOD detection is \textit{the norm of feature vectors} residing in the hidden layers of neural networks. 
Its known behavior is that a model trained on the ID data exhibits larger values of feature norm over ID samples than the OOD instances \cite{dhamija2018reducing,yu2020out,chen2020norm,meng2021magface}. However, the studies are mainly empirical and provide no underlying principle of the feature norm at a fundamental level.

\begin{figure}[!t]
\begin{center}
\includegraphics[width=.9\linewidth]{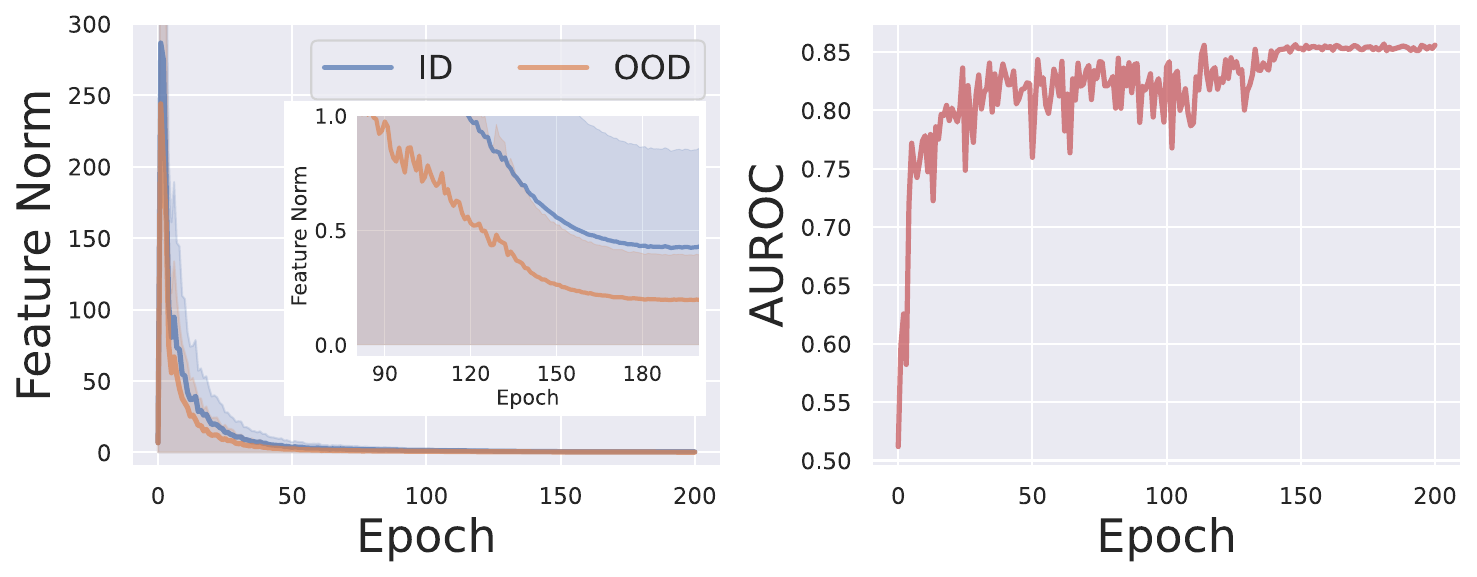}
\end{center}
\caption{
(left) As a discriminative model is trained, its hidden layer features exhibit higher vector norm on in-distribution samples (ID) and relatively lower norm on out-of-distribution (OOD) instances.
This phenomenon prevails even when the model reduces the overall feature norm (\eg by weight decay).
(right) As a result, ID samples are separated from OOD instances with respect to the feature norm.
To see its underlying cause, 
we analyze the discriminative structures concealed in the hidden layer.
}
\label{fig:preliminary}
\end{figure}

A preliminary attempt at understanding the feature norm has been given in the appendix of \cite{vaze2021open}. The authors of \cite{vaze2021open} argue that minimizing the cross entropy (CE) maximizes the feature norm of ID samples. However, the argument is not general. As we observe in Fig.~\ref{fig:preliminary}, training the weight-decayed model decreases the overall feature norm, but the separation between ID and OOD remains obvious. Hence, we require a new lens to understand the underlying cause of feature norm separation.

In this work, we study \textit{why} the feature norm separates ID from OOD. To this end, we both theoretically and empirically show that the feature norm is equal to a confidence value of a classifier hidden in the corresponding layer. Based on the existing theory on the classifier confidence \cite{fang2021learning}, the equality guarantees the detection capability of feature norm.

Furthermore, our analysis indicates that the feature norm is agnostic to the class label space. This suggests that the feature norm can detect OOD using any general discriminative model, including self-supervised classifiers.
We validate this postulation empirically under several aspects: Firstly, by considering inter- and intra-class learning independently, we show that inter-class learning enables the feature norm to separate OOD from the training fold of ID. The intra-class learning, on the other hand, generalizes the detection capability to the test environment, enabling the feature norm to differentiate OOD from the test fold of ID. The finding shows that inter- and intra-class learning corresponds to memorization and generalization, respectively, in the context of OOD detection.
Secondly, we show that the detection capability of feature norm is strongly correlated to the entropy of activation (\ie diversity of on/off status of neurons). As activation entropy is a class-agnostic characteristic, the finding reinforces our postulation.  

In addition to that, we observe that the conventional vector norm only captures the activation tendency of hidden layer neurons, but misses the deactivation counterpart. 
Failing to account for the deactivation tendencies results in the loss of important characteristics specific to ID samples, potentially leading to misidentification of such instances as OOD examples.
Motivated by this drawback, we derive a novel negative-aware norm that captures both the activation and deactivation tendencies of hidden layer neurons.

We perform a thorough assessment of the NAN and demonstrate its efficacy across OOD benchmarks. Additionally, we confirm that NAN is compatible with several state-of-the-art OOD detectors. Furthermore, NAN is free of hyperparameters, requires no classification layer, and does not necessitate expensive feature extraction from a bank set. Consequently, NAN can be readily deployed in scenarios where class labels are unavailable. We evaluate NAN in unsupervised environments using self-supervised models and assess its performance on one-class classification benchmarks.

The contributions of our work are summarized as follows:
\begin{itemize}
\item We demystify the OOD detection capability of the feature norm by showing that the feature norm is a confidence value of a classifier hidden in the corresponding layer (Sec.~\ref{sec:theory}). 

\item
We reveal that the feature norm is class-agnostic, hence able to detect OOD using general discriminative models (Sec.~\ref{sec:emp_analysis}). We validate this property under several aspects including inter/intra-class learning and activation entropy.

\item
We put forward a novel negative-aware norm (NAN), which captures both activation and deactivation tendencies of hidden layer neurons (Sec.~\ref{sec:method}). NAN is hyperparameter-free, label-free, and bank-set-free. NAN can be easily integrated with state-of-the-art OOD detectors. (Sec.~\ref{sec:exp})
\end{itemize}

\section{Background}
\label{sec:background}
% Let $\mathbf{x}$ be an input sample to be determined as either an ID or OOD sample. An OOD detection score function $S(\mathbf{x})$ can be used to determine $\mathbf{x}$ as OOD if $S(\mathbf{x}) > \tau $ for some threshold $\tau$ and as ID otherwise.
The goal of OOD detection is to devise a score function $S(\mathbf{x})$ that determines an input sample $\mathbf{x}$ as OOD if $S(\mathbf{x}) {<} \tau $ for some threshold $\tau$ and as ID otherwise.
There are several ways to derive such a score function from a discriminative model $p_\theta (y|\mathbf{x})$. A standard detection score is the maximum softmax probability (MSP) score \cite{hendrycks2016baseline}, which is defined as $S(\mathbf{x}) {=} \max_y p_\theta (y|\mathbf{x})$ with $p_\theta$ modeled by the softmax function. 
%\cite{rna} observed the magnitude ($l_2$ norm) of embedding vector $g(\mathbf{x})$ can also serve as a detection score based on the empirical observation that the norm of embedding tends to be larger over ID than OOD.

%The discriminative probability is formulated by $p_\theta (y|\mathbf{x}) {=} \softmax(\psi(\mathbf{x}))$ where $\psi(\mathbf{x}) {\in} \mathbb{R}^K$ is the classifier logit vector. \cite{maxlogit} showed the effectiveness of the direct usage of the maximum logit $S(\mathbf{x}) {=} {-} \max_k \psi_k(\mathbf{x})$ (\textbf{MaxLogit}) for OSR tasks. On the other hand, 

%MSP and MaxLogit as OOD detectors follow natural probabilistic interpretation since ideally calibrated posterior and its corresponding maximum logit should be large over ID exclusively. However, the representation magnitude of the OOD score is not firmly grounded, and its principal mechanism is unclear.
%In this work, we aim to fill in this gap, explaining how the OOD detection is enabled by the norm of the ReLU activation vector.

\noindent
\textbf{Other OOD detection scores} include the energy score \cite{liu2020energy} that extracts the energy function \cite{grathwohl2019your} from the classification layer. \cite{hendrycks2019scaling} proposes the KL divergence to the uniform prediction, while \cite{vaze2021open} applies only the maximum value of logit.

Other works propose the utilization of distance metrics for OOD detection.
\cite{lee2018simple} applied the Mahalanobis distance as an OOD detector based on a strong parametric assumption that each ID class follows a Gaussian distribution with a shared covariance.
A unified approach SSD \cite{sehwag2021ssd} generalizes the principle of \cite{lee2018simple}, exploiting  class clusters attained by unsupervised $K$-means. As SSD requires no class labels, its usage is general and applicable to both supervised and unsupervised models. ViM \cite{wang2022vim} adopts SSD but uses the orthogonal distance from principal components instead, and combines it with the energy score with manual calibration. 
CSI \cite{tack2020csi}, on the other hand, defines the detection score by combining a rotation classifier with the $k$-nearest neighbor distance. The effectiveness of CSI, however, comes from a deliberate design of image-specific data augmentations. As a simpler and model-agnostic approach, \cite{sun2022out} proposed the $k$-nearest neighbor (KNN) distance for OOD detection. Despite its broad applicability \cite{ming2022exploit}, KNN requires a careful hyperparameter search on the sampling ratio of the ID bank set and the number of neighbors.

Apart from the distance-based OOD detectors, an alternative approach to detecting OOD is by perturbing the signal of the network.
\cite{liang2017enhancing,hsu2020generalized} observed a particular input perturbation perturbs OOD samples severely but makes ID samples remain mostly invariant. \cite{sun2021react} proposed a rectification layer that clips out all values greater than a given threshold.
Despite their effectiveness, the perturbation methods rely on specific assumptions of network signal distributions and are sensitive to hyperparameters. 
% Moreover, the detection improvement comes at the expense of ID classification accuracy.

\noindent
\textbf{On feature norm.}
The first application of feature norm for OOD detection was reported by \cite{dhamija2018reducing}, whose authors observed that the magnitude ($l_2$-norm) of embedding vector tends to be larger for ID than OOD. The same trend was observed in the appendixes of \cite{tack2020csi,vaze2021open,huang2021importance} for generic images. 
In biometrics,
\cite{yu2020out} observed the same phenomenon for face images, thereby devising a score that can more effectively reject unseen identities based on the feature norm. \cite{meng2021magface} extended the application of feature norm, showing that it can measure the quality score of the face image. 
On the other hand, 
\cite{chen2020norm,chen2021norm} observed that the norm of feature embedding effectively differentiates a person from his/her surrounding background, and thus can be used to improve the performance and efficiency of person search.
Besides OOD detection, \cite{yuan2017feature} observed that the embedding vectors of highly discriminative samples lie in the area of the large norm. \cite{xu2019larger} extended this observation, demonstrating the samples with large feature norms are not only more discriminative but also more transferable for domain adaptation.

Although numerous works report empirical observations of the phenomenon, to our best knowledge, no work provides a systematic theoretical explanation of the underlying mechanism of feature norm.

\begin{figure}[t]
\begin{center}
\includegraphics[width=.9\linewidth]{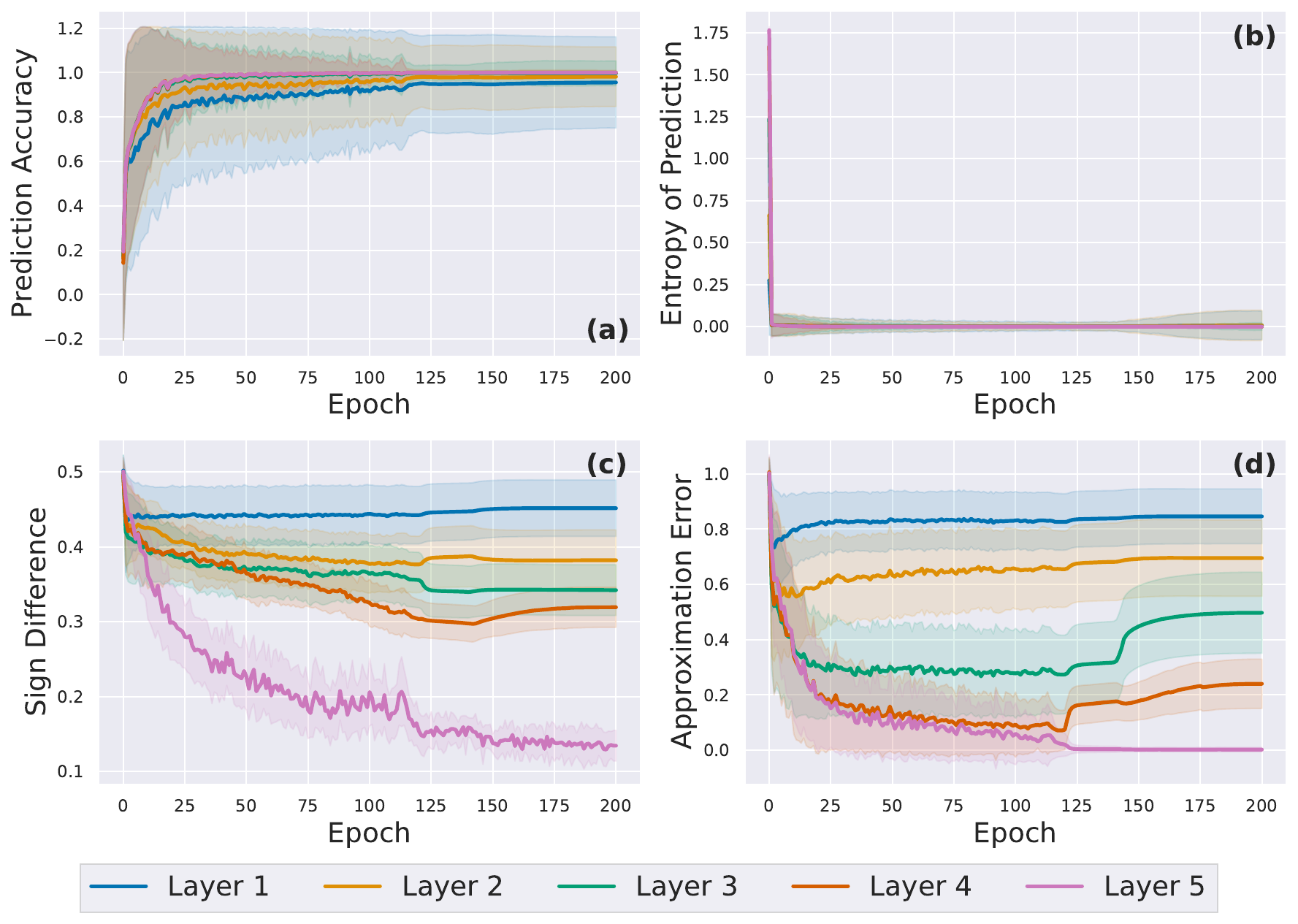}
\end{center}
\caption{
\textbf{The results on hidden classifiers} of MLP-5 trained on CIFAR-10 (ID). (a) The prediction accuracy of the hidden classifier increases through learning. (b) Accordingly, the prediction becomes more deterministic (\ie, confident). (c,d) As the sign difference between the feature vector and class weight $\mathbf{c}^{(l)}_y$ is reduced, the approximation error between the feature norm and the maximum value of the hidden classifier is reduced in a similar trend, \textit{verifying} our Thm.~\ref{thm:approx}. Results with other activation functions are in Sec.~\ref{asec:add_hidden_classifier_mlp}.
}
\label{fig:hidden_classifier}
\end{figure}

\section{Understanding Feature Norm as a Confidence of Hidden Classifier}
\label{sec:theory}

In this section, we show that the feature norm is a confidence value of a discriminative classifier covertly concealed in the corresponding layer. Specifically, under a regularity condition, the $l_1$-norm of the feature vector is equal to the maximum logit of a hidden classifier attained by binarizing the network weights. Hence, based on the theory from \cite{fang2022out}, the feature norm is guaranteed its detection proficiency.

\subsection{Theoretical analysis}
\label{sec:theory_analysis}
%To justify the efficacy of the feature norm for OOD detection, we show that the feature norm is a confidence value of a hidden classifier. The hidden classifier is attained by binarization of network weights.

\noindent
\textbf{Notation and setup.}
Let $\{(\mathbf{x}_i, y_i)\}_{i=1}^N$ be the train ID dataset where $y_i {\in} \mathcal{Y} {=} \{1,\dots,K\}$ are labels from $K$ classes. 
Suppose our model is a multi-layer perceptron (MLP) whose $l$-th hidden layer consists of the $d_l$-dimensional feature vector $\textbf{a}^{(l)}$ computed by $\mathbf{a}^{(l)} {=} \sigma( \mathbf{W}^{(l)T} \mathbf{a}^{(l-1)} )$ consecutively from the initial layer $l{=}0$ to the \textit{last hidden layer} $l{=}L$, where $\mathbf{a}^{(0)} {=} \mathbf{x}$. The vector of pre-activated units is denoted by $\mathbf{z}^{(l)}$, which satisfies $\mathbf{a}^{(l)} {=} \sigma(\mathbf{z}^{(l)})$. The activation function $\sigma$ is assumed to be a unit-wise rectifier such as ReLU \cite{nair2010rectified,fukushima1975cognitron}, GeLU \cite{hendrycks2016gaussian}, and Leaky ReLU \cite{xu2015empirical}. Each weight matrix $\mathbf{W}^{(l)} {\in} \mathbb{R}^{d_{l-1} \times d_{l}}$ constitutes trainable parameters $\theta$. The classifier logit $\psi(\mathbf{x}) {\in} \mathbb{R}^K$ is computed by $\psi(\mathbf{x}) {=} \mathbf{W}^{(L+1)T} \mathbf{a}^{(L)}$.

\noindent
\textbf{Assumption.}
We assume arbitrary class type for the label space $\mathcal{Y}$; classes can be supervised labels, instance classes, or even noisy labels.

To extract a hidden classifier from each hidden layer of the model, we first access the hidden layer through matrix multiplication.
\begin{prop}
\label{thm:decom}
The final logit is represented by
\begin{equation}
\label{eq:matrix_form}
\psi(\mathbf{x}) = \mathbf{C}^{(l)} \mathbf{a}^{(l)}
\end{equation}
for each hidden layer $l$, where 
\begin{equation}
\mathbf{C}^{(l)} {=} \left(\prod_{k=0}^{L-l-1} \mathbf{W}^{(L+1-k)T} \mathbf{D}^{(L-k)} \right) \mathbf{W}^{(l+1)T}
\end{equation}
with $\mathbf{D}^{(l)} {=} \diag(\frac{\sigma(z_1)}{z_1}, \dots, \frac{\sigma(z_{d_l})}{z_{d_l}})$ and the convention $\frac{\cdot}{0} = 0$. The matrix $\mathbf{C}^{(l)} = \mathbf{C}^{(l)}(\mathbf{x}) \in \mathbb{R}^{K \times d_l}$ depends on $\mathbf{x}$.
\end{prop}

\begin{proof}
All proofs are given in Sec.~\ref{asec:theory}.
\end{proof}

The multiplication by the coefficient matrix $\mathbf{C}^{(l)} = [\mathbf{c}^{(l)}_1, \dots, \mathbf{c}^{(l)}_K]^T$ resembles a classification layer with the column weight $\mathbf{c}^{(l)}_k {=} \mathbf{c}^{(l)}_k(\mathbf{x})$ as the $k$-th class proxy.

We note that $\psi$ is called a \textit{discriminative classifier} since the target class unit of logit is maximum $\psi_y > \psi_k$ for all $k {\neq} y$. If the output classifier $\psi$ is sufficiently discriminative, then binarizing the coefficient matrix $\mathbf{C}^{(l)}$ does not alter the prediction of the classifier. 
This leads us to a \textit{hidden classifier} $\overline{\psi}^{(l)} \in \mathbb{R}^K$ defined by binarizing the network weights:
\begin{equation}
\label{eq:hidden_classifier}
\overline{\psi}^{(l)} (\mathbf{x}) 
:= \mathbf{B}^{(l)} \mathbf{a}^{(l)}
:= \sign ( \mathbf{C}^{(l)} ) \mathbf{a}^{(l)}
\end{equation}
where  $\sign(x) =1$ if $x>0$ and $-1$ otherwise. 

\begin{prop}
\label{thm:hidden}
For all labeled sample $(\mathbf{x},y)$, suppose the discriminative learning of $\psi_k(\mathbf{x}) = \mathbf{c}^{(l)}_k \cdot \mathbf{a}^{(l)}$ increases and decreases the cosine similarities between $\mathbf{c}^{(l)}_k$ and $\mathbf{a}^{(l)}$ sufficiently for $k {=} y$ and $k {\neq} y$, respectively. Then $\overline{\psi}^{(l)}$ is a discriminative classifier with $\overline{\psi}^{(l)}_y > \overline{\psi}^{(l)}_k$ for all $k \neq y$.
\end{prop}

In the sufficient condition of Prop.~\ref{thm:hidden}, the network aligns the activation pattern $\sign(\mathbf{a}^{(l)})$ \cite{hanin2019deep} with the binary weight $\mathbf{b}^{(l)}_y$ that corresponds to the target class $y$. Here, $\mathbf{b}^{(l)}_y$ is the $y$-th row of $\mathbf{B}^{(l)}$. Due to the alignment, the feature norm becomes the prediction confidence $\max_k \overline{\psi}^{(l)}_{k} (\mathbf{x})$ of the hidden classifier.

\begin{thm}
\label{thm:approx}
Given the sufficient condition of Proposition \ref{thm:hidden}, the feature norm
\begin{equation}
\lVert \mathbf{a}^{(l)} \rVert_1 
\text{ converges to }
\overline{\psi}^{(l)}_{y} (\mathbf{x}) = \max_k \overline{\psi}^{(l)}_{k} (\mathbf{x}),
\end{equation}
in which case $\sign(\mathbf{a}^{(l)}) = \mathbf{b}^{(l)}_y$.
In general, for any $k$
\begin{equation}
0 \leq \lVert \mathbf{a}^{(l)} \rVert_1 - \overline{\psi}_k(\mathbf{x}) 
\leq 
\lVert \mathbf{a}^{(l)} \rVert_\infty
\lVert \sign(\mathbf{a}^{(l)}) - \mathbf{b}^{(l)}_{k} \rVert_1
%\sum_i 
%\lvert \sign(a^{(l)}_i) - b^{(l)}_{ki} \rvert.
\end{equation}
\end{thm}

Existing OOD theories on classifiers \cite{fang2022out,fang2021learning} assure that OOD samples have smaller prediction confidence than ID under regularity conditions. In this case, the feature norm of OOD also has a smaller value due to Thm.~\ref{thm:approx}:

\begin{cor}
\label{thm:separation}
If $\max_k \overline{\psi}^{(l)}_k(\mathbf{x}_{ood})$ is sufficiently small, then $\lVert \mathbf{a}^{(l)}(\mathbf{x}_{ood}) \rVert_1 < \lVert \mathbf{a}^{(l)}(\mathbf{x}_{ind}) \rVert_1$ for all ID samples $\mathbf{x}_{ind}$.
\end{cor}

\subsection{Empirical verification}
\label{sec:theory_emp_proof}

We empirically verify the above claims. We train a 5-layer MLP on CIFAR10 (ID) \cite{krizhevsky2009learning}. The full empirical setup is given in Sec.~\ref{asec:add_hidden_classifier}. Fig.~\ref{fig:hidden_classifier} shows that the hidden classifiers learn to increase their prediction accuracy while reducing the prediction uncertainty (entropy), verifying Prop.~\ref{thm:hidden}. As described in Thm.~\ref{thm:approx}, the discriminative training induces the sign alignment between the hidden layer feature and corresponding class weight $\mathbf{c}^{(l)}_y$, thereby reducing the gap between the feature norm and the maximum confidence of the hidden classifier.

\paragraph{Remark}
We remark that the trend of approximation error may not be precisely aligned with that of the sign difference (Fig.~\ref{fig:hidden_classifier}) as the sign difference is the sufficient condition but not a necessary one. Hence, when the sign difference is large, the approximation error can be either large or small; \ie they can be misaligned. However, due to its sufficiency, if the sign difference converges to 0, then the approximation error also decreases to 0.

\begin{figure}[t]
\centering
\includegraphics[width=.9\linewidth]{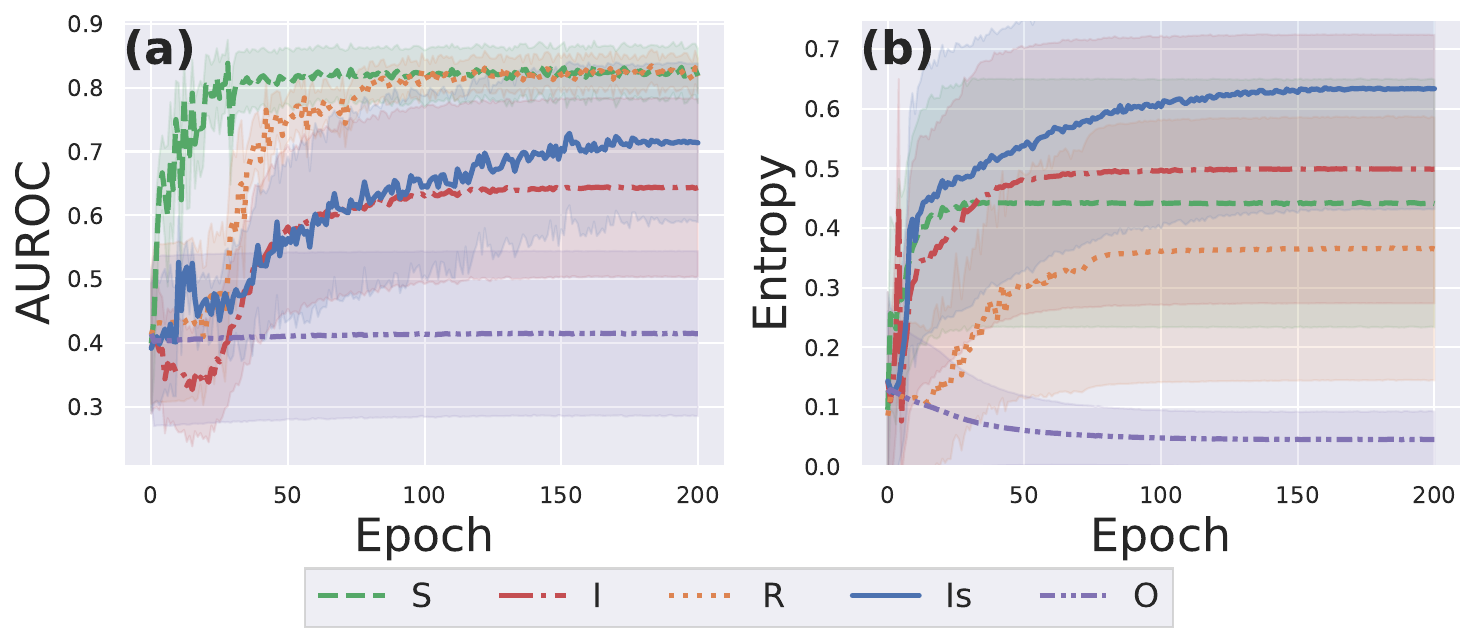}
\caption{
The results on ResNet-18 trained on CIFAR-10 (ID).
(a) Training the model increases the OOD detection performance of feature norm if and only if the model is discriminative.
(b) Accordingly, training the model increases the entropy of activation if and only if the model is discriminative.
Here, the models with S, I, R, and Is labeling schemes are discriminative, while model O is not discriminative.
}
\label{fig:entropy}
\end{figure}

\begin{figure}[t]
\begin{center}
\includegraphics[width=.7\linewidth]{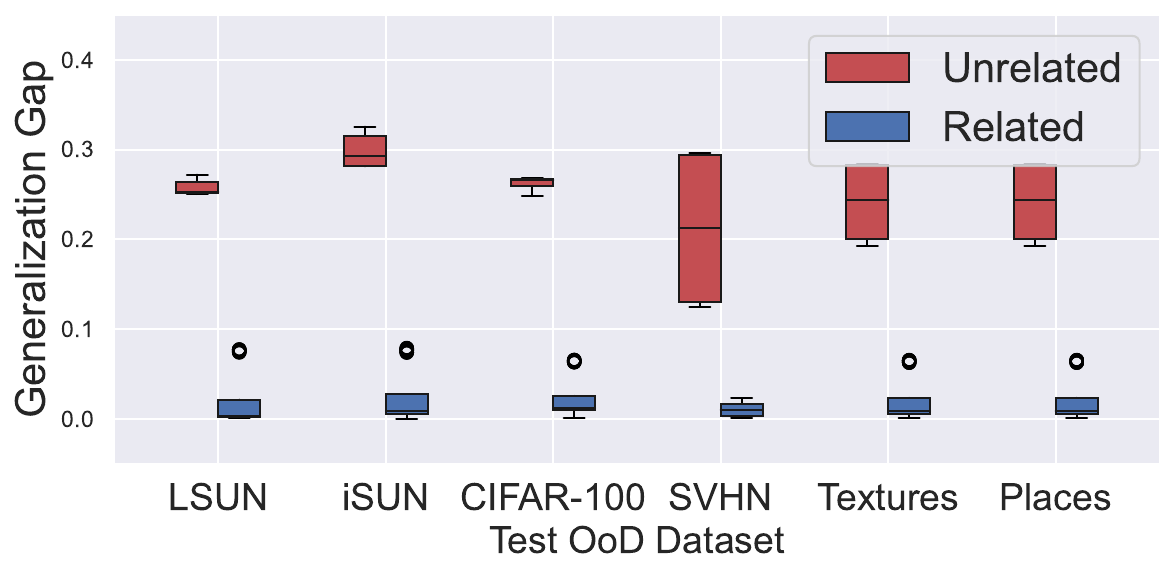}
\end{center}
\caption{
When the intra-class samples are \textit{related semantically} (\ie \{S,Is,O\}), the OOD detection performance is generalized to test environments (\ie small generalization gap). However, if intra-class samples are randomly related (R), or there is no more than one sample in each class (I), no generalization is observed.
}
\label{fig:detection_gap}
\end{figure}

\section{Class Agnosticity of Feature Norm}
\label{sec:emp_analysis}

The theoretical properties of feature norm proven in Sec.~\ref{sec:theory} hold true with respect to any type of label space, suggesting that the feature norm is class-agnostic and capable of detecting out-of-distribution (OOD) samples with any discriminative model. In this section, we conduct empirical analyses to validate this hypothesis across different aspects. Specifically, we observe that inter/intra-class learning generally enhances the feature norm's performance. We then demonstrate that the feature norm's performance is correlated with the entropy of activation, which is another class-agnostic characteristic of neural networks. The feature norm's dependence on class-agnostic factors provides further evidence supporting our hypothesis.

\subsection{Impact of inter/intra-class learning}
\label{sec:emp_analysis_interintra}

\noindent
\textbf{Setup.} 
We train a ResNet-18 on CIFAR-10, and test against different OODs, \ie, LSUN \cite{yu2015lsun}, iSUN \cite{xu2015turkergaze}, CIFAR-100 \cite{krizhevsky2009learning}, SVHN \cite{netzer2011reading}, Texture \cite{cimpoi2014describing}, and Places \cite{zhou2017places}. 

We consider five different training schemes by varying the label space. \textbf{`S'}: the supervised learning with generic object categories. \textbf{`I'}: the instance-discrimination learning with $y_i {=} i$. \textbf{`Is'}: instance-discrimination with data augmentation (\ie conventional self-supervision). \textbf{`R'}: learning with random binary labels. \textbf{`O'}: non-discriminative learning with every ID sample labeled by the same label `1'. 

The detection score we use is the feature norm $\lVert \mathbf{a}^{(L)} \rVert_1$ of the last hidden layer feature $\mathbf{a}^{(L)}$. The performance is measured by the area under receiving operating characteristic curve (AUROC). A more detailed description of the setup and full experimental results are given in Sec.~\ref{asec:setup_general_model}.

\noindent
\textbf{Inter-class learning.}
To analyze the effect of inter-class learning, we divide the training schemes into two: discriminative learning \{S,R,I,Is\}, and non-discriminative learning \{O\}. Fig.~\ref{fig:entropy} demonstrates that the feature norm separates OOD from the train fold of ID if and only if the model is trained with inter-class learning. In particular, the feature can detect OOD even if the model is trained with random noisy labels, indicating that its detection capability is independent of the class type of label space.

\noindent
\textbf{Intra-class learning.}
To examine the impact of intra-class learning, we divide the training schemes into two groups \{S,Is,O\} and \{R,I\}. In the former group \{S,Is,O\}, the intra-class samples are semantically related. On the latter group \{R,I\}, there is no semantic relation within the intra-class samples. Fig.~\ref{fig:detection_gap} indicates the generalization gap between train and test performances for OOD detection. The results support that the detection capability of feature norm is generalized to the test environment if and only if the intra-class samples are semantically related.

\noindent
\textbf{Summary on inter/intra-class learning.}
The detection capability of feature norms does not depend on a particular type of class label. Instead, any type of inter-class learning allows the feature norm to differentiate  OOD from the training fold of ID. On the other hand, intra-class learning with any appropriate semantics facilitates the separation of OOD from the test fold of ID. In general, inter-class learning corresponds to memorization, while intra-class is associated with generalization.

\begin{figure}[t]
\begin{center}
\includegraphics[width=.75\linewidth]{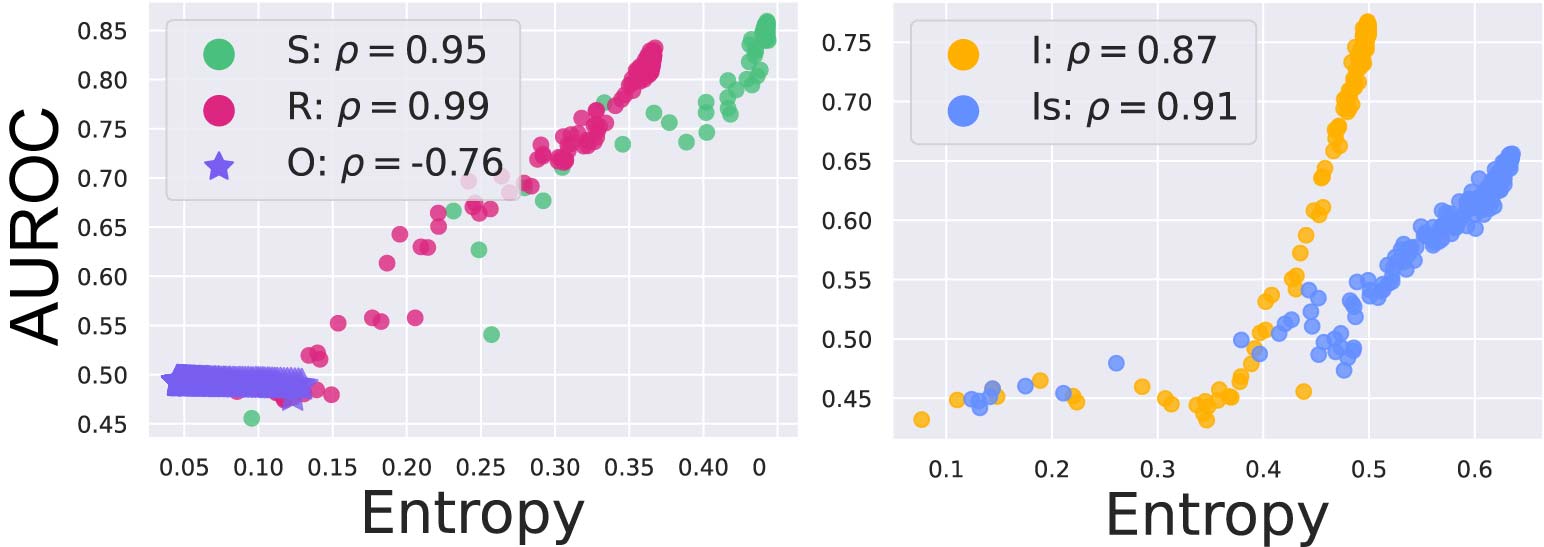}
\end{center}
\caption{
For discriminative models \{S,R,I,Is\}, the OOD detection performance of feature norm is positively \textit{correlated} to the averaged entropy of activation (Eq.~\eqref{eq:entropy}). However, no consistent correlation is found in the non-discriminative model O.
}
\label{fig:detection_relation}
\end{figure}

\subsection{The relation to the entropy of activation}
The feature norm's detection capability depends on the model's discriminative nature, not the class type. Here, we further show that the capability relies on the entropy of activation, which is another class-agnostic characteristic.

If the model is discriminative, target logits $\overline{\psi}^{(L)}_y(\mathbf{x})$ with different $y$
is maximized for ID samples $x$. Then, due to
\begin{equation}
\overline{\psi}^{(L)}_y(\mathbf{x}) 
%= \lVert \mathbf{a}^{(L)} \rVert_1
= \sum_i b_{y,i}^{(L)} a^{(L)}_i
\end{equation}
with $\mathbf{b}^{(L)}_y = (b^{(L)}_{y,1},\dots,b^{(L)}_{y,d_L}) \in \mathbb{R}^{d_L}$,
the unit $a^{(L)}_i$ is maximized for samples $\mathbf{x}$ in
$\{\mathbf{x} : b^{(L)}_{y,i}=1\}$, and minimized for $\mathbf{x}$'s in $\{\mathbf{x} : b^{(L)}_{y,i}=-1\}$.
Consequently, the \textit{entropy of activation} is maximized
\begin{equation}
\label{eq:entropy}
H(a^{(L)}_i) = {-}\sum_{c=0}^1 \mathbb{P}( 1_{a^{(l)}_i {>} 0} {=} c) \log \mathbb{P}( 1_{a^{(l)}_i {>} 0} {=} c)
\end{equation}
for each neuron $a^{(L)}_i$ of ID samples.

Conversely, if the model is not discriminative, \ie, $\mathcal{Y} = \{1\}$, then all ID samples likely have the same constant binary indicator; $b_{y,i}^{(L)}=c$ for all samples $\mathbf{x}$ where $c \in \{-1,1\}$. Hence, the activation entropy is minimized in this case.

This trend is empirically validated in Fig.~\ref{fig:entropy}b; only discriminative models maximize the activation entropy. Moreover, demonstrated by the strong correlation depicted in Fig.~\ref{fig:detection_relation}, the detection performance of the feature norm depends on the activation entropy, which is a characteristic independent of the class type of the label space. 

\section{Method: Negative-Aware Norm (NAN)}
\label{sec:method}

\begin{figure}[t]
\centering
\includegraphics[width=.995\linewidth]{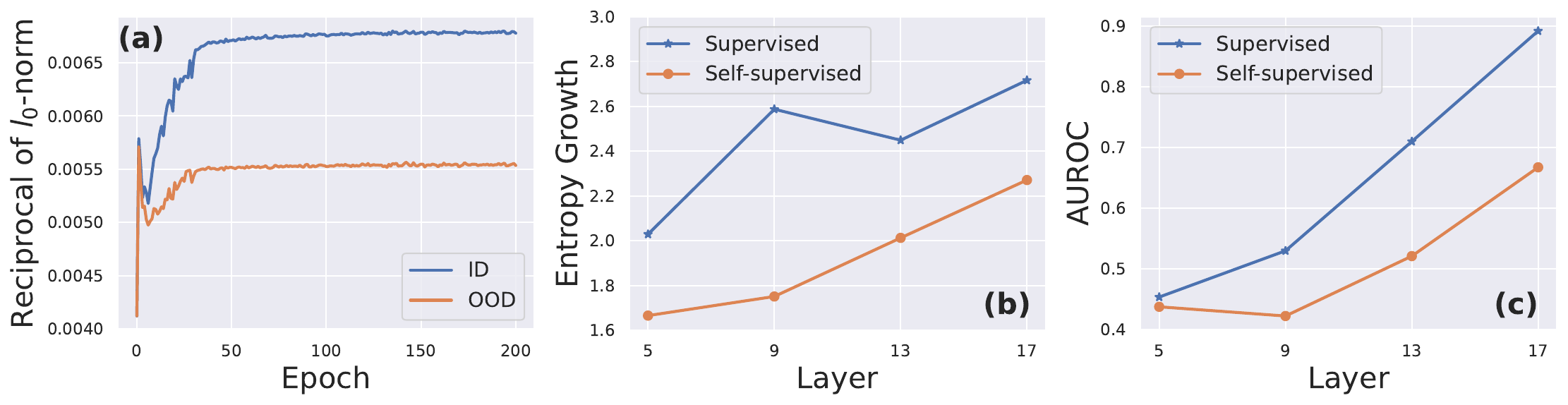}
\caption{
(a) The sparsity of activations, measured by $\lVert \mathbf{a}^{(L)} \rVert_0^{-1}$, is maximized and higher on ID samples than on OOD instances. 
(b) The entropy growth is larger in deeper layers. (c) The OOD detection performance is accordingly better in deeper layers.
}
\label{fig:method}
\end{figure}

\noindent
\textbf{A missing component in the conventional norm.}
The network training tends to maximize the  confidence of hidden classifier
\begin{equation}
\max \,
%\lVert \mathbf{a}^{(L)} \rVert_1
%= 
\overline{\psi}_y(\mathbf{x}) 
= \overbrace{\sum_{i:b^{(L)}_{y,i}=1} a^{(L)}_i}^{\text{maximized}}
%\lVert \mathbf{a}^{(L)}\rVert_1}
- \overbrace{\sum_{j:b^{(L)}_{y,j}=-1} a^{(L)}_j}^{\text{minimized}}
\end{equation}
on ID samples $\mathbf{x}$
under a regularity condition (Prop.~\ref{thm:hidden}). This maximization is stronger on ID samples than on OOD instances \cite{fang2022out}, and hence serves as a key factor that separates OOD from ID (Cor.~\ref{thm:separation}).

The maximization of confidence can be disentangled to maximization of the positive summand $A {:=} \sum_{i:b^{(L)}_{y,i}=1} a^{(L)}_i$ and minimization of the negative summand $D {:=} \sum_{j:b^{(L)}_{y,j}=-1} a^{(L)}_j$, which correspond to activation and deactivation of neurons, respectively. 
The conventional $l_1$ feature norm $\lVert \textbf{a} \rVert_1$ captures the maximization trend of activation neurons as the summand $A$ converges to $\lVert \textbf{a} \rVert_1$. However the $l_1$ norm fails to reflect the deactivation responses as the negative summand is diminished with $D {\approx} 0$ due to to the nature of the activation function (\eg. ReLU). Hence, this can lead to potential misidentification of ID samples when the naive $l_1$ norm is used for OOD detection.

\begin{table*}[t]
\centering
\resizebox{.995\linewidth}{!}{
\begin{tabular}{l ccc cc cc cc cc cc c}
\toprule
~ & \multirow{2}{*}{hyper.-free} & \multirow{2}{*}{label-free} & \multirow{2}{*}{bank-free} & \multicolumn{2}{c}{iNaturalist} & \multicolumn{2}{c}{SUN} & \multicolumn{2}{c}{Places} & \multicolumn{2}{c}{Texture} & \multicolumn{2}{c}{Average} & \multirow{2}{*}{ID ACC$\uparrow$} \\ 
~ & ~ & ~ & ~ & AUROC$\uparrow$ & FPR95$\downarrow$ & AUROC$\uparrow$ & FPR95$\downarrow$ & AUROC$\uparrow$ & FPR95$\downarrow$ & AUROC$\uparrow$ & FPR95$\downarrow$ & AUROC$\uparrow$ & FPR95$\downarrow$ & ~ \\ \midrule
\multicolumn{15}{l}{\textbf{\textit{With Supervised Labels of ID:}}} \\
MSP & \cmark &  & \cmark & 93.78 & 29.74 & 84.56 & 59.54 & 84.28 & 60.94 & 84.90 & 50.02 & 86.88 & 50.06 & 78.73 \\ 
Energy & \cmark &  & \cmark & 96.17 & 20.98 & 88.91 & 47.05 & 87.70 & 51.15 & 88.90 & 39.31 & 90.42 & 39.62 & 78.73 \\ 
MaxLogit & \cmark &  & \cmark & 95.99 & 22.06 & 88.43 & 50.90 & 87.37 & 53.78 & 88.42 & 42.25 & 90.05 & 42.25 & 78.73 \\ 
KL & \cmark &  & \cmark & 96.17 & 20.98 & 88.91 & 47.06 & 87.70 & 51.15 & 88.90 & 39.31 & 90.42 & 39.63 & 78.73 \\ 
Mahalanobis & \cmark &  & \cmark & 94.79 & 35.04 & 86.55 & 64.99 & 83.92 & 70.31 & 95.52 & 15.02 & 90.20 & 46.34 & 78.73 \\ 
% GradNorm & \cmark &  & \cmark & 94.89 & 27.54 & 91.92 & 34.91 & 89.16 & 43.69 & 88.91 & 37.80 & 91.22 & 35.99 & 78.73 \\ 
ViM &  &  & \cmark & 95.54 & 27.75 & 89.85 & 48.12 & 87.05 & 57.82 & 95.18 & 14.47 & 91.91 & 37.04 & 78.73 \\ 
SSD &  & \cmark & \cmark & 94.08 & 37.77 & 88.06 & 58.38 & 84.70 & 63.89 & 96.96 & 11.63 & 90.95 & 42.92 & 78.73 \\ 
KNN &  & \cmark &  & 94.15 & 38.25 & 87.75 & 58.19 & 84.93 & 61.80 & 94.24 & 19.29 & 90.27 & 44.38 & 78.73 \\ 
\rowcolor{lightgray}
NAN (ours) & \cmark & \cmark & \cmark & 96.94 & 15.86 & 92.77 & 29.81 & 91.46 & 37.21 & 88.09 & 43.46 & \textbf{92.32} & \textbf{31.59} & 78.73 \\
\midrule
\multicolumn{15}{l}{\textbf{\textit{Without Supervised Labels of ID (detectors based on supervised labels are not available):}}} \\
SSD &  & \cmark & \cmark & 60.34 & 93.87 & 80.89 & 78.41 & 77.23 & 81.26 & 90.19 & 33.53 & 77.16 & 71.77 & 71.10 \\ 
KNN &  & \cmark &  & 84.53 & 78.71 & 82.26 & 76.06 & 77.50 & 80.65 & 91.99 & 24.61 & 84.07 & 65.01 & 71.10 \\ 
\rowcolor{lightgray}
NAN (ours) & \cmark & \cmark & \cmark & 92.90 & 36.09 & 86.76 & 56.27 & 83.22 & 65.08 & 87.57 & 46.86 & \textbf{87.61} & \textbf{51.08} & 71.10 \\
\bottomrule
\end{tabular}
}
\caption{
\textbf{Results on ImageNet-1k} with ResNet-50. `hyper.-free' indicates that the detection score does not require a hyperparameter.
}
\label{table:sota_large}
\end{table*}

\noindent
\textbf{Derivation.}
To mitigate this drawback, we capture the \textit{deactivation tendency} by the sparsity of activations $\lVert \mathbf{a}^{(L)} \rVert_0^{-1}$. The sparsity term reflects the number of deactivated neurons by
\begin{equation}
\lVert \mathbf{a}^{(L)} \rVert_0 
= d_L - |\{ i : a_i^{(L)} \leq 0 \}|.
\end{equation}
 Combining the sparsity term with the conventional vector norm, we derive a novel \textit{negative-aware norm (NAN)}
\begin{equation}
\lVert \mathbf{a} \rVert_{\text{NAN}}
= 
\lVert \mathbf{a}^{(L)} \rVert_1 \cdot \lVert \mathbf{a}^{(L)} \rVert_0^{-1}.
\end{equation}
NAN captures both the activation and deactivation tendencies of ID samples' neurons. Fig.~\ref{fig:method}a shows the sparsity term is higher on ID samples than OOD instances, demonstrating that the deactivation tendency is stronger in ID samples' neurons. Hence, capturing the deactivation tendency likely improves the conventional norm. 
We conduct extensive experiments on NAN in the next section to validate its effectiveness.

We remark that similar to the $l_1$ feature norm, the negative-aware norm (NAN) exhibits class-agnostic characteristics, as verified through analyses of inter/intra-class learning and activation entropy in Sec.~\ref{asec:setup_general_model}.

\noindent
\textbf{Additional consideration.}
We utilize the last hidden layer $\mathbf{a} {=} \mathbf{a}^{(L)}$ for OOD detection as the last hidden layer exhibits a higher growth in activation entropy and accordingly better performance  (Fig.~\ref{fig:method}bc).

\begin{table}[t]
\centering
\resizebox{.7\linewidth}{!}{
\begin{tabular}{l c c}
\toprule
~ & AUROC$\uparrow$ & FPR95$\downarrow$ \\ 
\midrule
NAN & 92.32 & 31.59 \\ 
NAN + KNN \cite{sun2022out} & 92.99 & 29.26 \\
NAN + SSD \cite{sehwag2021ssd} & 93.42 & 27.51 \\
NAN + ReAct \cite{sun2021react} & 93.91 & 29.23 \\ 
NAN + ReAct \cite{sun2021react} + KNN \cite{sun2022out} & 94.37 & 24.94 \\ 
NAN + ReAct \cite{sun2021react} + SSD \cite{sehwag2021ssd} & \textbf{94.61} & \textbf{24.57} \\ 
\bottomrule
\end{tabular}
}
\caption{
Compatibility of NAN to existing detectors. The ID is ImageNet-1k. The value is averaged over all test OOD datasets.
}
\label{table:sota_compatibility}
\end{table}

\begin{table*}[t]
\centering
\resizebox{.995\linewidth}{!}{
\begin{tabular}{l cc cc cc cc cc cc c}
\toprule
OOD & \multicolumn{2}{c}{LSUN-fix} & \multicolumn{2}{c}{ImageNet-fix} & \multicolumn{2}{c}{CIFAR100} & \multicolumn{2}{c}{SVHN} & \multicolumn{2}{c}{Places} & \multicolumn{2}{c}{Average} & \multirow{2}{*}{ID ACC$\uparrow$} \\ 
~ & AUROC$\uparrow$ & FPR95$\downarrow$ & AUROC$\uparrow$ & FPR95$\downarrow$ & AUROC$\uparrow$ & FPR95$\downarrow$ & AUROC$\uparrow$ & FPR95$\downarrow$ & AUROC$\uparrow$ & FPR95$\downarrow$ & AUROC$\uparrow$ & FPR95$\downarrow$ &  \\ 
\midrule
\multicolumn{14}{c}{\textbf{\textit{With} supervised labels of ID}} \\
ODIN* \cite{liang2017enhancing} & - & - & - & - & - & - & 88.3 & 60.4 & 90.6 & 45.5 & - & - & - \\ 
CSI* \cite{tack2020csi} & 92.1 & - & 92.4 & - & 90.5 & - & 96.5 & - & - & - & - & - & - \\ 
MSP & 90.3 & 59.1 & 89.7 & 61.3 & 88.0 & 64.1 & 96.9 & 19.8 & 88.5 & 61.7 & 90.7 & 53.2 & 94.5 \\ 
Energy  & 86.8 & 50.9 & 84.7 & 55.1 & 81.6 & 59.6 & 93.9 & 22.1 & 86.7 & 48.4 & 86.7 & 47.2 & 94.5 \\ 
MaxLogit & 86.8 & 51.7 & 84.7 & 56.0 & 81.6 & 60.1 & 94.1 & 22.0 & 86.6 & 49.8 & 86.8 & 47.9 & 94.5 \\ 
KL & 88.8 & 50.3 & 89.4 & 50.0 & 87.2 & 55.1 & 98.8 & 6.6 & 88.0 & 49.2 & 90.4 & 42.2 & 94.5 \\ 
Mahalanobis & 92.5 & 38.3 & 90.6 & 47.3 & 88.0 & 54.8 & 99.0 & 5.9 & 90.9 & 41.0 & 92.2 & 37.5 & 94.5 \\ 
ViM & 92.8 & 41.0 & 91.3 & 43.7 & 87.3 & 52.5 & 95.0 & 22.5 & 94.1 & 28.2 & 92.1 & 37.6 & 94.5 \\ 
KNN & 96.0 & 25.7 & 95.1 & 31.4 & 92.2 & 44.2 & 99.8 & 1.1 & 94.3 & 32.4 & 95.5 & 27.0 & 94.5 \\ 
SSD & 96.5 & 20.2 & 94.2 & 35.0 & 88.8 & 51.4 & \textbf{99.9} & 0.4 & 92.2 & 42.3 & 94.3 & 29.9 & 94.5 \\ 
\rowcolor{lightgray}
NAN (ours) & 94.7 & 36.6 & 94.5 & 34.4 & 91.7 & 44.8 & 99.7 & 1.3 & 94.2 & 33.3 & 95.0 & 30.1 & 94.5 \\ 
\rowcolor{lightgray}
NAN + KNN & 96.0 & 26.7 & 95.5 & 29.0 & \textbf{92.7} & \textbf{40.9} & \textbf{99.9} & 0.6 & \textbf{94.9} & \textbf{28.2} & \textbf{95.8} & 25.1 & 94.5 \\ 
\rowcolor{lightgray}
NAN + SSD & \textbf{96.7} & \textbf{19.9} & \textbf{95.6} & \textbf{27.6} & 91.8 & 43.6 & \textbf{99.9} & \textbf{0.3} & 94.6 & 30.3 & 95.7 & \textbf{24.3} & 94.5  \\ 
\midrule
\multicolumn{14}{c}{\textbf{\textit{Without} supervised labels of ID}} \\
RotNet* \cite{hendrycks2019using} & 81.6 & - & 86.7 & - & 82.3 & - & 97.8 & - & - & - & - & - & - \\ 
GOAD* \cite{bergman2020classification} & 78.8 & - & 83.3 & - & 77.2 & - & 96.3 & - & - & - & - & - & - \\ 
CSI* \cite{tack2020csi} & 90.3 & - & 93.3 & - & 89.2 & - & 99.8 & - & - & - & - & - & - \\ 
KNN & 95.0 & 30.5 & 93.7 & 36.7 & 89.7 & 50.3 & 99.4 & 3.0 & 88.6 & 58.2 & 93.3 & 35.7 & 90.7 \\ 
SSD & 94.1 & 30.2 & 90.8 & 47.4 & 85.9 & 57.6 & 98.5 & 8.3 & 88.8 & 51.9 & 91.6 & 39.1 & 90.7 \\ 
\rowcolor{lightgray}
NAN (ours) & 94.9 & 28.8 & 93.7 & 36.1 & 88.6 & 52.4 & 96.1 & 22.0 & 89.3 & 51.5 & 92.5 & 38.1 & 90.7 \\ 
\rowcolor{lightgray}
NAN + KNN & 95.8 & 24.6 & \textbf{94.8} & \textbf{32.6} & \textbf{90.1} & \textbf{49.4} & 98.4 & 8.8 & 90.5 & 50.5 & \textbf{93.9} & 33.2 & 90.7 \\ 
\rowcolor{lightgray}
NAN + SSD & \textbf{96.0} & \textbf{21.3} & 94.5 & 33.6 & 89.4 & 49.7 & \textbf{98.5} & \textbf{8.3} & \textbf{91.2} & \textbf{45.6} & \textbf{93.9} & \textbf{31.7} & 90.7 \\ 
\bottomrule
\end{tabular}
}
\vspace{-2mm}
\caption{
\textbf{Results on CIFAR-10} with ResNet-18. * indicates the values are taken from the references.
}
\label{table:sota_small}
\vspace{-2mm}
\end{table*}

\section{Experiments on NAN}
\label{sec:exp}

The objective of this experiment is to assess the OOD detection capabilities of NAN across diverse configurations using general discriminative models. To achieve this goal, we evaluate NAN's performance using both supervised and self-supervised models, and assess it in large-scale and small-scale benchmarks, including the one-class classification setting.
Additionally, we consider the compatibility of NAN, namely, whether NAN can be combined with other detectors for performance gain. We conclude this section with ablation studies of NAN. A detailed description of the complete experiment setup can be found in Sec.~\ref{asec:setup_exp_nan}.

\noindent
\textbf{Evaluation metrics}
The performance is reported by the widely-used metrics: (1) the area under the receiver operating characteristic curve (AUROC), (2) the false positive rate (FPR95) on the OOD samples when the true positive rate of ID samples is at 95\%, (3) closed-set classification accuracy (ACC) of ID.

\subsection{Evaluation on large-scale benchmark}

%\paragraph{Setup}
\noindent
\textbf{Setup.}
We utilize a ResNet-50 trained on ImageNet-1k. The model is trained either by (1) supervised labels using the contrastive loss \cite{khosla2020supervised} or (2) self-supervised instance discrimination loss using momentum embeddings \cite{chen2020improved}. In the case of the supervised contrastive learning, the classification layer is learned after training and freezing the backbone representation. For fair comparison, all detection scores are applied on the same backbone.

% All baseline OOD scores are implemented following the corresponding references. Particularly, the Mahalanobis (MH) score follows the procedure given by SSD \cite{sehwag2021ssd}. For a fair comparison, baseline OOD scores are evaluated on the same model backbone (except for scores that require score-specific training \eg GODIN \cite{hsu2020generalized}).

Following the widely-used ImageNet-1k benchmark \cite{huang2021mos}, we test against four test OOD datasets: fine-grained plant images of iNaturalist \cite{van2018inaturalist}, scene images from SUN \cite{xiao2010sun} and Places \cite{zhou2017places}, and texture images from Texture \cite{cimpoi2014describing}. All OOD datasets are processed so that no overlapping category is present with ImageNet-1k. 

%\paragraph{NAN is comparable to the state-of-the-art on ImageNet-1k}
\noindent
\textbf{Results.}
Table \ref{table:sota_large} shows that NAN is comparable to the state-of-the-art detectors on the ImageNet-1k benchmark. Compared to the OOD detection scores that require a supervised classification layer (\ie MSP, Energy, MaxLogit, and KL), NAN shows significant improvement on both AUROC and FPR95. Moreover, NAN can be instantly applied to the contrastive models without a classification layer and label supervision. 

Distance-based scores (Mahalanobis, SSD, and KNN) outperform NAN on the far-OOD dataset Texture. This is because NAN inherently is a classifier confidence, which can exhibit overconfidence when dealing with far OOD instances.
On average, however, NAN is more robust and produces a significant reduction on the FPR95 metric (11-26\%) without any hyperparameter. Rather than competing with the state-of-the-art distanced-based detectors,
we show NAN can be integrated with them easily for further improvement.

\subsection{Evaluation on NAN compatibility}
We examine whether NAN can be integrated with existing OOD scores. To this end, we consider the state-of-the-art perturbation method ReAct \cite{sun2021react} and the label-free distance-based scores SSD and KNN. NAN is combined with SSD and KNN by simple score division as follows: given a distance function to the ID bank set or prototypes in the form of $d(\mathbf{x}, X_{bank})$, we re-calibrate the distance by $d(\mathbf{x}, X_{bank}) / \lVert \mathbf{a}^{(L)} \rVert_{\text{NAN}}$ where $\mathbf{a}^{(L)}$ is the last hidden layer feature of the test input $\mathbf{x}$. Table \ref{table:sota_compatibility} shows that the combination improves both metrics in all cases, demonstrating the compatibility of NAN.

\subsection{Evaluation on standard benchmark}
\label{sec:exp_standard}
We evaluate NAN on the standard CIFAR-10 benchmark that consists of low-resolution images.

\noindent
\textbf{Setup.} 
We utilize a ResNet-18 trained on CIFAR10. The model is trained by either of the two standard training schemes: cross-entropy minimization with supervised labels  and self-supervised learning (MoCo-v2) without the supervised labels. Following the popular benchmark, we choose the following datasets as OOD test datasets: LSUN-fix \cite{tack2020csi}, ImageNet-fix \cite{tack2020csi}, CIFAR100 \cite{krizhevsky2009learning}, SVHN \cite{netzer2011reading}, and Places \cite{zhou2017places}. All images are of size 32 $\times$ 32.

\noindent
\textbf{Evaluation results.}
Table \ref{table:sota_small} shows that the proposed score NAN is comparable to state-of-the-art scores specifically designed for OOD detection.  We highlight that only NAN is a hyperparameter-free approach among the top-performing methods. The label-free distance-based scores KNN and SSD exhibit robustness, but their results are attained by carefully fine-tuning their method-specific hyperparameters. 
Despite not utilizing any hyperparameters, NAN exhibits comparable performance to the label-free state-of-the-art detectors (SSD and KNN) in terms of AUROC and FPR95 metrics on average.
CSI also shows marginal superiority in two cases out of eight, but CSI requires  complicated training with image rotation prediction, and its inference must be combined with KNN in an intricate manner. In contrast, NAN is simple and can be easily integrated to KNN. Combined with the distance-based scores SSD and KNN, NAN exhibits a consistent performance boost and outperforms all reported detectors.

\subsection{Evaluation on one-class classification}
As NAN requires neither classifier nor supervised labels, it can be applied to one-class classification (OCC). To assess the OCC performance, we evaluate the standard one-class benchmark of CIFAR-10/100.
A class randomly chosen in CIFAR-10 is regarded as the ID data, and the rest of the 9 classes in CIFAR-10 constitute OOD instances. We conduct a similar experimental procedure on CIFAR-100 superclasses. For a fair comparison, we compare with one-class classification baselines that do not utilize extra training data and pretrained weights attained from large-scale data. For evaluation, we apply NAN on the MoCo-v2 model that is trained on the one-class data from scratch.

Table \ref{table:sota_occ} indicates that NAN is comparable to the state-of-the-art one-class classifier CSI without any complicated training and hyperparameter tuning. Combined with the distance-based detectors, NAN performs equally well and improves the distance-based detectors on both CIFAR-10/100 data sets.

\begin{table}[t]
\centering
\resizebox{.65\linewidth}{!}{
\begin{tabular}{lll}
\toprule
~ & CIFAR10 & CIFAR100 \\ 
\midrule
\multicolumn{3}{l}{\textit{\textbf{Without} bank set:}} \\
OC-SVM* \cite{ocsvm} & 58.8 & 63.1 \\ 
Deep-SVDD* \cite{ruff2018deep} & 64.8 & - \\
AnoGAN* \cite{schlegl2017unsupervised} & 61.8 & - \\
OCGAN* \cite{perera2019ocgan} & 65.7 & - \\
Geom* \cite{geom} & 86.0 & 78.7 \\ 
GOAD* \cite{goad} & 88.2 & - \\ 
NAN (ours) & \textbf{93.7} & \textbf{88.2} \\ 
\midrule
\multicolumn{3}{l}{\textit{\textbf{With} bank set:}} \\
% SSD* & 90.0 & - \\ 
CSI* & \textbf{94.3} & - \\ 
SSD & 91.1 & 85.7 \\ 
SSD + NAN (ours) & \textbf{94.3}{\footnotesize(\textbf{+3.2})} & \textbf{88.7}{\footnotesize(\textbf{+2.0})} \\ 
KNN & 92.1 & 87.1 \\ 
KNN + NAN (ours) & \textbf{94.3}{\footnotesize(\textbf{+2.2})} & 88.3{\footnotesize(\textbf{+1.0})} \\
\bottomrule
\end{tabular}
}
\caption{
The average one-class classification (OCC) performance in AUROC. * indicates the values are taken from the references.
}
\label{table:sota_occ}
\end{table}

\subsection{Ablation study}
% \paragraph{Ablation on the NAN Score}
\noindent
\textbf{Ablation on the NAN Score}
The primary innovation of NAN is the inclusion of a sparsity term (\ie, the denominator of NAN), which accounts for the hidden layer neurons' tendency to deactivate. We analyze the impact of this component by ablating it. Table \ref{table:ablation_na} shows the effectiveness of the sparsity term in both large-scale and small-scale settings. In the large-scale setting (ImageNet-1k), OOD is mostly differentiated from ID by the deactivation tendency of hidden layer neurons. In the case of the small-scale CIFAR-10 dataset, capturing both deactivation and activation tendencies is crucial for enhancing the OOD detection performance.
In general, the inclusion of the sparsity term to capture the deactivation tendency enhances the robustness of the OOD detection score.

\begin{table}[t]
\centering
\resizebox{.95\linewidth}{!}{
\begin{tabular}{l c c c c}
\toprule
~ & \multicolumn{2}{c}{ImageNet-1k} & \multicolumn{2}{c}{CIFAR-10}   \\ 
~ & AUROC$\uparrow$ & FPR95$\downarrow$ & AUROC$\uparrow$ & FPR95$\downarrow$  \\
\midrule
%embedding magnitude & 84.09 & 72.85 & 93.00 & 43.40  \\ 
NAN w/o sparsity term & 57.99 & 95.22 & 92.40 & 43.00  \\ 
NAN & \textbf{92.32} & \textbf{31.59} & \textbf{94.90} & \textbf{30.10} \\
\bottomrule
\end{tabular}
}
\caption{
The ablation study examines the effect of NAN's sparsity term, which accounts for the hidden layer neurons' deactivation tendency. The ID is either ImageNet-1k or CIFAR-10. The value is averaged over all corresponding test OOD datasets. 
}
\label{table:ablation_na}
\end{table}

\begin{figure}[t]
\centering
\includegraphics[width=.95\linewidth]{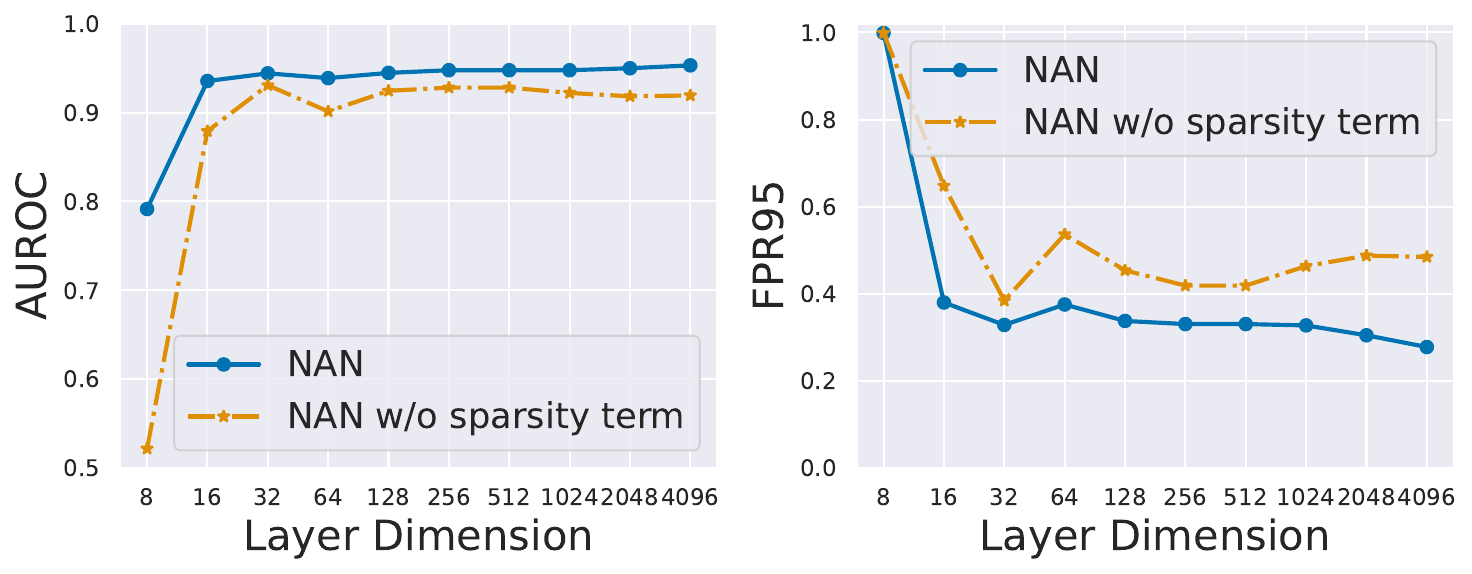}
\caption{
The ablation study of NAN with respect to the dimension $d_L$ of the last hidden layer $\mathbf{a}^{(L)}$. The ID data is CIFAR-10. The reported metric numbers are values averaged over test OOD datasets.
}
\label{fig:ablation_dim}
\end{figure}

\noindent
\textbf{Ablation on the Architectural Component: the Last Hidden Layer Dimension}
Although NAN is a hyperparameter-free OOD score, its effectiveness is still influenced by the network architecture, much like other detection scores. Specifically, the performance of NAN may primarily depend on the dimension $d_L$ of the last hidden layer $\mathbf{a}^{(L)}$. To assess the impact of this dimension on the performance of NAN, we evaluate NAN on multiple ResNet-18 models with different dimensions $d_L$. We train the models on CIFAR-10 using supervised cross-entropy loss and evaluate them on various OOD datasets, including LSUN-fix, ImageNet-fix, CIFAR-100, and SVHN. We report the average performance over all test OOD datasets.

We hypothesize that a wider hidden layer would better capture the deactivation tendency of neurons, and hence improve the performance. Fig.~\ref{fig:ablation_dim} evidences the hypothesis; increasing the dimension of the last hidden layer tends to improve the performance of NAN. Particularly on the FPR95 metric, the improvement is not marginal. Moreover, the performance is fairly robust unless the layer dimension is unreasonably small. 
Interestingly, the comparison between NAN and the standard $l_1$-norm score without the sparsity term unveils an intriguing finding; NAN's ability to capture the deactivation tendency makes the score more robust to changes in the layer dimension $d_L$. This result suggests that measuring the deactivation tendency is critical for effective OOD detection.

\noindent
\textbf{Additional ablations and limitation.}
Further ablation on architectural components and the limitation of NAN are given in Sec.~\ref{asec:more_ablation} and \ref{asec:limitation}, respectively.

\section{Conclusion}

We have conducted a thorough investigation of the feature norm to gain insights into its underlying mechanism for OOD detection. Specifically, we have demonstrated that the feature norm's ability to detect OOD stems from its function as classifier confidence. Additionally, we have established that the feature norm can detect OOD using any discriminative model, making it independent of class label type. Through our formulation of the feature norm as a hidden classifier, we have identified that the conventional feature norm neglects neurons that tend to deactivate, leading to the potential misidentification of ID samples. To address this limitation, we have proposed a novel negative-aware norm NAN that captures both the activation and deactivation tendencies of hidden layer neurons. Our empirical results have demonstrated the effectiveness of NAN across diverse OOD detection benchmarks.

\paragraph{Acknowledgments}
This work was supported by the Materials/Parts Technology Development Program grant funded by the Korea government (MOTIE) (No.~1415187441) and the National Research Foundation of Korea (NRF) grant funded by the Korea government (MSIP) (No.~NRF-2022R1A2C1010710).

% We systematically justified the usage of feature norm for OOD detection by interpreting the feature norm as a confidence value of a hidden classifier obtained by network weight binarization. Based on the classifier-based interpretation, we further discovered that the feature norm is a class-agnostic OOD detector, and that its performance is correlated to the diversity of activation patterns in the layer. Moreover, by generalizing the notion of feature norm based on its classifier-based interpretation, we proposed a novel norm-based indicator, negative-aware norm (NAN), which can effectively capture both activation and deactivation patterns of feature units. We verified the effectiveness of NAN on both small-scale and large-scale OOD detection benchmarks.
%of instance discriminator for unsupervised OOD detection. Notably, we verified the behavioral equivalence between the final ReLU activation norm and the maximum logit of a classifier. We observed that a discriminative labeling scheme with semantic consistency in intra-class samples could enable OOD detection. Based on the observed principles, we proposed using the final ReLU layer for deriving our proposed norm-based scores and showed the effectiveness thereof for both independent and combined usages with distanced-based OOD detectors.

%%%%%%%%% REFERENCES
{\small
\bibliographystyle{ieee_fullname}
\bibliography{egbib}
}

%%% SUPPLEMENTARY MATERIALS - START
\clearpage

\appendix

\onecolumn

\setcounter{thm}{0}

\part*{Supplementary Materials}

\section{Supplementary to the Analysis of Hidden Classifier}
\label{asec:theory}

\subsection{Proofs for the properties of hidden classifier}
\label{asec:theory_main_proof}

\paragraph{Notation (detailed)}
Each hidden layer feature $\mathbf{a}^{(l)}$ is defined by consecutive computation of the post-activated feature vector
\begin{equation}
\mathbf{a}^{(l)} = \sigma(\mathbf{W}^{(l)T} \mathbf{a}^{(l-1)})
\end{equation}
from the input layer $l=0$ to the last hidden layer $l=L$. The pre-activated features satisfy $\mathbf{a}^{(l)} = \sigma( \mathbf{z}^{(l)} )$, where the activation function $\sigma$ is a rectifier (\eg ReLU, GeLU, Leaky ReLU).
The penultimate embedding is $g(\mathbf{x}) = \mathbf{U}^T\mathbf{a}^{(L)}$, which computes the network classification logit $\psi(\mathbf{x}) \in \mathbb{R}^K$ by 
\begin{equation}
\psi(\mathbf{x}) = \mathbf{W}^Tg(\mathbf{x}).
\end{equation}
$\mathbf{W}$ is the weight matrix for the classification layer.
%In the main text, we assumed $\mathbf{N}$ is an identity matrix for notation simplicity. Here, we assume $N$ is either an identity matrix or $N = (1/T)\diag(\frac{1}{\lVert \mathbf{w}_1 \rVert_2 \lVert g(\mathbf{x}) \rVert_2}, \dots, \frac{1}{\lVert \mathbf{w}_K \rVert_2 \lVert g(\mathbf{x}) \rVert_2})$ where $T$ is a temperature. The latter case is for networks whose final outputs are scaled cosine similarities. 
For notation simplicity, let $\mathbf{W}^{(L+1)} := \mathbf{U}\mathbf{W}$ such that $\psi(\mathbf{x}) = \mathbf{W}^{(L+1)} \mathbf{a}^{(L)}$.
The sign function $\sign(\cdot)$, on the other hand, that binarizes a scalar to either $1$ or ${-}1$ is applied point-wise.

\paragraph{Note}
For the embedding computation, $\mathbf{U}$ is a fixed identity matrix in supervised models, while $\mathbf{U}$ serves as a learnable parameters for self-supervised models with projection head \cite{chen2020improved}.

\begin{prop}
\label{prop:supp_decom}
The final logit is represented by
\begin{equation}
\psi(\mathbf{x}) = \mathbf{C}^{(l)} \mathbf{a}^{(l)}
\end{equation}
for each hidden layer $l$, where 
\begin{equation}
\mathbf{C}^{(l)} = \left(\prod_{k=0}^{L-l-1} \mathbf{W}^{(L+1-k)T} \mathbf{D}^{(L-k)} \right) \mathbf{W}^{(l+1)T}
\end{equation}
with $\mathbf{D}^{(l)} = \diag(\frac{\sigma(z_1)}{z_1}, \dots, \frac{\sigma(z_{d_l})}{z_{d_l}})$ with the convention $\frac{\cdot}{0} = 0$. $\mathbf{C}^{(l)} = \mathbf{C}^{(l)}(\mathbf{x}) \in \mathbb{R}^{K \times d_l}$ depends on $\mathbf{x}$.
\end{prop}

\begin{proof}
Observe inductively that
\begin{alignat}{2}
\psi(\mathbf{x}) & = \textbf{W}^{(L+1)T} \mathbf{a}^{(L)} \\
& = \textbf{W}^{(L+1)T} \mathbf{D}^{(L)} \mathbf{z}^{(L)} \\
& = \textbf{W}^{(L+1)T} \mathbf{D}^{(L)} \textbf{W}^{(L)T} \mathbf{a}^{(L-1)} \\
& = \textbf{W}^{(L+1)T} \mathbf{D}^{(L)} \textbf{W}^{(L)T} \mathbf{D}^{(L-1)} \mathbf{z}^{(L-1)} \\
& = \cdots,
\end{alignat}
obtaining
\begin{equation}
\psi (\mathbf{x})
=
\left(\prod_{k=0}^{L-l-1} \mathbf{W}^{(L+1-k)T} \mathbf{D}^{(L-k)} \right) \mathbf{W}^{(l+1)T} \mathbf{a}^{(l)}
\end{equation}
\end{proof}

\begin{remk*}
We note that both $\mathbf{D}^{(l)} = \mathbf{D}^{(l)}(\mathbf{x})$ and $\mathbf{C}^{(l)} = \mathbf{C}^{(l)}(\mathbf{x})$ depend on $\mathbf{x}$ as they depend on $\mathbf{a}^{(l)}$. Also, note that the dimension of $\mathbf{C}^{(l)}$ is $K \times d_l$.
\end{remk*}

%\paragraph{Terminology}
%We say that a signal $Q = Q(\theta)$ of a network parameterized by the parameter $\theta$ is maximized if there exists a sequence $(\theta_n)_{n=1}^\infty$ of parameters such that $Q(\theta_n) \to M:= \max \{ Q(\theta) : \theta \in \Theta \}$ where $\Theta$ is the set of all possible parameters. 
%To avoid any unnecessary mathematical complication, we assume $M = \infty$. Namely, the optimization is unconstrained. Minimization of the signal is similarly defined. 

%\begin{remk*}
%$Q(\theta_n) \to M=\infty$ practically means that the signal $Q(\theta_n)$ has an increasing trend.
%\qed
%\end{remk*}

%Below, we let $(\mathbf{x}, y)$ denote a labeled sample. 
Recall that $\mathbf{C}^{(l)} = [\mathbf{c}^{(l)}_1, \dots, \mathbf{c}^{(l)}_K]^T$.

\begin{prop}
\label{prop:supp_hidden}
Let $(\mathbf{x},y)$ be arbitrary labeled ID sample.
Suppose that $\psi_y(\mathbf{x})$ is maximized in a manner to reduce the angle between $\mathbf{c}^{(l)}_y$ and $\mathbf{a}^{(l)}$ sufficiently that $\sign(\mathbf{c}^{(l)}_y) = \sign(\mathbf{a}^{(l)})$. 
Suppose that $\psi_k(\mathbf{x})$ is minimized in a manner to increase the angle between $\mathbf{c}^{(l)}_k$ and $\mathbf{a}^{(l)}$ sufficiently that $\measuredangle(\sign(\mathbf{c}^{(l)}_k), \mathbf{a}^{(l)}) > \pi/2$. 
Then, $\overline{\psi}^{(l)}$ becomes a discriminative classifier with $\overline{\psi}^{(l)}_y(\mathbf{x}) > \overline{\psi}^{(l)}_k(\mathbf{x})$.
\end{prop}

\begin{proof}
For notational simplicity, ignore the superscript index $l$, and let $\mathbf{a} = \mathbf{a}^{(l)}$, $\mathbf{b}_k = \mathbf{b}^{(l)}_k$, $\mathbf{c}_k = \mathbf{c}^{(l)}_k$, and $\overline{\psi} = \overline{\psi}^{(l)}$.
First, observe $\mathbf{b}_y = \sign(\mathbf{c}_y) = \sign(\mathbf{a})$ implies $0 \leq \measuredangle(\mathbf{b}_y, \mathbf{a}) < \pi/2$. Therefore, 
\begin{equation}
\overline{\psi}_y(\mathbf{x}) = \mathbf{b}_y \cdot \mathbf{a} 
= \lVert \mathbf{b}_y \rVert_2 \lVert \mathbf{a} \rVert_2 
\cos ( \measuredangle(\mathbf{b}_y, \mathbf{a}) ) > 0.
\end{equation}
On the other hand, $\measuredangle(\sign(\mathbf{c}_k), \mathbf{a}) > \pi/2$ means $\pi \geq \measuredangle( \mathbf{b}_k, \mathbf{a}) > \pi/2$
by the definition of $\mathbf{b}_k$ for $k \neq y$. Therefore, 
\begin{equation}
\overline{\psi}_k(\mathbf{x}) = \mathbf{b}_k \cdot \mathbf{a} 
= \lVert \mathbf{b}_k \rVert_2 \lVert \mathbf{a} \rVert_2 
\cos ( \measuredangle(\mathbf{b}_k, \mathbf{a}) ) < 0.
\end{equation}
Since $(\mathbf{x},y)$ was arbitrary, we have proved the desired.
\end{proof}

The main message of Prop.~\ref{prop:supp_hidden} is that the discriminative optimization of the original classifier should be powerful enough to optimize the \textit{angle} between the hidden layer feature and the binary weight. Then, in this case, the hidden classifier becomes discriminative.

\begin{thm}
\label{thm:supp_approx}
Under the sufficient condition of Prop.~\ref{thm:hidden}, for any labeled ID sample $(\mathbf{x},y)$,
\begin{equation}
\lVert \mathbf{a}^{(l)} \rVert_1
\text{ converges to }
\overline{\psi}^{(l)}_{y} (\mathbf{x})
= \max_k \overline{\psi}^{(l)}_{k} (\mathbf{x})
\end{equation}
in which case $\sign(\mathbf{a}^{(l)}) = \mathbf{b}^{(l)}_y$.
In general, for any $k$ and for any sample $\mathbf{x}$ (either ID or OOD),
\begin{equation}
0 \leq \lVert \mathbf{a}^{(l)} \rVert_1 - \overline{\psi}^{(l)}_k(\mathbf{x}) \leq \lVert \mathbf{a}^{(l)} \rVert_\infty
\lVert \sign(\mathbf{a}^{(l)}) - \mathbf{b}^{(l)}_{k} \rVert_1.
\end{equation}
\end{thm}

%Here, the sign function is applied element-wise, and defined by
%\begin{equation}
%\sign(x) = 
%\begin{cases}
%1 & \text{ if } x>0 \\
%-1 & \text{ if } x \leq 0
%\end{cases}
%\end{equation}
%for a scalar value $x$. 

\begin{proof}
For notational simplicity, ignore the superscript index $l$, and let $\mathbf{a} = \mathbf{a}^{(l)}$, $\mathbf{b}_k = \mathbf{b}^{(l)}_k$, $\mathbf{c}_k = \mathbf{c}^{(l)}_k$, and $\overline{\psi} = \overline{\psi}^{(l)}$.
First, observe that
\begin{equation}
\lVert \mathbf{a} \rVert_1
= \sum_i | a_i |
\geq \sum_i b_{ki} a_i
= \mathbf{b}_k \cdot \mathbf{a} = \overline{\psi}_k(\mathbf{x})
\end{equation}
where $\mathbf{b}_k = (b_{k1}, \dots, b_{kd_l}) \in \{-1, 1\}^{d_l}$. This proves that $\lVert \mathbf{a} \rVert_1 \geq \overline{\psi}_k(\mathbf{x})$ for all $k$.

Now, observe that $|a_i| = \sign(a_i) a_i$. Therefore,
\begin{equation}
\lVert \mathbf{a} \rVert_1 - \overline{\psi}_k(\mathbf{x})
= \sum_i ( \sign(a_i) - b_{ki} ) a_i
\leq \sum_i \lvert \sign(a_i) - b_{ki} \rvert \lvert a_i \rvert
\leq
\lVert \mathbf{a} \rVert_\infty
\lVert \sign(\mathbf{a}) - \mathbf{b}_{k} \rVert_1,
\end{equation}
proving a general upper bound of the difference between the hidden classifier output and the feature norm.

Now, under the sufficient condition of Prop.~\ref{prop:supp_hidden}, the binary weight becomes the activation pattern by the assumption; $\sign(\mathbf{a}) = \mathbf{b}_y$. Therefore, in this case,
\begin{equation}
0 \leq \lVert \mathbf{a} \rVert_1 - \overline{\psi}_y(\mathbf{x}) \leq \lVert \mathbf{a} \rVert_\infty \cdot 0 = 0,
\end{equation}
proving the desired.
\end{proof}

\begin{cor}
If $\max_k \overline{\psi}^{(l)}_k(\mathbf{x}_{ood})$ is sufficiently small such that
\begin{equation}
\max_k \overline{\psi}^{(l)}_k(\mathbf{x}_{ood}) + \delta
< \max_k \overline{\psi}^{(l)}_k (\mathbf{x}_{ind})
\end{equation}
for all ID samples $\mathbf{x}_{ind}$
where 
\begin{equation}
\delta \geq
\lVert \mathbf{a}^{(l)} (\mathbf{x}_{ood}) \rVert_\infty 
\cdot
\lVert \sign(\mathbf{a}^{(l)} (\mathbf{x}_{ood})) - \mathbf{b}^{(l)}_{k_0} \rVert_1
\end{equation}
and $k_0 = \arg \max_k  \overline{\psi}^{(l)}_k(\mathbf{x}_{ood})$, then
\begin{equation}
\lVert \mathbf{a}^{(l)}(\mathbf{x}_{ood}) \rVert_1 
< 
\lVert \mathbf{a}^{(l)}(\mathbf{x}_{ind}) \rVert_1
\end{equation}
for all ID samples $\mathbf{x}_{ind}$.
\end{cor}

\begin{proof}
By Thm.~\ref{thm:supp_approx},
\begin{equation}
\lVert \mathbf{a}^{(l)}(\mathbf{x}_{ood}) \rVert_1
\leq 
\overline{\psi}^{(l)}_{k_0} (\mathbf{x}_{ood}) + \lVert \mathbf{a}^{(l)}(\mathbf{x}_{ood}) \rVert_\infty
\lVert \sign(\mathbf{a}^{(l)})(\mathbf{x}_{ood}) - \mathbf{b}^{(l)}_{k_0} \rVert_1
<
\max_k \overline{\psi}^{(l)}_k (\mathbf{x}_{ind})
\leq 
\lVert \mathbf{a}^{(l)}(\mathbf{x}_{ind}) \rVert_1.
\end{equation}
\end{proof}

\subsection{Additional Theoretical Consideration}

We present additional results of the theoretical analysis on the hidden classifier.

\subsubsection{Relation to General $l_p$-norms}

We have proved that $l_1$-norm can differentiate OOD from ID. This capability of $l_1$-norm extends to the general $l_p$-norm by Holder's inequality.

\begin{thm}[Holder's inequality]
\label{thm:holder}
For $0 < p \leq q < \infty$ and $\mathbf{x} \in \mathbb{R}^d$, 
\begin{equation}
\lVert \mathbf{x} \rVert_q \leq \lVert \mathbf{x} \rVert_p 
\leq d^{1/p - 1/q} \lVert \mathbf{x} \rVert_q.
\end{equation}
\end{thm}

%\begin{thm}
%For $0 < p \leq q < \infty$, $\lVert \mathbf{x} \rVert_q \leq \lVert \mathbf{x} \rVert_p$
%\end{thm}
%\begin{proof}
%For a fixed vector $\mathbf{x} \in \mathbb{R}^n$, the mapping $p \mapsto \lVert \mathbf{x} \rVert_p$ is a non-increasing function for $p > 0$.
%\end{proof}
%
%\begin{thm}[Corollary of Holder's Inequality]
%For a vector $\mathbf{x} \in \mathbb{R}^d$, $\lVert \mathbf{x} \rVert_p \leq d^{1/p - 1/q} \lVert \mathbf{x} \rVert_q$ where $p < q $.
%\end{thm}

Thus, for an activation vector $\mathbf{a}^{(l)} \in \mathbb{R}^{d_l}$ and for $p>1$, we have
\begin{equation}
\lVert \mathbf{a}^{(l)} \rVert_p
\leq
\lVert \mathbf{a}^{(l)} \rVert_1
\leq d_l^{-1/p} \lVert \mathbf{a}^{(l)} \rVert_p
\end{equation}
Therefore, if $\lVert \mathbf{a}^{(l)} \rVert_1$ is large or small, then $\lVert \mathbf{a}^{(l)} \rVert_p$ is also large or small, respectively. Thus, different $l_p$-norms have similar mechanisms for OOD detection. Note, however, that different $l_p$-norms have different priors on the computation of units in the activation vector. Accordingly, the OOD detection performance will vary depending on which $l_p$-norm is used.

\subsubsection{Extension to Pre-Activation Layer}

Extending the framework in Sec.~\ref{sec:theory} to the pre-activation layer feature vector $\mathbf{z}^{(l)}$ is trivial, where the pre-activation layer feature is the vector satisfying $\mathbf{a}^{(l)} = \sigma ( \mathbf{z}^{(l)} )$ with the activation function $\sigma$. Here, we provide the properties of the pre-activation layer that correspond to the ones given in Sec.~\ref{sec:theory}.

\begin{prop}
\label{prop:supp_decom_pre}
The final logit is represented by
\begin{equation}
\psi(\mathbf{x}) = \widehat{\mathbf{C}}^{(l)} \mathbf{z}^{(l)}
\end{equation}
for each hidden layer $l$, where 
\begin{equation}
\widehat{\mathbf{C}}^{(l)} = \left(\prod_{k=0}^{L-l} \mathbf{W}^{(L+1-k)T} \mathbf{D}^{(L-k)} \right) 
\end{equation}
with $\mathbf{D}^{(l)} = \diag(\frac{\sigma(z_1)}{z_1}, \dots, \frac{\sigma(z_{d_l})}{z_{d_l}})$ with the convention $\frac{\cdot}{0} = 0$. $\widehat{\mathbf{C}}^{(l)} = \widehat{\mathbf{C}}^{(l)}(\mathbf{x}) \in \mathbb{R}^{K \times d_l}$ depends on $\mathbf{x}$.
\end{prop}

Define a hidden classifier corresponding to $\mathbf{z}^{(l)}$ by
\begin{equation}
\widehat{\psi}(\mathbf{x}) := \sign(\widehat{C}^{(l)}) \mathbf{z}^{(l)} 
= \widehat{\mathbf{B}}^{(l)} \mathbf{z}^{(l)}
\end{equation}
where $\widehat{\mathbf{C}}^{(l)} = [\widehat{\mathbf{c}}^{(l)}_1, \dots, \widehat{\mathbf{c}}^{(l)}_K]^T$ and $\widehat{\mathbf{B}}^{(l)} = [\widehat{\mathbf{b}}^{(l)}_1, \dots, \widehat{\mathbf{b}}^{(l)}_K]^T$.

\begin{prop}
\label{prop:supp_hidden_pre}
Let $(\mathbf{x},y)$ be an arbitrary labeled sample.
Suppose that $\psi_y(\mathbf{x})$ is maximized in a manner to reduce the angle between $\widehat{\mathbf{c}}^{(l)}_y$ and $\mathbf{z}^{(l)}$ sufficiently that $\sign(\widehat{\mathbf{c}}^{(l)}_y) = \sign(\mathbf{z}^{(l)})$. 
Suppose that $\psi_k(\mathbf{x})$ is minimized in a manner to increase the angle between $\widehat{\mathbf{c}}^{(l)}_k$ and $\mathbf{z}^{(l)}$ sufficiently that $\measuredangle(\sign(\widehat{\mathbf{c}}^{(l)}_k), \mathbf{z}^{(l)}) > \pi/2$. 
Then, $\widehat{\psi}^{(l)}$ becomes a discriminative classifier with $\widehat{\psi}^{(l)}_y(\mathbf{x}) > \widehat{\psi}^{(l)}_k(\mathbf{x})$.
\end{prop}

\begin{thm}
\label{thm:supp_approx_pre}
Under the sufficient condition of Prop.~\ref{prop:supp_hidden_pre}, 
\begin{equation}
\lVert \mathbf{z}^{(l)} \rVert_1
\text{ converges to }
\widehat{\psi}^{(l)}_{y} (\mathbf{x})
= \max_k \widehat{\psi}^{(l)}_{k} (\mathbf{x})
\end{equation}
in which case $\sign(\mathbf{z}^{(l)}) = \widehat{\mathbf{b}}^{(l)}_y$.
In general, for any $k$
\begin{equation}
0 \leq \lVert \mathbf{z}^{(l)} \rVert_1 - \widehat{\psi}_k(\mathbf{x}) \leq 
\lVert \mathbf{z}^{(l)} \rVert_\infty
\lVert \sign(\mathbf{z}^{(l)}) - \widehat{\mathbf{b}}^{(l)}_k \rVert_1
\end{equation}
\end{thm}

\subsubsection{On Bias}
In Sec.~\ref{sec:theory}, we ignored the bias in the computation of features for simplicity. We can preserve the properties of features given in Sec.~\ref{sec:theory} while including the bias terms.
To observe this, consider
\begin{equation}
\mathbf{a}^{(l)} = \sigma( \mathbf{W}^{(l)T} \mathbf{a}^{(l-1)} + \boldsymbol{\beta}^{(l)})
= \mathbf{D}^{(l)} \mathbf{W}^{(l)T} \mathbf{a}^{(l-1)} + \mathbf{D}^{(l)} \boldsymbol{\beta}^{(l)}.
\end{equation}
Thus, if $\Psi$ denotes the logit computed with bias, then
\begin{equation}
\Psi(x) = \mathbf{C}^{(l)} \mathbf{a}^{(l)}
+ \sum_{j=l}^L \widehat{\mathbf{C}}^{(j+1)} \boldsymbol{\beta}^{(j+1)}
= \psi(x) + \boldsymbol{\Gamma}
\end{equation}
with $\boldsymbol{\Gamma} = \boldsymbol{\Gamma}(l,\mathbf{x}) = \sum_{j=l}^L \widehat{\mathbf{C}}^{(j+1)} \boldsymbol{\beta}^{(j+1)}$ and the convention that $\mathbf{\widehat{C}}^{(L+1)} = \mathbf{I}$. Hence, if the discriminative learning of $\Psi$ is not trivially achieved by the optimization of the bias term $\boldsymbol{\Gamma}$, and if the discriminative learning of $\psi$ is thus sufficiently powerful, then the properties in Sec.~\ref{sec:theory} hold.

%We represent the second term on the RHS by
%\begin{equation}
%\sum_{j=l}^L \widehat{\mathbf{C}}^{(j+1)} \beta^{(j+1)}
%\end{equation}

%Indeed, the bias can be ignored due to the linear form of layer-wise computation. To observe this, temporarily, assume that the pre-activated unit is computed by
%\begin{equation}
%\mathbf{z}^{(l+1)} = \mathbf{W}^{(l+1)T} \mathbf{a}^{(l)} + \mathbf{b}^{(l+1)}.
%\end{equation}
%Then, the final logit 
%\begin{equation}
%\psi(\mathbf{x}) = \mathbf{C}^{(l)} \psi^{(l)}(\mathbf{x}) + \sum_{j=l}^{L} \mathbf{C}^{(j+1)} \mathbf{b}^{(j+1)}
%\end{equation}
%with the convention that $\mathbf{C}^{(L+1)} = \mathbf{I}$ is the identity matrix.
%Hence, the computation of bias can be linearly decoupled. Accordingly, the bias can be ignored in our context.

\subsubsection{On Cosine Similarity Logit}
We assumed that the classification logit is the output of the inner product in Sec.~\ref{sec:theory}.
Here, we show that changing the inner product logit by a (scaled) cosine similarity logit does not alter the major behavior of discriminative learning, and hence they are equivalent in our theoretical consideration. Thus, the theory developed in the inner-product logit also holds in the (scaled) cosine similarity logit.

To observe this, note that the scaled cosine similarity logit is defined as
\begin{equation}
\label{eq:cos_sim}
\phi_k (\mathbf{x}) = \frac{1}{T} \frac{\mathbf{w}_k  }{\lVert \mathbf{w}_k \rVert_2}
\cdot \frac{g(\mathbf{x})}{\lVert g(\mathbf{x}) \rVert_2}  
\end{equation}
where $\mathbf{w}_k$ are class weight vectors (prototypes) of trainable parameters and $g(\mathbf{x}) = \mathbf{U}^T \mathbf{a}^{(L)}$ with a matrix $\mathbf{U}$ of trainable parameters. $T$ is the temperature that modifies the scale of similarity. Without loss of generality, we assume $T=1$. Let $\psi_k(\mathbf{x}) = \mathbf{w}_k \cdot g(\mathbf{x})$ denote the inner-product logit that we originally used. Thus, we have
\begin{equation}
\phi_k(x) = \psi_k(x) (\lVert \mathbf{w}_k \rVert_2 \lVert g(\mathbf{x}) \rVert_2)^{-1}.
\end{equation}

During discriminative learning, the model maximizes
\begin{equation}
(-1)^{1_{y \neq k}} \phi_k(\mathbf{x}) 
=
(-1)^{1_{y \neq k}} \psi_k(\mathbf{x}) (\lVert \mathbf{w}_k \rVert_2 \lVert g(\mathbf{x}) \rVert_2)^{-1}.
\end{equation}
Assuming $\psi_y(\mathbf{x}) =  \mathbf{w}_y \cdot g(\mathbf{x}) >0$ and $\psi_k(\mathbf{x}) = \mathbf{w}_k \cdot g(\mathbf{x}) < 0$, the above maximization is equivalent to minimizing its negative log
\begin{equation}
-\log(
(-1)^{1_{y \neq k}} \phi_k(\mathbf{x}) 
)
=
- \log \left( (-1)^{1_{y \neq k}} \psi_k(\mathbf{x}) \right)
+ \log \left( \lVert \mathbf{w}_k \rVert_2 \lVert g(\mathbf{x}) \rVert_2 \right),
\end{equation}
which can be considered as the constrained minimization of
\begin{equation}
- \log \left( 
(-1)^{1_{y \neq k}} \psi_k(\mathbf{x})
\right)
\equiv
-(-1)^{1_{y = k}} \psi_k(\mathbf{x}) 
\end{equation}
constraint to 
\begin{equation}
\lVert \mathbf{w}_k \rVert_2 \lVert g(\mathbf{x}) \rVert_2 \leq e^{\eta_0} = \eta
\end{equation}
for some $\eta$. 
Thus, optimization of the cosine similarity logit is equivalent to the constrained optimization of the inner product logit.

\begin{prop}
The maximization
\begin{equation}
\max_\phi \quad (-1)^{1_{y \neq k}} \phi_k(\mathbf{x}) 
\end{equation}
is equivalent to 
\begin{equation}
\begin{aligned}
\max_\psi \quad & (-1)^{1_{y = k}} \psi_k(\mathbf{x})  \\
\text{subject to} \quad & \lVert \mathbf{w}_k \rVert_2 \lVert g(\mathbf{x}) \rVert_2 \leq \eta
\end{aligned}
\end{equation}
for some $\eta >0$ if $\psi_y > 0$ and $\psi_k < 0$.
\end{prop}

%\begin{table}[t]
%\centering
%\resizebox{.65\linewidth}{!}{
%\begin{tabular}{l l l}
%\toprule
%Title & ID & OOD  \\ 
%\midrule
%CIFAR10 vs Rest & 6 classes of CIFAR10 & rest of 4 classes in CIFAR10  \\
%CIFAR10 vs CIFAR100 & 6 classes of CIFAR10 & all of CIFAR100 \\
%CIFAR10 vs LSUN & 6 classes of CIFAR10 & all of LSUN \\
%CIFAR10 vs SVHN & 6 classes of CIFAR10 & 6 classes of SVHN \\
%TinyImageNet vs Rest & 20 classes of TinyImageNet & rest of 80 classes in TinyImageNet \\
%TinyImageNet vs LSUN & 20 classes of TinyImageNet & all of LSUN \\
%TinyImageNet vs SVHN & 20 classes of TinyImageNet & 6 classes of SVHN \\
%\bottomrule
%\end{tabular}
%}
%\caption{
%The ID and OOD configuration of the datasets used in the empirical analysis (Sec.~\ref{sec:emp_analysis}). For the selection of particular classes in a dataset, we follow the OSR protocol given by \cite{neal2018open,vaze2021open}
%}
%\label{table:supp_dataset_exp_anal}
%\end{table}

\subsection{Supplementary to empirical validation of hidden classifier}
\label{asec:add_hidden_classifier}

Here, we provide a detailed description of the experiments conducted to validate the theoretical analysis presented in Sec.~\ref{sec:theory}.

\subsubsection{On MLP}
\label{asec:add_hidden_classifier_mlp}

\paragraph{Setup}
We train an MLP with 5 hidden layers. The hidden layer dimension is fixed to 512, and likewise for the embedding layer dimension. The embedding is normalized, and the cosine similarity logit is divided by a temperature of 0.1. The model is trained by AdamW for 200 epochs with batch size 256. The learning rate decays from 0.001 to 0 by the cosine scheduler. Other setups follow the default setting in PyTorch.

\begin{figure*}[t]
\centering
% CIFAR-10, ReLU
\begin{subfigure}{0.495\linewidth}
\includegraphics[width=1.0\linewidth]{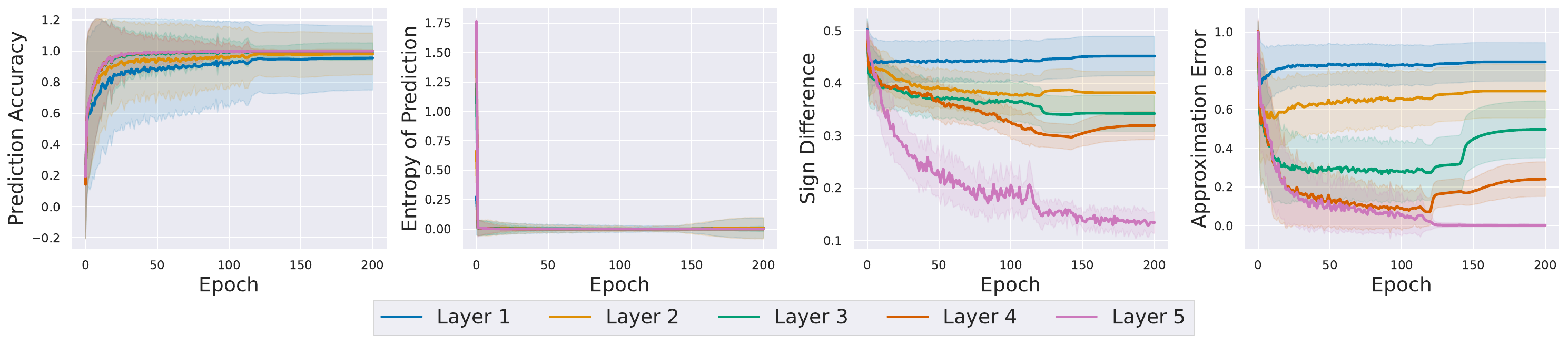}
\caption{CIFAR-10, ReLU, w/o bias, post-activation}
\end{subfigure}
\hfill
\begin{subfigure}{0.495\linewidth}
\includegraphics[width=1.0\linewidth]{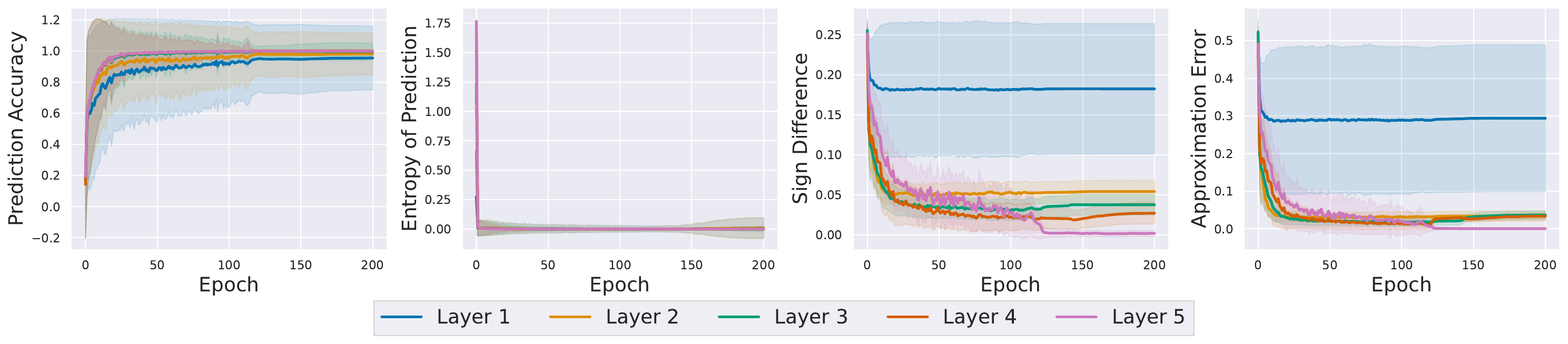}
\caption{CIFAR-10, ReLU, w/o bias, pre-activation}
\end{subfigure}
\begin{subfigure}{0.495\linewidth}
\includegraphics[width=1.0\linewidth]{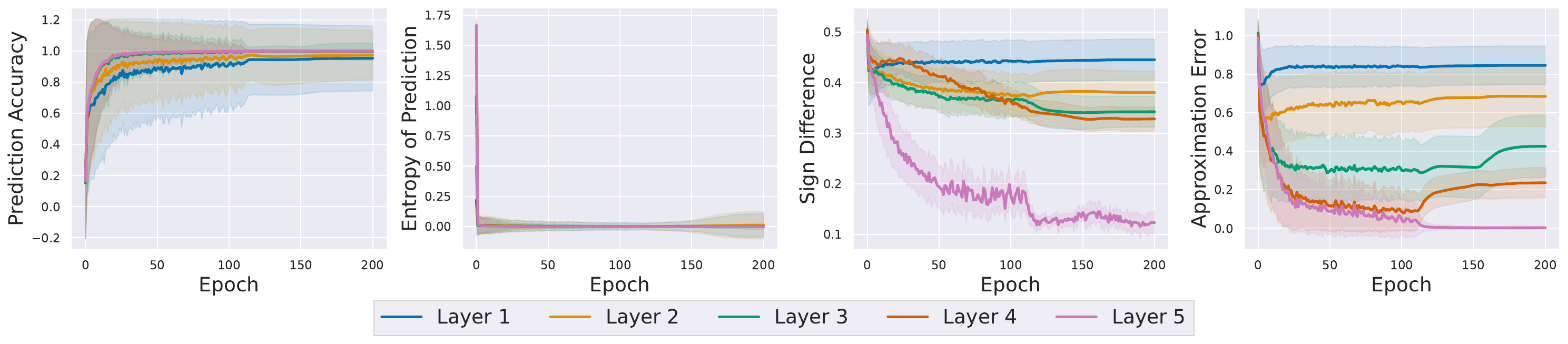}
\caption{CIFAR-10, ReLU, w/ bias, post-activation}
\end{subfigure}
\hfill
\begin{subfigure}{0.495\linewidth}
\includegraphics[width=1.0\linewidth]{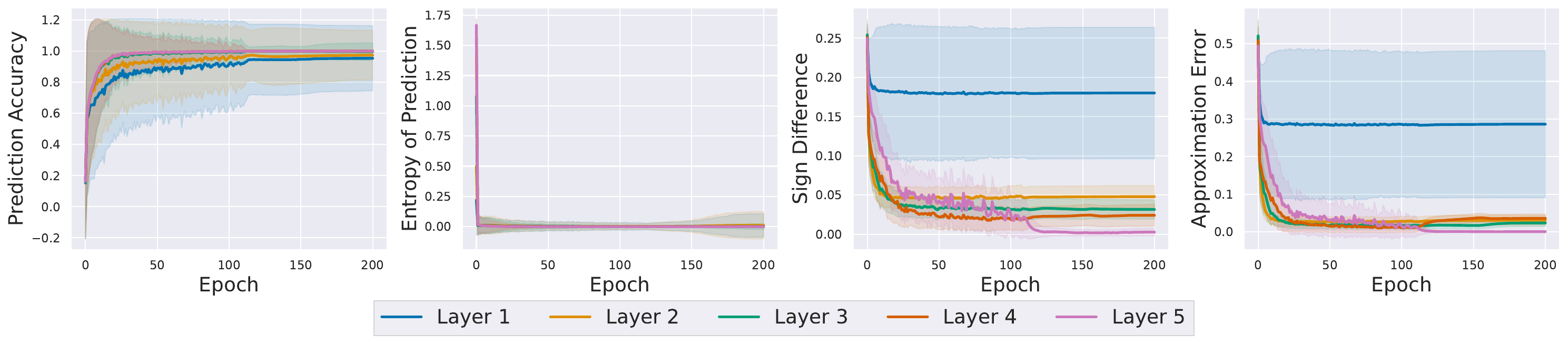}
\caption{CIFAR-10, ReLU, w/ bias, pre-activation}
\end{subfigure}
% CIFAR-10, Leaky ReLU
\begin{subfigure}{0.495\linewidth}
\includegraphics[width=1.0\linewidth]{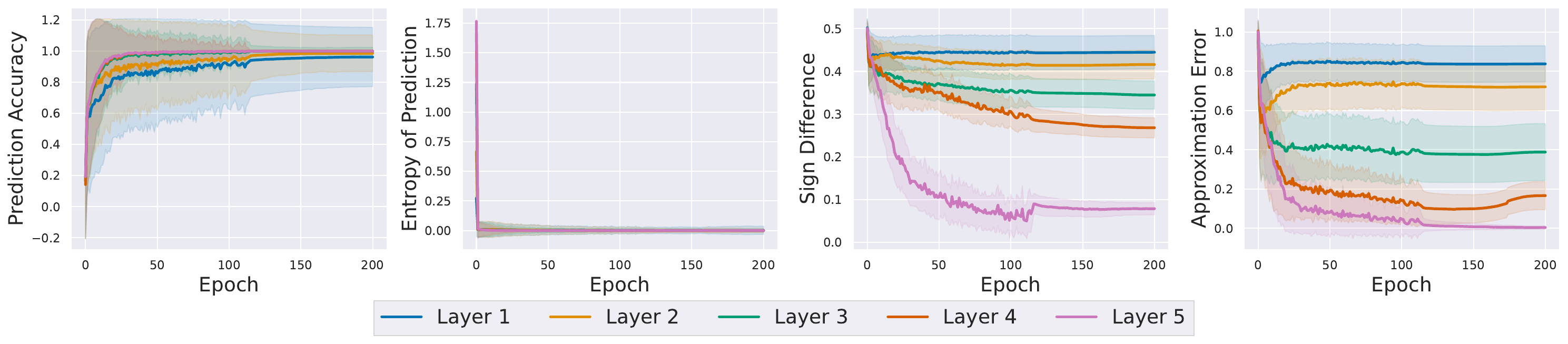}
\caption{CIFAR-10, Leaky ReLU, w/o bias, post-activation}
\end{subfigure}
\hfill
\begin{subfigure}{0.495\linewidth}
\includegraphics[width=1.0\linewidth]{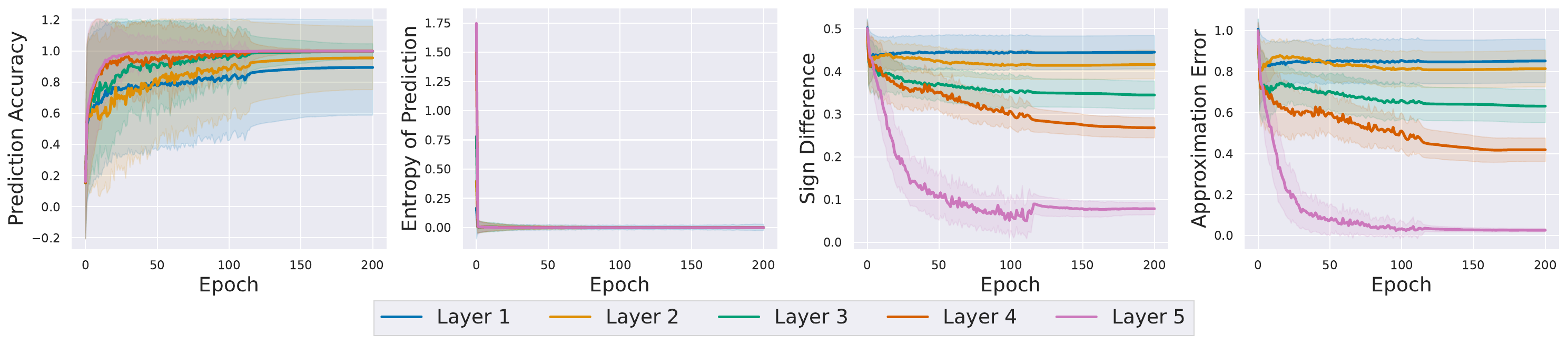}
\caption{CIFAR-10, Leaky ReLU, w/o bias, pre-activation}
\end{subfigure}
\begin{subfigure}{0.495\linewidth}
\includegraphics[width=1.0\linewidth]{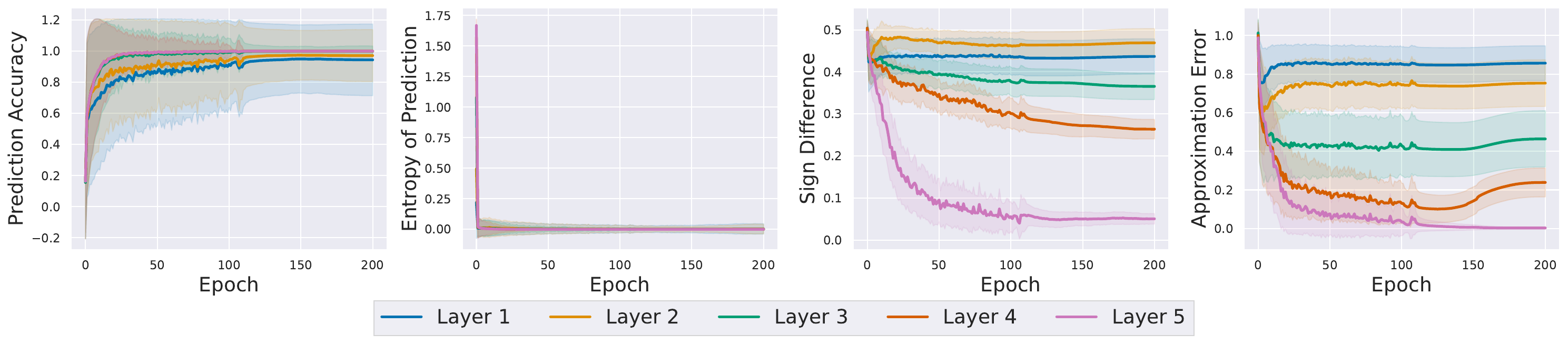}
\caption{CIFAR-10, Leaky ReLU, w/ bias, post-activation}
\end{subfigure}
\hfill
\begin{subfigure}{0.495\linewidth}
\includegraphics[width=1.0\linewidth]{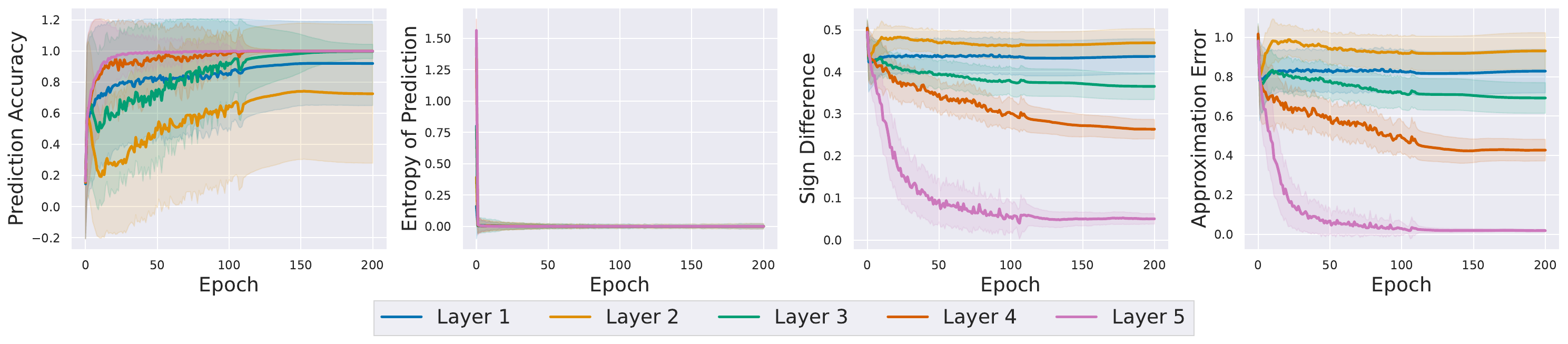}
\caption{CIFAR-10, Leaky ReLU, w/ bias, pre-activation}
\end{subfigure}
% CIFAR-10, GeLU
\begin{subfigure}{0.495\linewidth}
\includegraphics[width=1.0\linewidth]{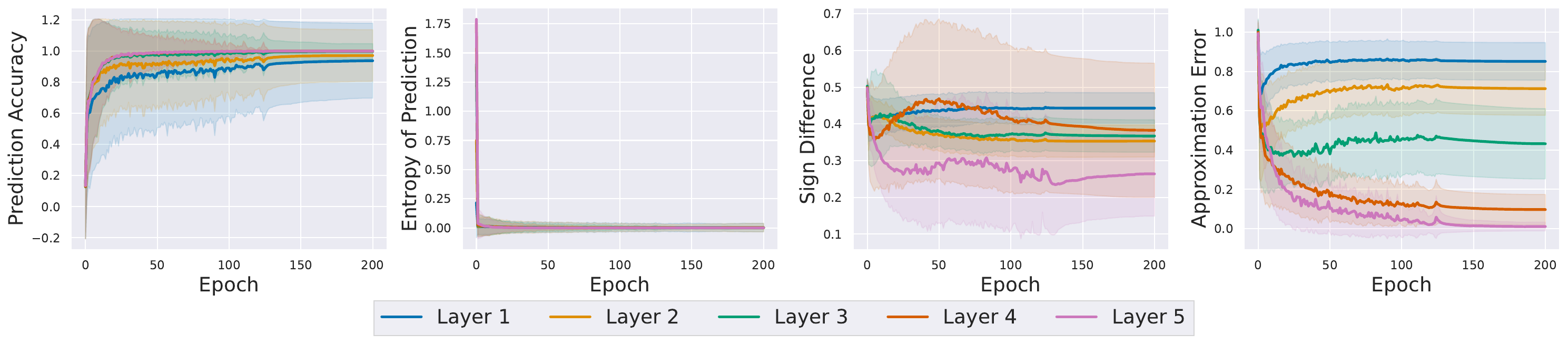}
\caption{CIFAR-10, GeLU, w/o bias, post-activation}
\end{subfigure}
\hfill
\begin{subfigure}{0.495\linewidth}
\includegraphics[width=1.0\linewidth]{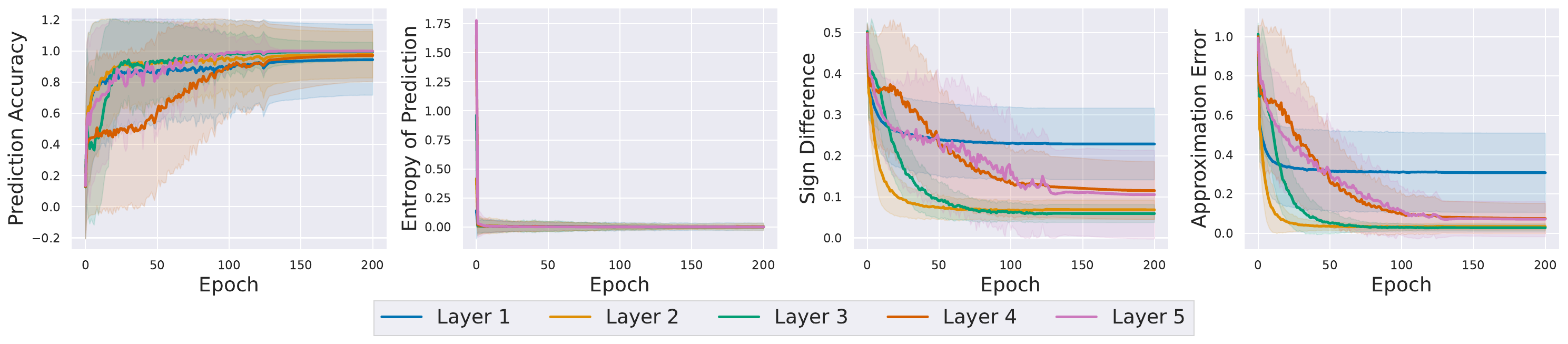}
\caption{CIFAR-10, GeLU, w/o bias, pre-activation}
\end{subfigure}
\begin{subfigure}{0.495\linewidth}
\includegraphics[width=1.0\linewidth]{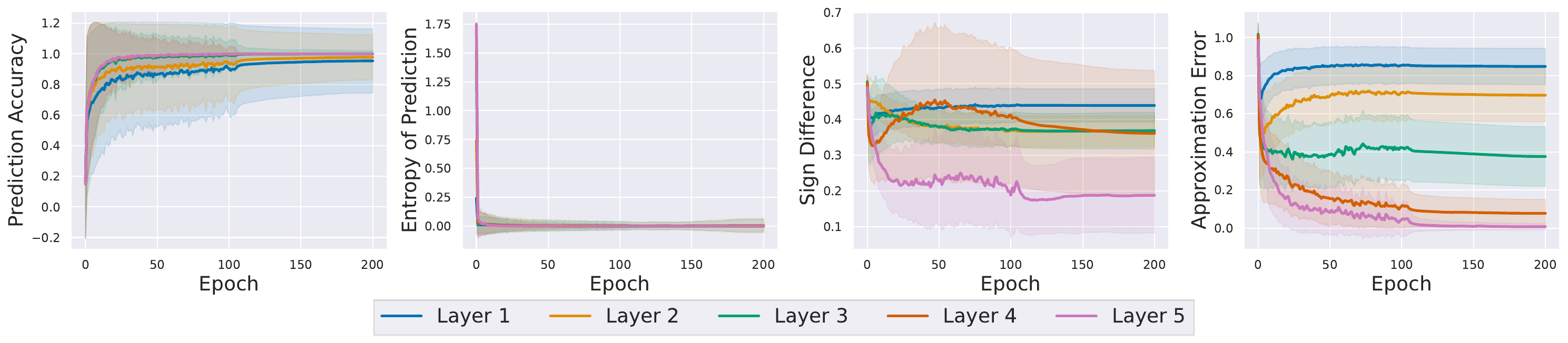}
\caption{CIFAR-10, GeLU, w/ bias, post-activation}
\end{subfigure}
\hfill
\begin{subfigure}{0.495\linewidth}
\includegraphics[width=1.0\linewidth]{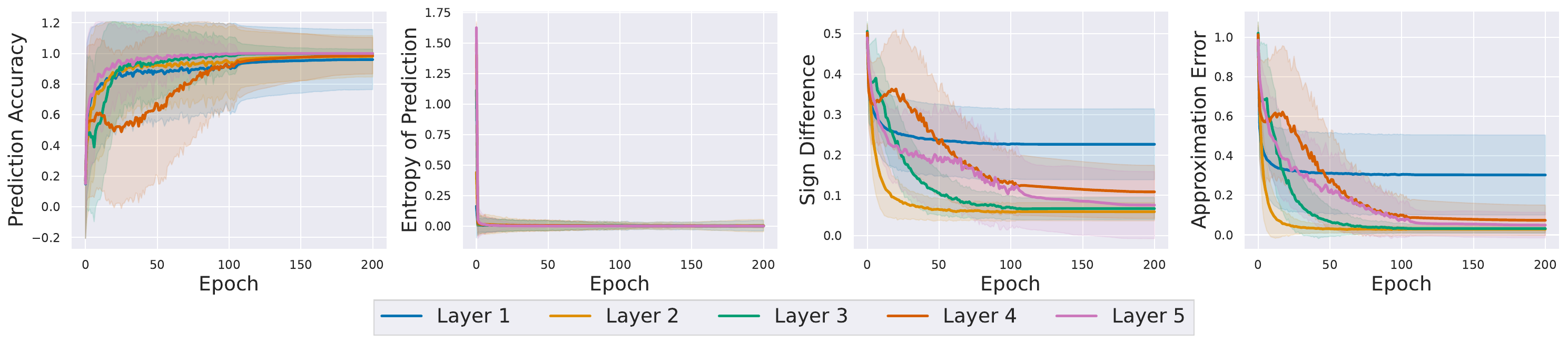}
\caption{CIFAR-10, GeLU, w/ bias, pre-activation}
\end{subfigure}
\caption{Results of hidden classifiers with different activation functions (ReLU, Leaky ReLU, and GeLU) on CIFAR-10.}
\label{fig:supp_hidden_cifar10}
\end{figure*}

\begin{figure*}[t]
\centering
% SVHN, ReLU
\begin{subfigure}{0.495\linewidth}
\includegraphics[width=1.0\linewidth]{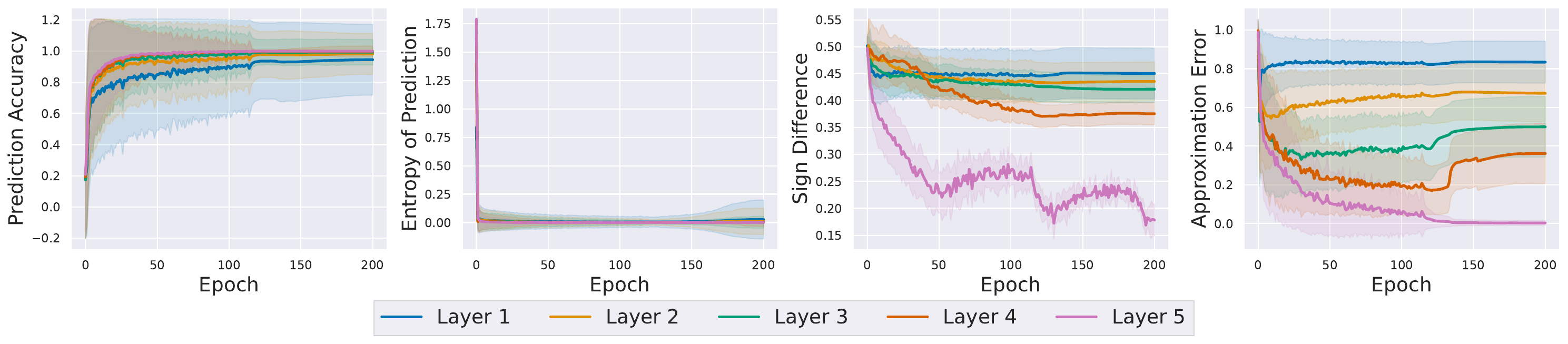}
\caption{SVHN, ReLU, w/o bias, post-activation}
\end{subfigure}
\hfill
\begin{subfigure}{0.495\linewidth}
\includegraphics[width=1.0\linewidth]{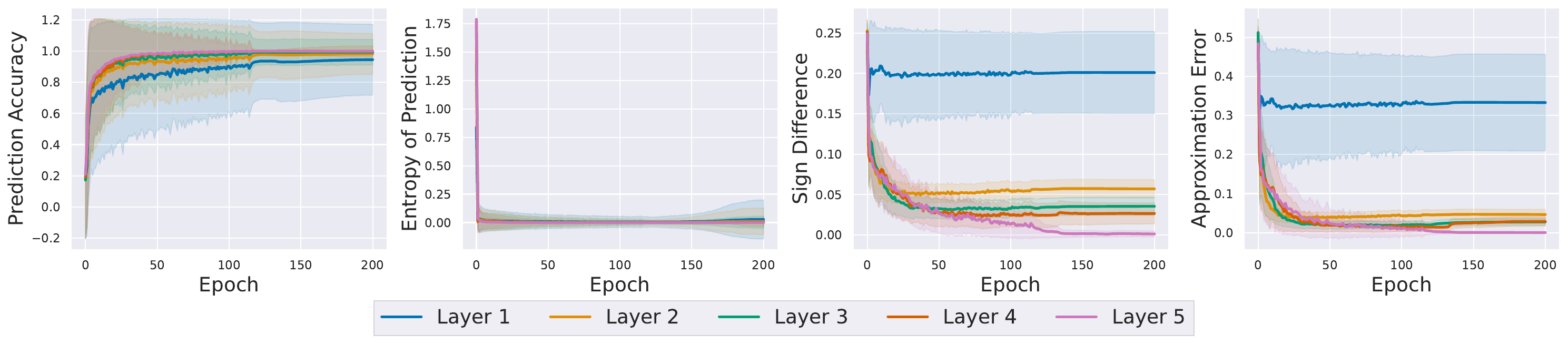}
\caption{SVHN, ReLU, w/o bias, pre-activation}
\end{subfigure}
\begin{subfigure}{0.495\linewidth}
\includegraphics[width=1.0\linewidth]{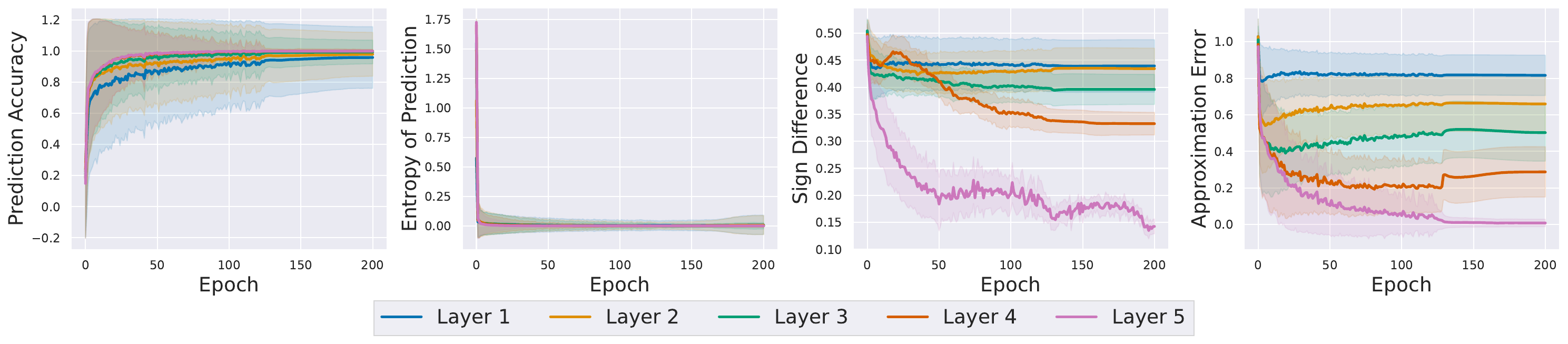}
\caption{SVHN, ReLU, w/ bias, post-activation}
\end{subfigure}
\hfill
\begin{subfigure}{0.495\linewidth}
\includegraphics[width=1.0\linewidth]{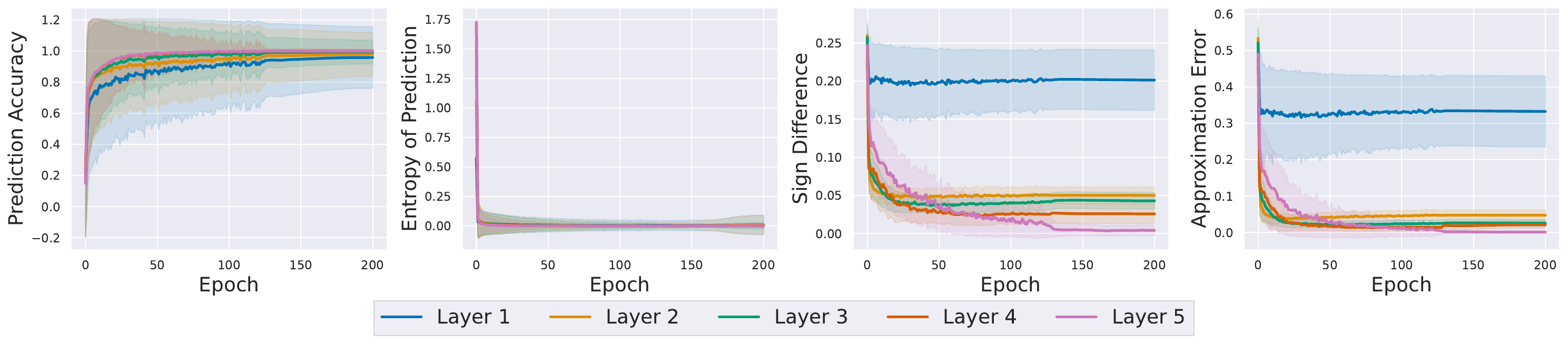}
\caption{SVHN, ReLU, w/ bias, pre-activation}
\end{subfigure}
% SVHN, Leaky ReLU
\begin{subfigure}{0.495\linewidth}
\includegraphics[width=1.0\linewidth]{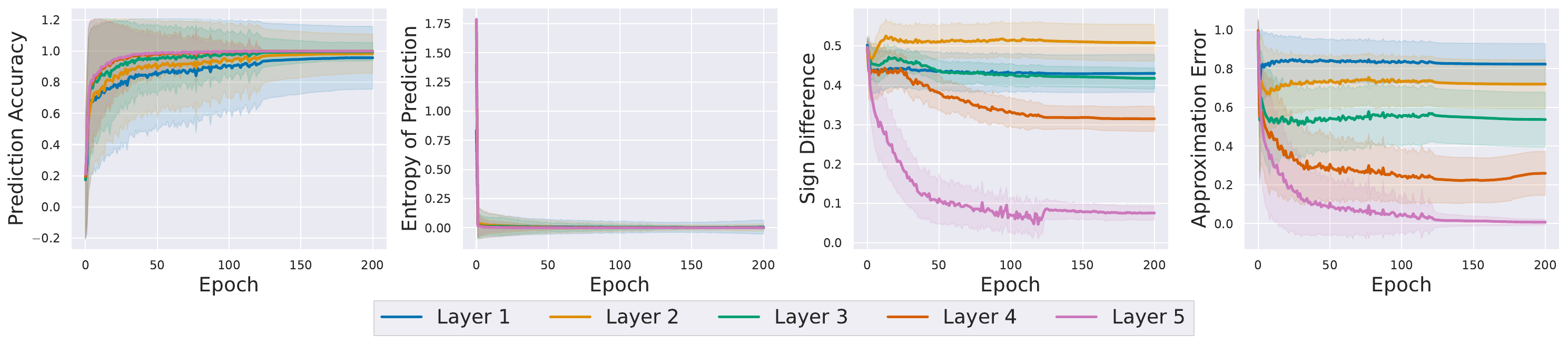}
\caption{SVHN, Leaky ReLU, w/o bias, post-activation}
\end{subfigure}
\hfill
\begin{subfigure}{0.495\linewidth}
\includegraphics[width=1.0\linewidth]{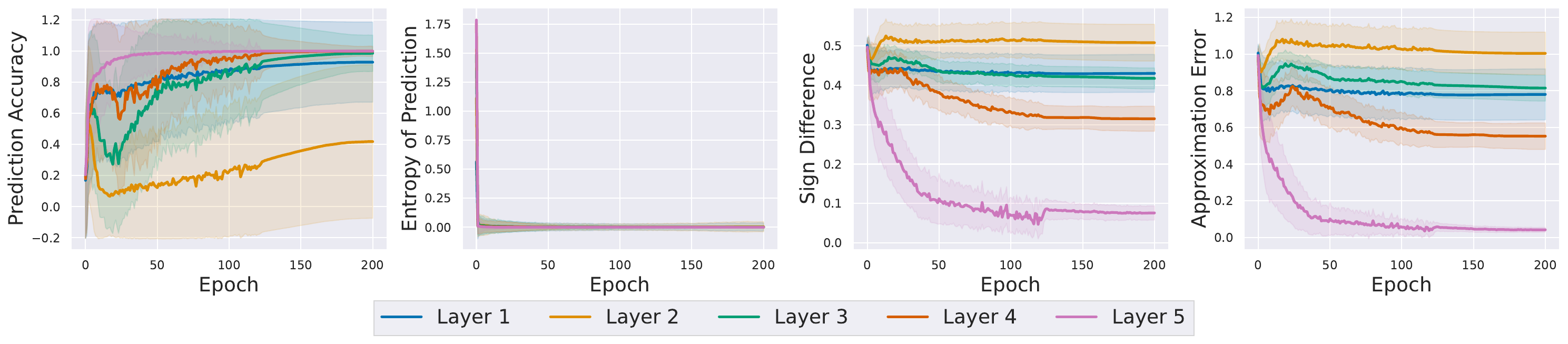}
\caption{SVHN, Leaky ReLU, w/o bias, pre-activation}
\end{subfigure}
\begin{subfigure}{0.495\linewidth}
\includegraphics[width=1.0\linewidth]{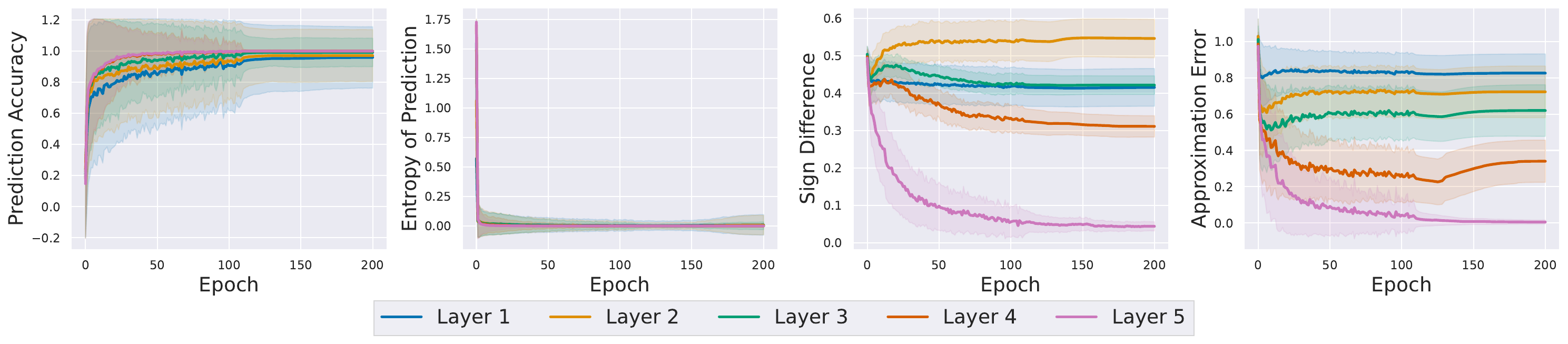}
\caption{SVHN, Leaky ReLU, w/ bias, post-activation}
\end{subfigure}
\hfill
\begin{subfigure}{0.495\linewidth}
\includegraphics[width=1.0\linewidth]{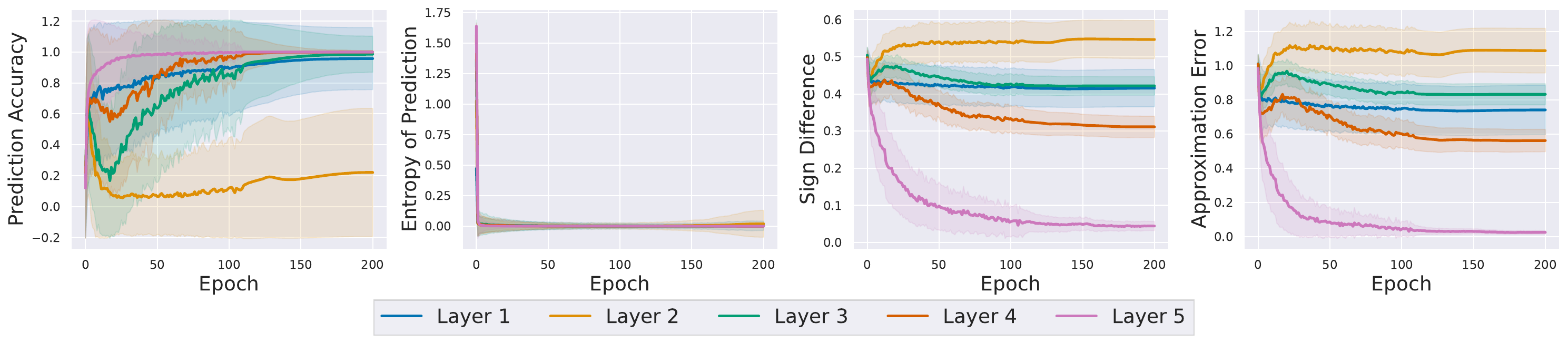}
\caption{SVHN, Leaky ReLU, w/ bias, pre-activation}
\end{subfigure}
% SVHN, GeLU
\begin{subfigure}{0.495\linewidth}
\includegraphics[width=1.0\linewidth]{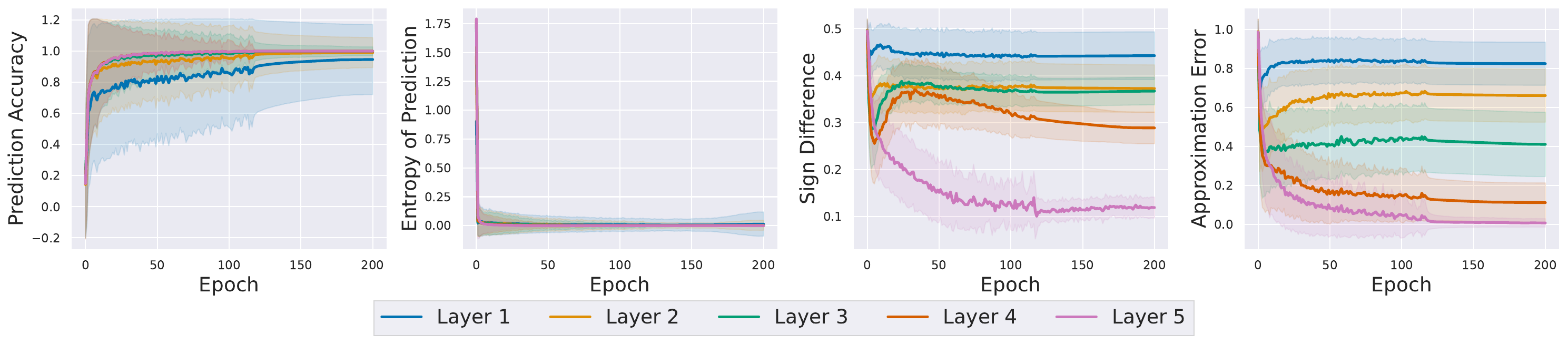}
\caption{SVHN, GeLU, w/o bias, post-activation}
\end{subfigure}
\hfill
\begin{subfigure}{0.495\linewidth}
\includegraphics[width=1.0\linewidth]{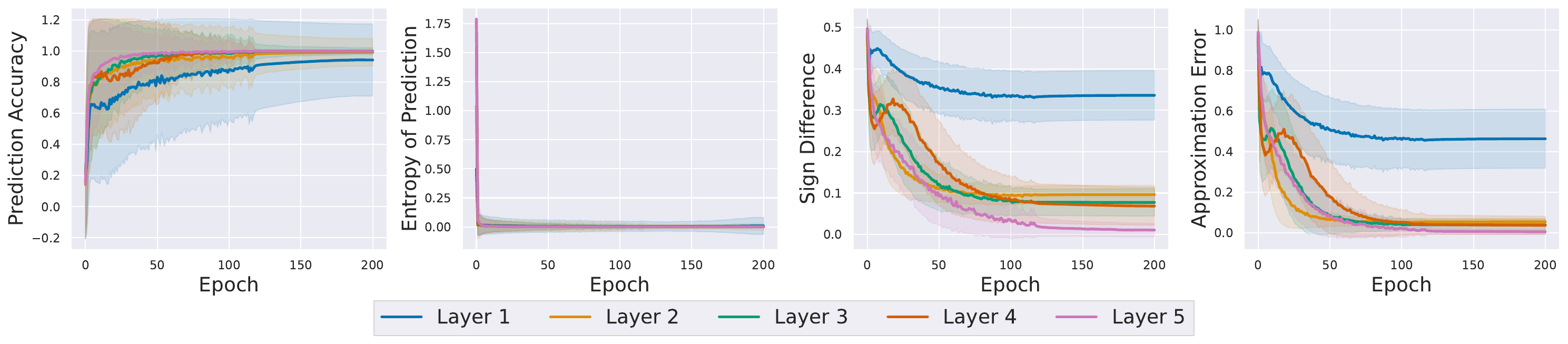}
\caption{SVHN, GeLU, w/o bias, pre-activation}
\end{subfigure}
\begin{subfigure}{0.495\linewidth}
\includegraphics[width=1.0\linewidth]{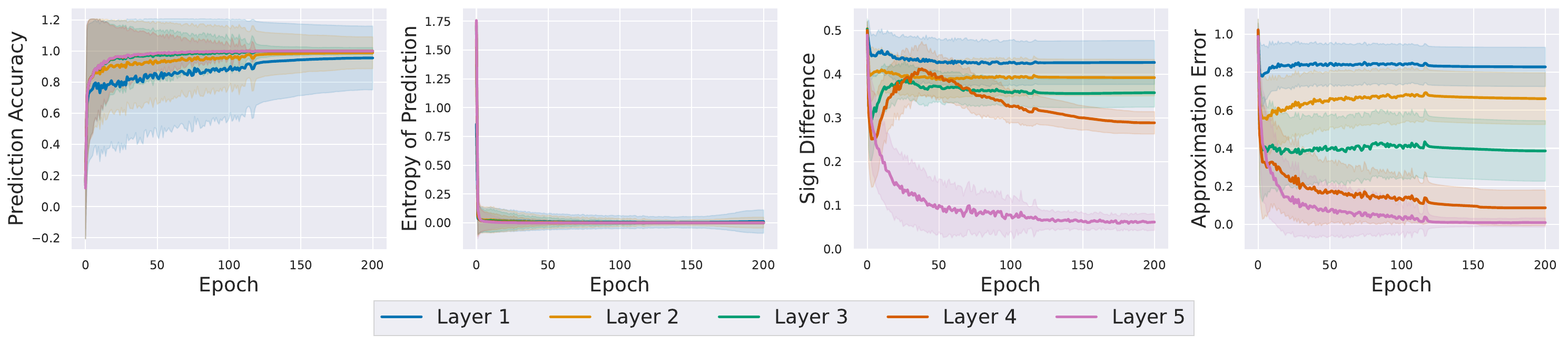}
\caption{SVHN, GeLU, w/ bias, post-activation}
\end{subfigure}
\hfill
\begin{subfigure}{0.495\linewidth}
\includegraphics[width=1.0\linewidth]{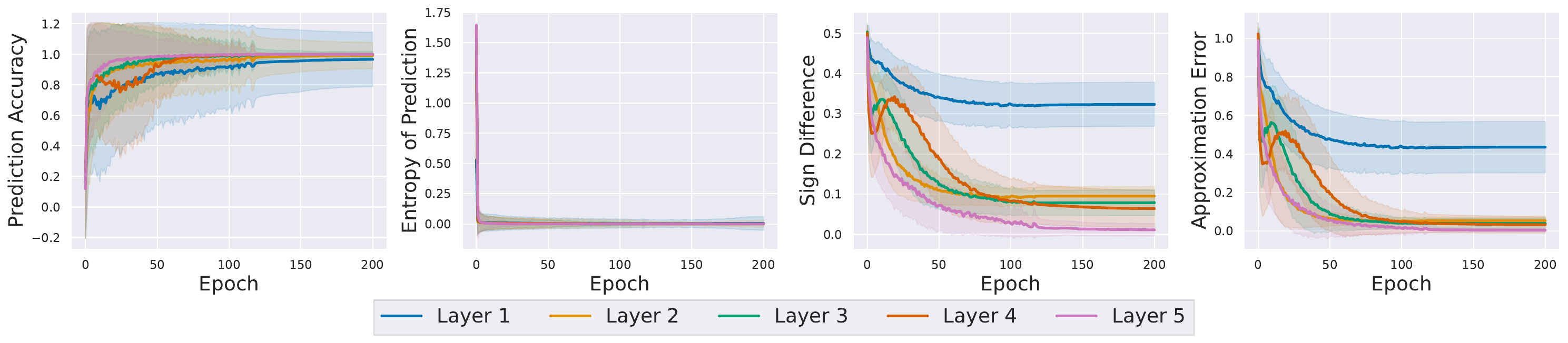}
\caption{SVHN, GeLU, w/ bias, pre-activation}
\end{subfigure}
\caption{Results of hidden classifiers with different activation functions (ReLU, Leaky ReLU, and GeLU) on SVHN.}
\label{fig:supp_hidden_svhn}
\end{figure*}

\begin{figure*}[t]
\centering
% MNIST, ReLU
\begin{subfigure}{0.495\linewidth}
\includegraphics[width=1.0\linewidth]{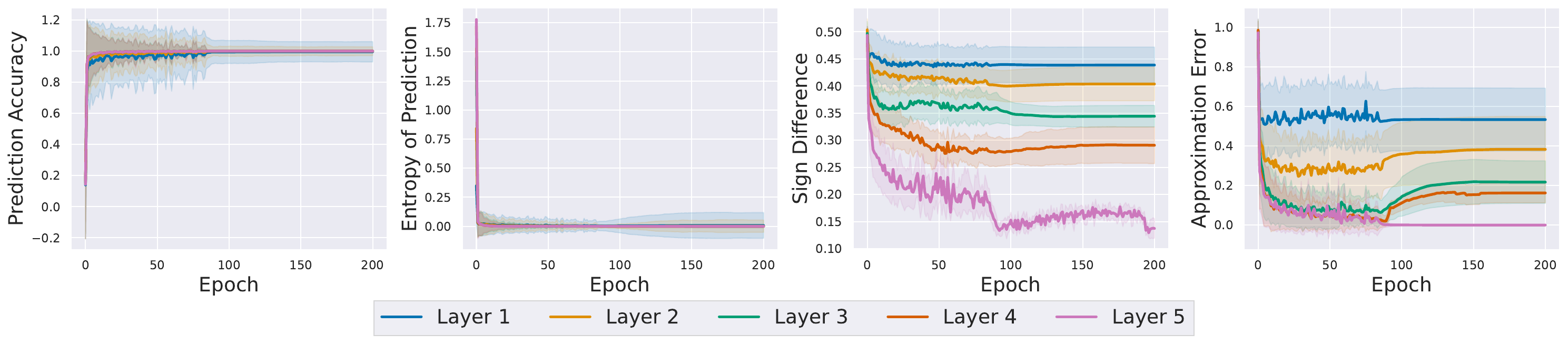}
\caption{MNIST, ReLU, w/o bias, post-activation}
\end{subfigure}
\hfill
\begin{subfigure}{0.495\linewidth}
\includegraphics[width=1.0\linewidth]{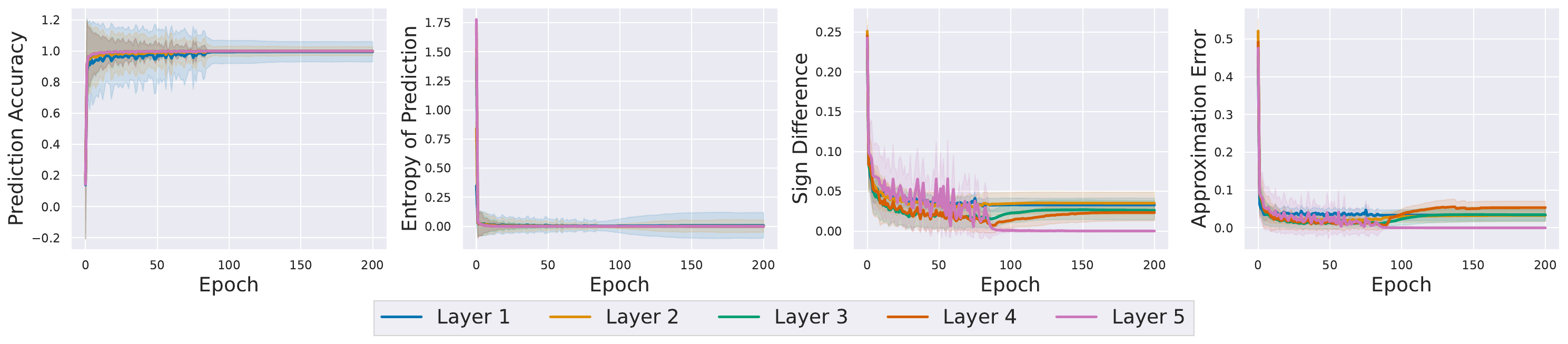}
\caption{MNIST, ReLU, w/o bias, pre-activation}
\end{subfigure}
\begin{subfigure}{0.495\linewidth}
\includegraphics[width=1.0\linewidth]{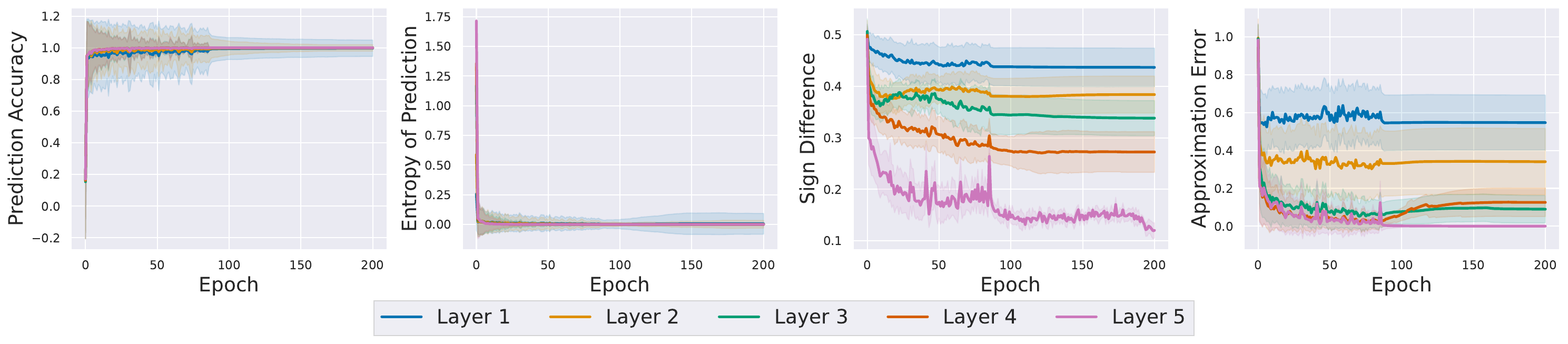}
\caption{MNIST, ReLU, w/ bias, post-activation}
\end{subfigure}
\hfill
\begin{subfigure}{0.495\linewidth}
\includegraphics[width=1.0\linewidth]{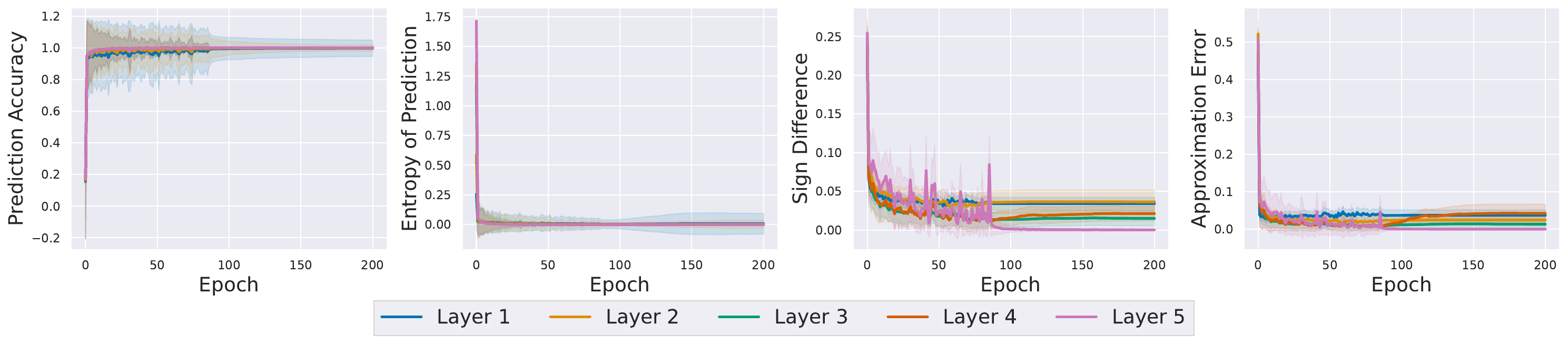}
\caption{MNIST, ReLU, w/ bias, pre-activation}
\end{subfigure}
% SVHN, Leaky ReLU
\begin{subfigure}{0.495\linewidth}
\includegraphics[width=1.0\linewidth]{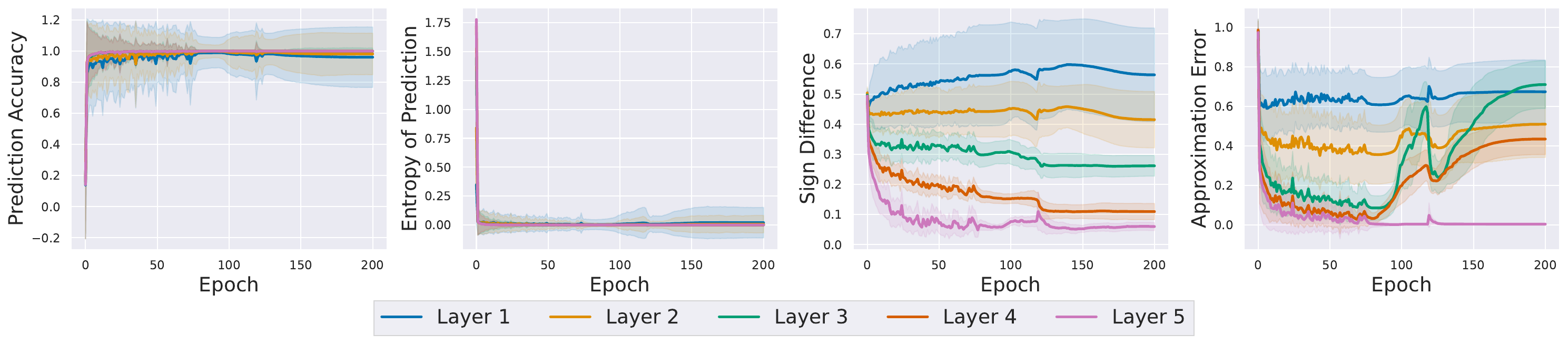}
\caption{MNIST, Leaky ReLU, w/o bias, post-activation}
\end{subfigure}
\hfill
\begin{subfigure}{0.495\linewidth}
\includegraphics[width=1.0\linewidth]{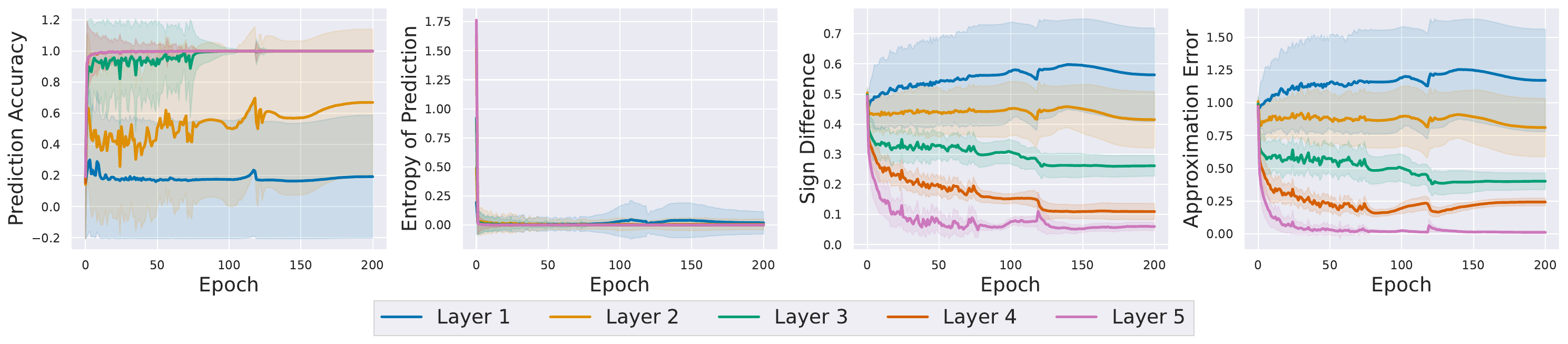}
\caption{MNIST, Leaky ReLU, w/o bias, pre-activation}
\end{subfigure}
\begin{subfigure}{0.495\linewidth}
\includegraphics[width=1.0\linewidth]{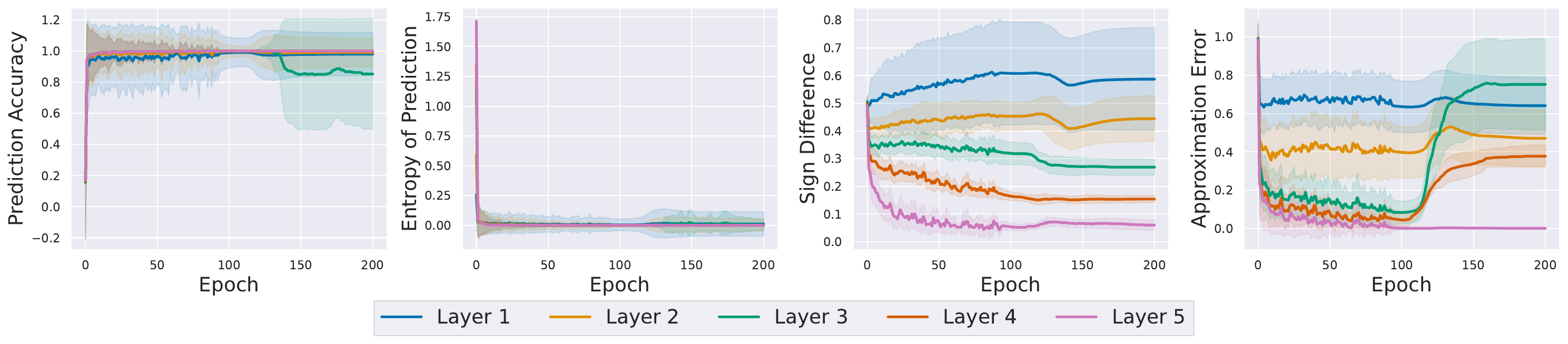}
\caption{MNIST, Leaky ReLU, w/ bias, post-activation}
\end{subfigure}
\hfill
\begin{subfigure}{0.495\linewidth}
\includegraphics[width=1.0\linewidth]{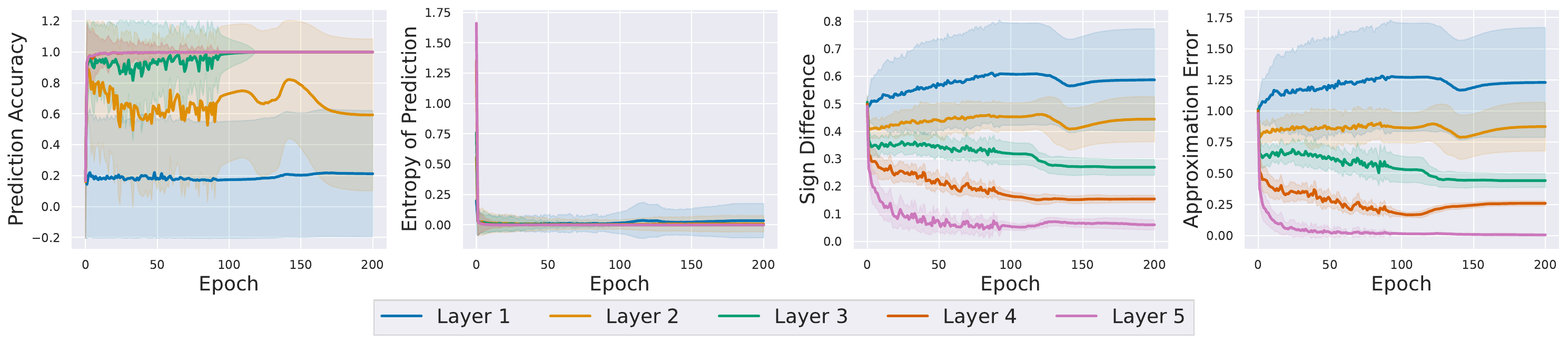}
\caption{MNIST, Leaky ReLU, w/ bias, pre-activation}
\end{subfigure}
% SVHN, GeLU
\begin{subfigure}{0.495\linewidth}
\includegraphics[width=1.0\linewidth]{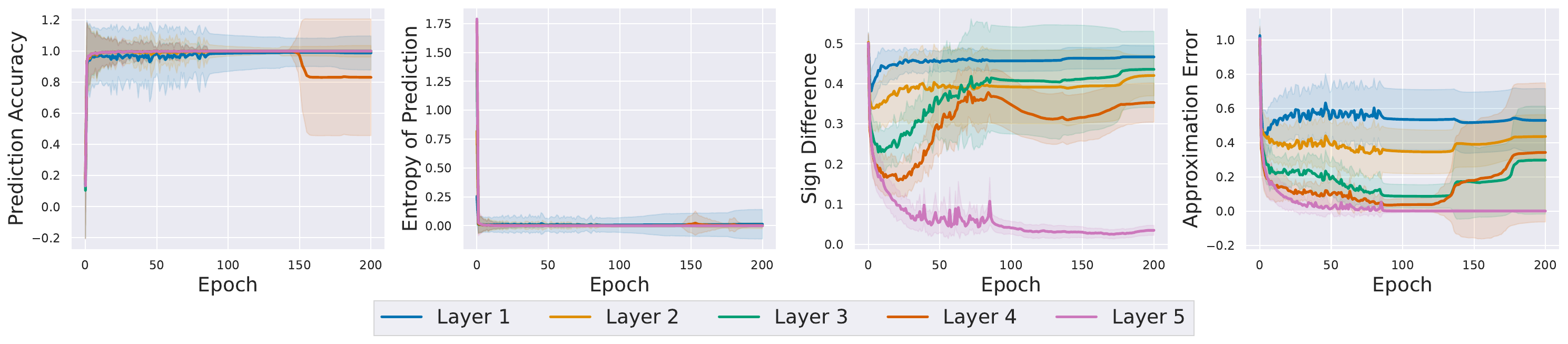}
\caption{MNIST, GeLU, w/o bias, post-activation}
\end{subfigure}
\hfill
\begin{subfigure}{0.495\linewidth}
\includegraphics[width=1.0\linewidth]{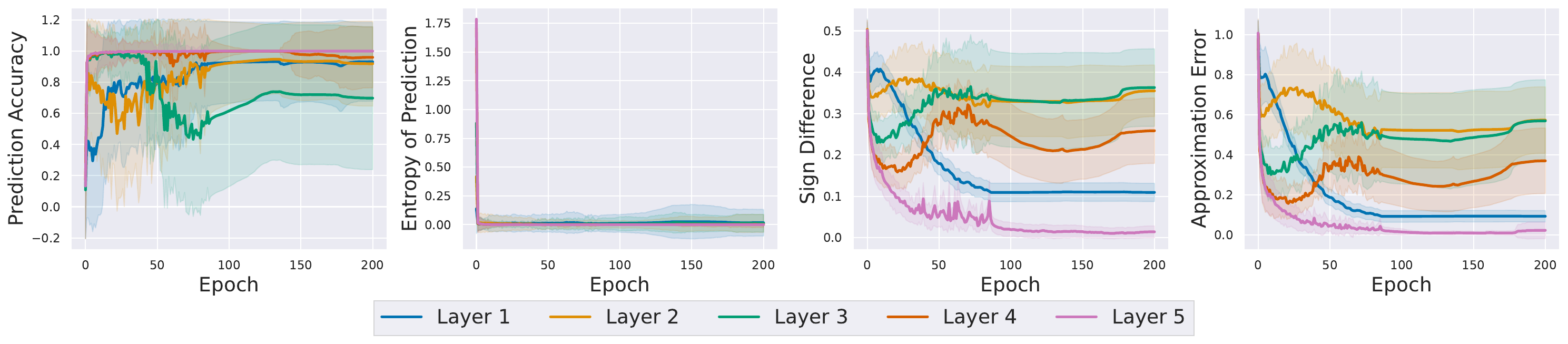}
\caption{MNIST, GeLU, w/o bias, pre-activation}
\end{subfigure}
\begin{subfigure}{0.495\linewidth}
\includegraphics[width=1.0\linewidth]{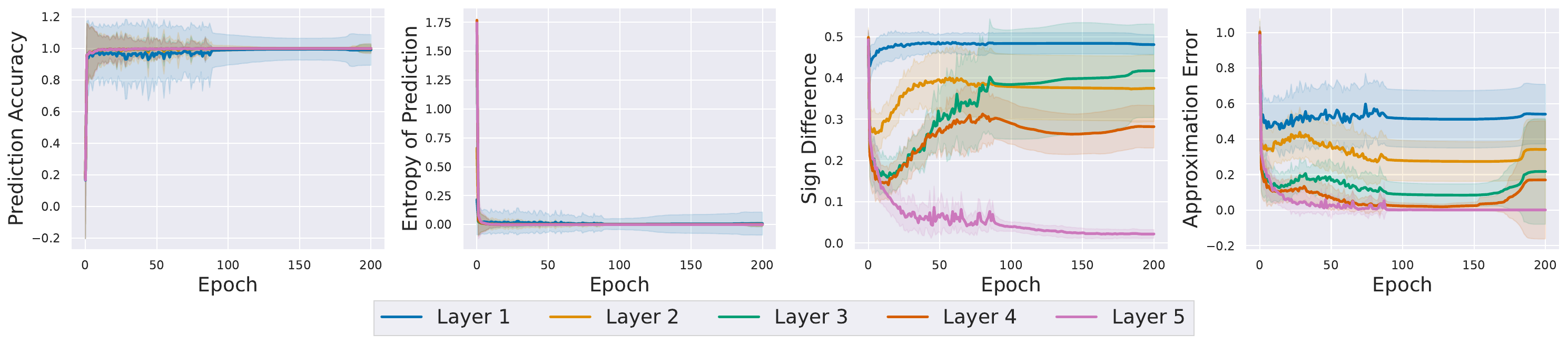}
\caption{MNIST, GeLU, w/ bias, post-activation}
\end{subfigure}
\hfill
\begin{subfigure}{0.495\linewidth}
\includegraphics[width=1.0\linewidth]{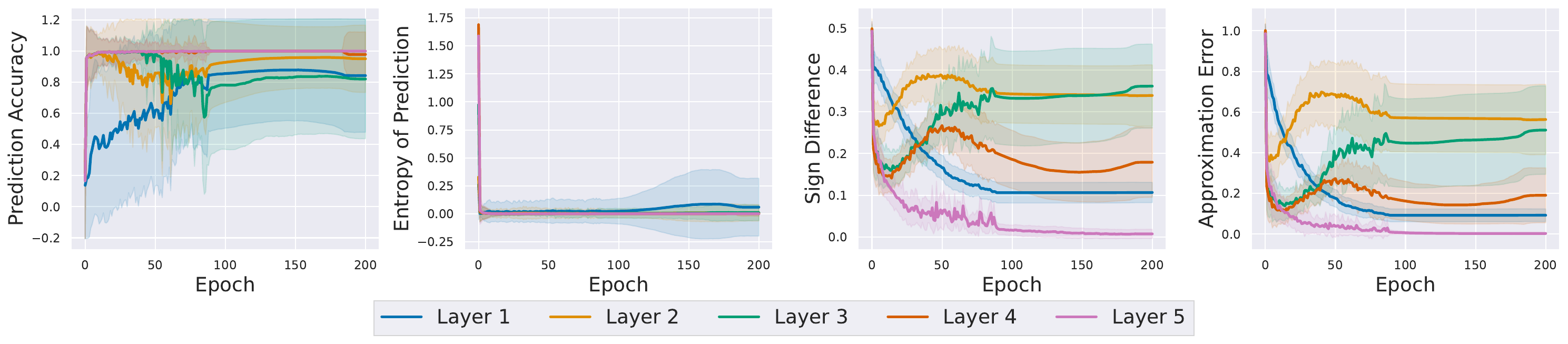}
\caption{MNIST, GeLU, w/ bias, pre-activation}
\end{subfigure}
\caption{Results of hidden classifiers with different activation functions (ReLU, Leaky ReLU, and GeLU) on MNIST.}
\label{fig:supp_hidden_mnist}
\end{figure*}

\paragraph{Results}
The results are given in Fig.~\ref{fig:supp_hidden_cifar10}, \ref{fig:supp_hidden_svhn}, and \ref{fig:supp_hidden_mnist}. They have similar trends that we expected and thus verify our theoretical claims.

\subsubsection{On Convolutional Network}

\paragraph{Setup}
The experiment setup is given as in Sec.~\ref{asec:setup_general_model}.

\begin{figure*}[t]
\centering
% CIFAR-10, ReLU
\begin{subfigure}{0.95\linewidth}
\includegraphics[width=1.0\linewidth]{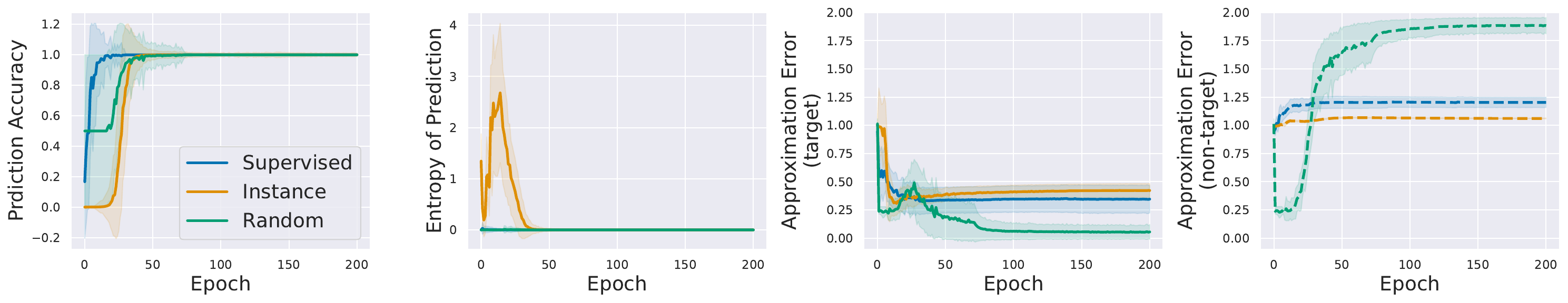}
\caption{post-activation}
\end{subfigure}
\begin{subfigure}{0.95\linewidth}
\includegraphics[width=1.0\linewidth]{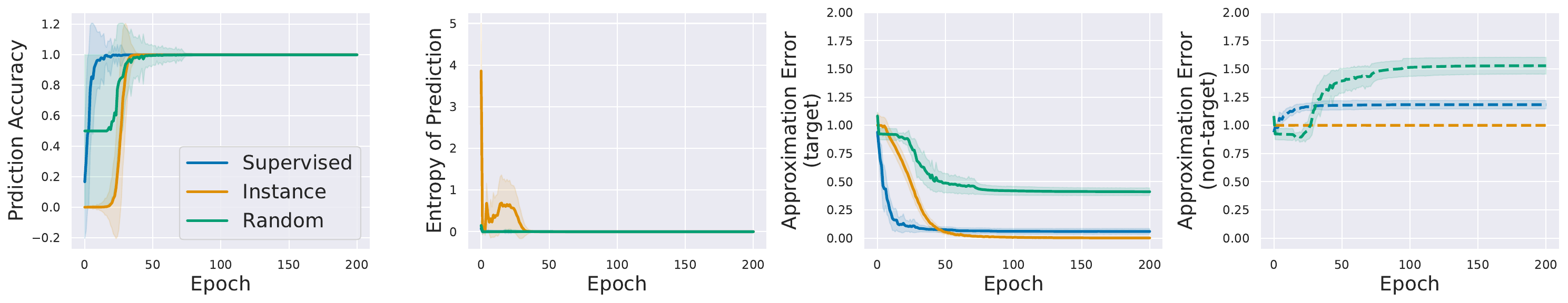}
\caption{pre-activation}
\end{subfigure}
\caption{Results of hidden classifiers of ResNet-18 with different class labeling schemes on CIFAR-10. The approximation error on the target unit measures the normalized error $(\lVert \mathbf{a} \rVert_1 - \overline{\psi}_y(\mathbf{x})) / \lVert \mathbf{a} \rVert_1$, while the approximation error on the non-target unit is the average of $(\lVert \mathbf{a} \rVert_1 - \overline{\psi}_k(\mathbf{x})) / \lVert \mathbf{a} \rVert_1$ with respect to $k \neq y$. In the case of post-activation, the vector $\mathbf{a}$ is $\mathbf{a} = \mathbf{a}^{(L)}$. In the case of pre-activation, the vector $\mathbf{a}$ is $\mathbf{a} = \mathbf{z}^{(l)}$ }
\label{fig:supp_hidden_resnet18}
\end{figure*}

\paragraph{Results}
In the cases of both instance discrimination (I), supervised learning (S), and random binary label discrimination (R), the hidden classifier of the last hidden layer in ResNet-18 is trained to be discriminative (Fig.~\ref{fig:supp_hidden_resnet18}).

\section{Supplementary to the Analysis of Feature Norm's Class Agnosticity}
\label{asec:setup_general_model}

\noindent
\textbf{Setup.}
We train a ResNet-18 on CIFAR-10. We add an MLP projection head as in MoCo-v2 \cite{chen2020improved}. The embedding is normalized, and the cosine similarity logit is divided by a temperature of 0.1. The model is trained for 200 epochs and batch size 256 with the SGD optimizer, cosine learning rate (0.06 to 0), and momentum 0.9.
Each model is trained in a different manner based on a different class labeling scheme:
\begin{itemize}
\item \textbf{S}: The class labels $y_i$ are supervised labels (\eg plane, dog, cat, ...). No data augmentation is applied.
\item \textbf{I}: The class labels $y_i$ are instance labels $y_i = i$. No data augmentation is applied such that each instance class has only one intra-class sample.
\item \textbf{Is}: The class labels $y_i$ are instance labels $y_i = i$. Data augmentation is applied such that each instance class has multiple intra-class samples.
\item \textbf{R}: The class labels $y_i$ are labeled randomly by a binary number $y_i \in \{0,1\}$.
\item \textbf{O}: The class labels $y_i$ are labeled with a single label $y_i = 0$ such that every sample is in the same class.
\end{itemize}
Other setups follow the default setting in PyTorch.

\paragraph{Full results on the impact of inter/intra-class learning}

\begin{figure*}[t]
\centering
\begin{subfigure}{0.3\linewidth}
\includegraphics[width=.995\linewidth]{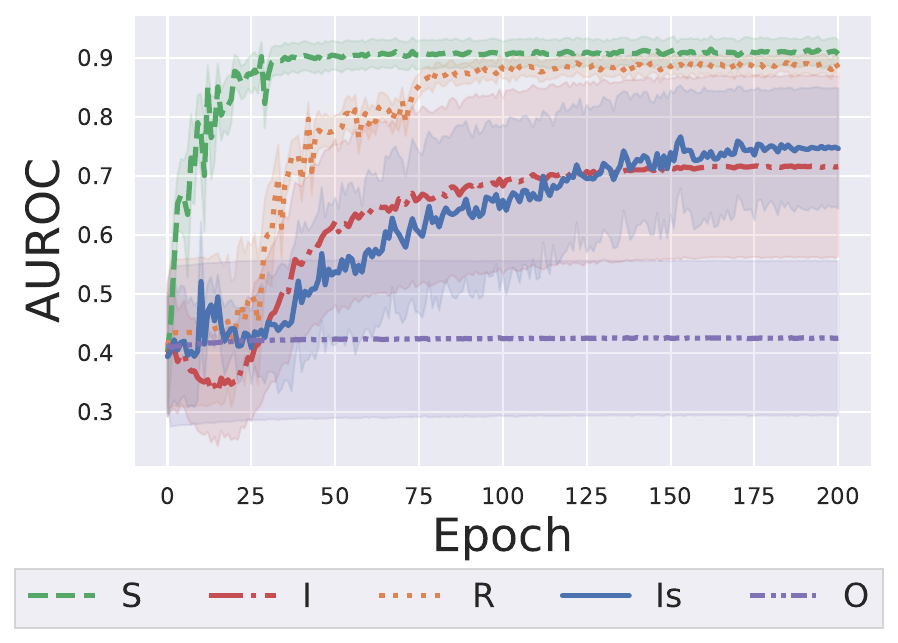}
\caption{}
\end{subfigure}
\begin{subfigure}{0.3\linewidth}
\includegraphics[width=.995\linewidth]{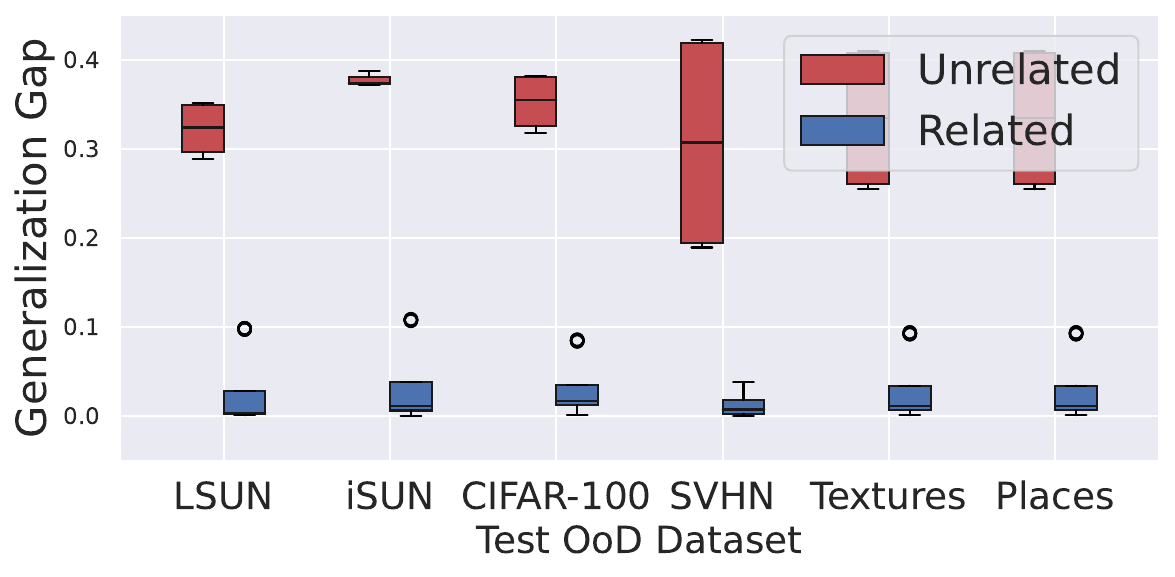}
\caption{}
\end{subfigure}
\caption{
(a) The detection performance of NAN versus the learning epoch across different types of training schemes
(b) The generalization gap of NAN based on the intra-class semantics.
}
\label{fig:supp_interintra}
\end{figure*}

The additional results on NAN is given in Fig.~\ref{fig:supp_interintra}, which NAN exhibits the same trend of memorization and generalization as the conventional feature norm.

\paragraph{Full results on the relation to entropy}

\begin{figure*}[t]
\centering
\includegraphics[width=.995\linewidth]{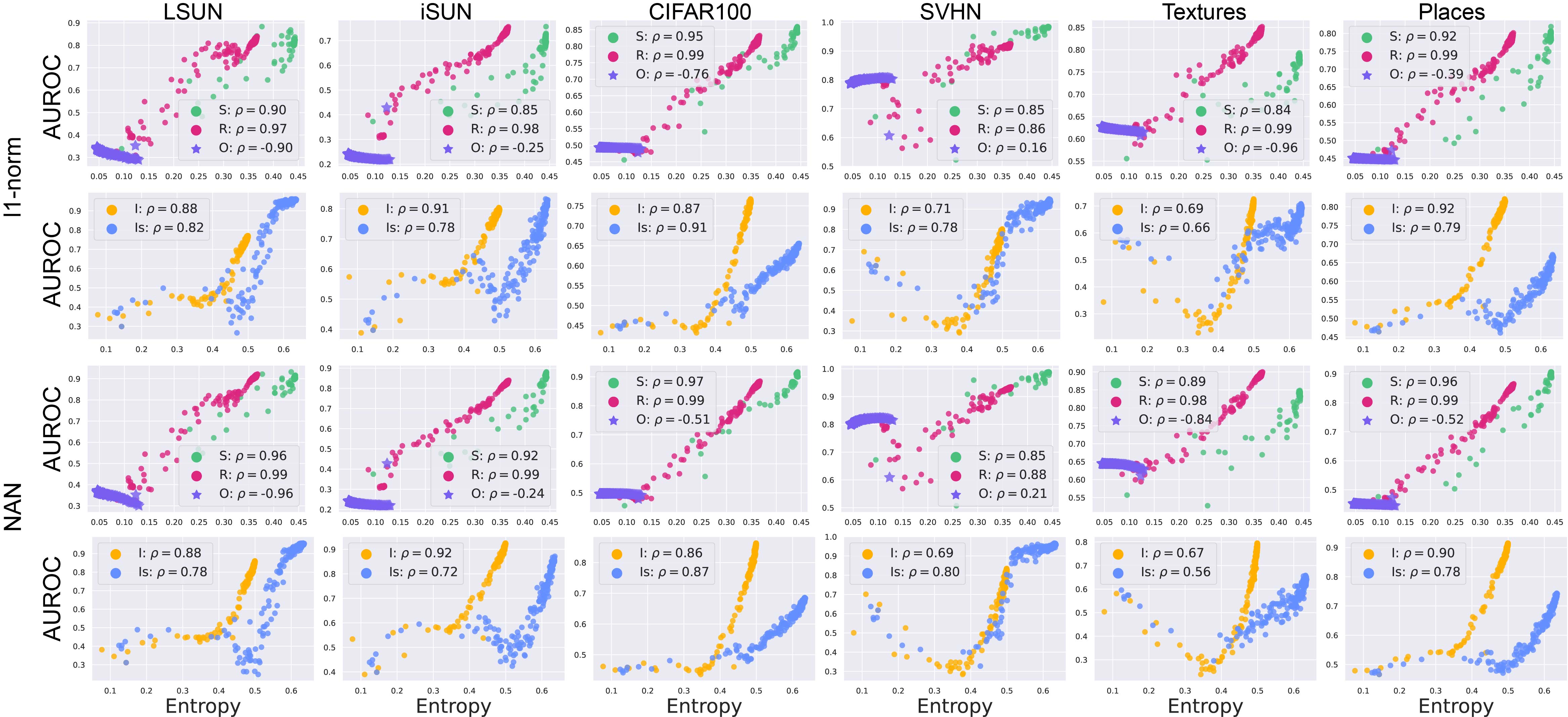}
\caption{
The graph of the detection performance versus the activation entropy. The performance is measured at every training epoch.
}
\label{fig:supp_detection_relation}
\end{figure*}

The full results on the relation between the activation entropy and the detection performance is given in Fig.~\ref{fig:supp_detection_relation}.

\paragraph{On the activation pattern}

\begin{figure*}[t]
\centering
\includegraphics[width=.995\linewidth]{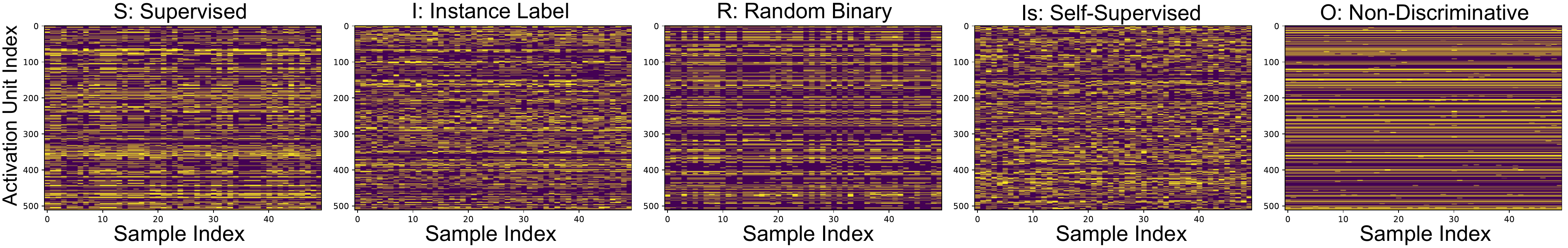}
\caption{
\textbf{Activation patterns} of randomly chosen 50 ID samples after training.
Each column corresponds to the activation pattern $\sign(\mathbf{a}^{(L)}) {\in} \{-1,1\}^{512}$ of an ID sample.
Discriminative training \{S,R,I,Is\} results in \textit{diverse} activation patterns, while the activation pattern \textit{collapses} for the non-discriminative model O.
}
\label{fig:supp_activation_patterns}
\end{figure*}

If the model is trained in a non-discriminative manner with a single class, then the entropy of activation is diminished. In this case, the activation pattern collapses as shown in Fig.~\ref{fig:supp_activation_patterns}.

\section{The Detailed Setup for the Experiments on NAN}
\label{asec:setup_exp_nan}

\subsection{Setup}

\paragraph{Setup: ImageNet-1k}

For the supervised model trained by the cross entropy, we utilize the ResNet-50 backbone trained on ImageNet-1k. The model is provided by the PyTorch model zoo.

For the supervised model trained by the contrastive loss (thanks to the authors of \cite{sun2022out}), we utilize the pretrained ResNet-50 model provided from the official GitHub page of KNN \cite{sun2022out}, which is trained on ImageNet-1k by the supervised contrastive loss \cite{khosla2020supervised} with the MLP projection head.

For the self-supervised contrastive model trained without the supervised labels of ID, thanks to the authors of MoCo-v2, we utilize the pretrained MoCo-v2 model provided from the official GitHub page of MoCo-v2 (the one with 71.1 accuracies on ImageNet-1k).

\paragraph{Setup: OOD CIFAR-10}
For the evaluation results of OOD detection `with supervised labels of ID'  in Table \ref{table:sota_small}, we train a cross-entropy model with supervised labels of CIFAR-10. The model has trained on CIFAR-10 over 800 epochs with the SGD optimizer and its momentum is 0.9. The learning rate decays to 0 from 0.03 by the cosine scheduler. The batch size is 512. The backbone is ResNet-18, accompanied by an MLP projection head on top of the encoder as in MoCo-v2. The embedding is normalized, and the cosine similarity logit is divided by the temperature 0.1.

For the evaluation results of OOD detection `without supervised labels of ID' in Table \ref{table:sota_small}, we train MoCo-v2 on CIFAR-10. The model is trained over 800 epochs with the SGD optimizer and its momentum 0.9. The batch size is 512. The learning rate is decayed by the cosine scheduler from 0.06 to 0. The model backbone is ResNet-18 combined with an MLP projection head.  For the other configurations, we follow those given in the link\footnote{\url{https://colab.research.google.com/github/facebookresearch/moco/blob/colab-notebook/colab/moco_cifar10_demo.ipynb}}. 
After training the MoCo-v2 model, the NAN score is computed over multiple (9 overall) translated images of the test sample including the original image, and the scores are aggregated by average \cite{tack2020csi}. This aggregation technique is used exclusively for the model trained by MoCo-v2.

\paragraph{Setup: OOD CIFAR-10}
The model training configuration for OCC is similar to that of label-free OOD detection on CIFAR-10 except that the train dataset is augmented randomly with 90-degree rotations. During the inference, the rotation is not used.

\subsection{Score Fusion}
A distance-based score $S_{dist}(\mathbf{x}) = d(X_{ind}, \mathbf{x})$ (\eg KNN, SSD, or Mahalanobis) can be combined with NAN in a simple manner by
\begin{equation}
S_{dist + NAN}(\mathbf{x}) = d(X_{ind}, \mathbf{x}) / \lVert \mathbf{a}^{(L)} \rVert_{\text{NAN}}.
\end{equation}

\section{Further Analysis on NAN}
\label{asec:more_ablation}

\paragraph{Setup}
We follow the same setup given in Sec.~\ref{sec:exp}. When CIFAR-10 is the ID data, the test OOD datasets are LSUN-fix, ImageNet-fix, CIFAR-100, SVHN, and Places. 
When ImageNet-1k is the ID data, the test OOD datasets are iNaturalist, SUN, Places, and Texture.

\subsection{Analysis on Projection Head}
\label{asec:more_ablation_proj_head}

\begin{table*}[t]
\centering
\resizebox{.9\linewidth}{!}{
\begin{tabular}{l ll cc}
\toprule
~ & ~ & ~ & AUROC$\uparrow$ & FPR95$\downarrow$ \\ 
ID & Architecture & Last hidden layer $\mathbf{a}^{(L)}$ & $l_1$-norm / NAN & $l_1$-norm / NAN \\ 
\midrule
\multirow{2}{*}{CIFAR-10} & ResNet-18 & average pool & 93.27 / 93.56 (\textbf{+0.29}) & 40.42 / 38.86 (\textbf{-1.56}) \\ 
~ & ResNet-18 + projection head & hidden layer in projection head & 92.43 / 94.94 (\textbf{+2.51}) & 43.02 / 30.08 (\textbf{-12.94}) \\ 
\multirow{2}{*}{ImageNet-1k} & ResNet-50 & average pool & 87.09 / 86.33 ({\color{red} \textbf{-0.76}}) & 44.67 / 46.56 ({\color{red} \textbf{+1.89}})\\ 
~ & ResNet-50 + projection head & hidden layer in projection head & 57.99 / 92.32 (\textbf{+34.33}) & 95.22 / 31.59 (\textbf{-63.63})\\ 
\bottomrule
\end{tabular}
}
\caption{
\textbf{Ablation of NAN with respect to the \textit{projection head}.} The sparsity term in NAN is particularly effective when applied to the network architecture that contains the MLP projection head. Note that the $l_1$-norm here refers to the NAN score without the sparsity term.
The reported performance here is obtained by averaging over all test OOD datasets.
}
\label{table:supp_ablation_proj}
\end{table*}

We analyze NAN with respect to \textbf{the projection head}. Table \ref{table:supp_ablation_proj} indicates that NAN is more effective when it is applied to the hidden layer of the projection head rather than the average pooling layer.

NAN (\ie particularly its sparsity term) becomes effective when the network learns to increase the number of deactivated units of ID samples (or have a relatively larger number of deactivated units for ID samples than OOD instances). Due to the entanglement of the feature map units in the average pooling layer, the network may not effectively increase the number of deactivated units in the average pooling layer. Hence, NAN can be sub-optimal for the average pooling layer.

\subsection{Analysis on Activation Function}

\begin{table*}[t]
\centering
\resizebox{.75\linewidth}{!}{
\begin{tabular}{l cc cc cc}
\toprule
~ & \multicolumn{2}{c}{ReLU}  & \multicolumn{2}{c}{Leaky ReLU}  & \multicolumn{2}{c}{GeLU}  \\ 
~ & AUROC$\uparrow$ & FPR95$\downarrow$ & AUROC$\uparrow$ & FPR95$\downarrow$ & AUROC$\uparrow$ & FPR95$\downarrow$ \\ 
\midrule
NAN w/o sparsity term ($l_1$-norm) & 92.43 & 43.02 & 92.40 & 44.65 & 92.68 & 43.84 \\ 
NAN & \textbf{94.94} & \textbf{30.08} & \textbf{94.92} & \textbf{30.56} & \textbf{94.05} & \textbf{35.02} \\ 
\bottomrule
\end{tabular}
}
\caption{
\textbf{Ablation of NAN with respect to the \textit{activation functions}} used in the last hidden layer. The ID data is CIFAR-10. The results indicate two aspects: (1) The performance of NAN is fairly robust with different choices of the activation function. (2) The sparsity term in NAN is always effective. 
The reported performance here is obtained by averaging over all test OOD datasets.
}
\label{table:supp_ablation_actv}
\end{table*}

We evaluate NAN with \textbf{different activation functions}. We follow the same experimental protocol given in Sec.~\ref{sec:exp_standard}. We apply different activation functions in the hidden layer of the projection head. The results given in Table \ref{table:supp_ablation_actv} shows that NAN is robust with respect to the choice of the activation function.

\subsection{Comparison with Embedding Magnitude}

\begin{table*}[t]
\centering
\resizebox{.7\linewidth}{!}{
\begin{tabular}{l l | c cc | c cc}
\toprule
ID & &  \multicolumn{3}{c|}{ImageNet-1k}  &  \multicolumn{3}{c}{CIFAR-10}  \\ 
~ & Formula & $d$ & AUROC$\uparrow$ & FPR95$\downarrow$ & $d$ & AUROC$\uparrow$ & FPR95$\downarrow$ \\ 
\midrule
embedding magnitude & $\lVert g(\mathbf{x}) \rVert_2$ & 128 & 84.09 & 72.85 & 128 & 93.00 & 43.40 \\ 
NAN w/o sparsity term & $\lVert \mathbf{a}^{(L)} \rVert_1$ & 2048 & 57.99 & 95.22 & 512 & 92.40 & 43.00 \\ 
NAN & $\lVert \mathbf{a}^{(L)} \rVert_{\text{NAN}}$ & 2048  & \textbf{92.32} & \textbf{31.59} & 512 & \textbf{94.90} & \textbf{30.10} \\ 
\bottomrule
\end{tabular}
}
\caption{
Comparison of NAN with the embedding magnitude. The embedding magnitude has been widely used in previous works. Here $d$ indicates the dimension of the corresponding layer. The dimension of the embedding layer is often chosen small for effective training of the model. Due to its small layer dimension, the embedding magnitude may not fully capture the activation patterns, and hence can be sub-optimal.
The reported performance here is obtained by averaging over all test OOD datasets. 
}
\label{table:supp_ablation_embedding}
\end{table*}

For the sake of extensiveness, we compare NAN with the \textbf{embedding magnitude}. The embedding magnitude has been widely used in prior works for OOD detection-related tasks. The dimension of the embedding layer is often chosen to be a small number to avoid the curse of dimensionality during training. This may have a trade-off to OOD detection as the embedding of a small dimension may not capture diverse activation patterns of embedding layer units and therefore its norm may not effectively differentiate OOD from ID. This hypothesis seems consistent to the results given in Table \ref{table:supp_ablation_embedding}.

\subsection{Evaluation of NAN on ViT}

\begin{table*}[!t]
\centering
\resizebox{.995\linewidth}{!}{
\begin{tabular}{l cc cc cc cc cc c}
\toprule
OOD & \multicolumn{2}{c}{iNaturalist} & \multicolumn{2}{c}{SUN}  & \multicolumn{2}{c}{Places} & \multicolumn{2}{c}{Texture} & \multicolumn{2}{c}{Average} &  \multirow{2}{*}{ID ACC}  \\ 
~ & AUROC$\uparrow$ & FPR95$\downarrow$ & AUROC$\uparrow$ & FPR95$\downarrow$ & AUROC$\uparrow$ & FPR95$\downarrow$ & AUROC$\uparrow$ & FPR95$\downarrow$ & AUROC$\uparrow$ & FPR95$\downarrow$ & ~ \\ 
\midrule
MSP & 89.63 & 50.57 & 80.64 & 75.54 & 79.78 & 76.24 & 82.98 & 65.14 & 83.26 & 66.87 & 81.07 \\ 
Energy & 83.76 & 49.68 & 56.50 & 75.22 & 54.77 & 78.38 & 72.44 & 65.09 & 66.87 & 67.09 & 81.07 \\ 
Mahalanobis & 91.96 & 43.76 & 75.62 & 86.01 & 61.50 & 89.74 & 84.60 & 67.93 & 78.42 & 71.86 & 81.07 \\ 
KNN & 91.43 & 50.04 & 83.45 & 75.76 & 79.46 & 78.41 & 89.25 & 50.78 & \textbf{85.90} & 63.75 & 81.07 \\ 
embedding magnitude & 81.26 & 66.16 & 78.64 & 67.44 & 75.81 & 69.37 & 82.93 & 57.11 & 79.66 & 65.02 & 81.07 \\ 
NAN w/o sparsity term (\ie $l_1$-norm) & 54.93 & 83.98 & 67.05 & 80.47 & 65.25 & 81.01 & 67.87 & 72.54 & 63.78 & 79.50 & 81.07 \\ 
\rowcolor{lightgray}
NAN & 92.46 & 45.82 & 82.11 & 67.62 & 80.46 & 69.66 & 87.24 & 57.77 & 85.57 & \textbf{60.22} & 81.07 \\ 
\bottomrule
\end{tabular}
}
\caption{
\textbf{Results on ImageNet-1k (ID) with \textit{ViT-B/16}.}
}
\label{table:supp_sota_large_vit}
\end{table*}

We evaluate NAN on the \textbf{vision transformer ViT}. We utilize ViT-B/16 pretrained on ImageNet-1k, which can be downloaded from PyTorch\footnote{\url{https://pytorch.org/vision/main/models/generated/torchvision.models.vit_b_16.html}}. Analogous to the observations in Sec.~\ref{asec:more_ablation_proj_head}, direct usage of NAN on the pretrained ViT can be sub-optimal because the class token output of ViT is the LayerNorm layer, which can cancel out the norm information therein. Therefore, we add an MLP projection head on top of the pretrained ViT, and fine-tune the projection head while freezing the pretrained ViT backbone. The MLP projection head consists of a single hidden layer whose dimension is 786 and its activation function is ReLU.  The embedding of the projection head is normalized and divided by the temperature $0.2$, and trained by the cross entropy with 10 epochs under SGD, using the learning rate 0.03 that decays to 0 by the cosine scheduler.

For comparison, the KNN and Mahalanobis scores are applied on the original class token output of the pretrained ViT, and hence are independent of the projection head fine-tuning. Other OOD detection scores (MSP, Energy, and embedding magnitude) are applied to the fine-tuned classifier of the projection head. NAN utilizes the hidden layer in the projection head as this layer is the last hidden layer that involves the activation function computation.

Table \ref{table:supp_sota_large_vit} shows that NAN is effective for the ViT network as well. In addition, NAN is comparable to the state-of-the-art OOD detection scores.

\paragraph{Note on the ViT performance of KNN}
Note that the performance of KNN in Table \ref{table:supp_sota_large_vit} is lower than that of KNN reported in \cite{sun2022out}. This is because the KNN we implemented is applied on ViT pretrained on ImageNet-1k, while the KNN reported in \cite{sun2022out} is applied on ViT pretrained on ImageNet-21k.

\subsection{Evaluation of Hidden Classifier for OOD Detection}

\begin{table*}[!t]
\centering
\resizebox{.995\linewidth}{!}{
\begin{tabular}{l  l cc cc cc cc cc  cc}
\toprule
test OOD datasets & ~ & \multicolumn{2}{c}{LSUN-fix} & \multicolumn{2}{c}{ImageNet-fix}  & \multicolumn{2}{c}{CIFAR-100}  & \multicolumn{2}{c}{SVHN}  & \multicolumn{2}{c}{Places}  & \multicolumn{2}{c}{Average}  \\ 
Score & Formula & AUROC$\uparrow$ & FPR95$\downarrow$ & AUROC$\uparrow$ & FPR95$\downarrow$ & AUROC$\uparrow$ & FPR95$\downarrow$ & AUROC$\uparrow$ & FPR95$\downarrow$ & AUROC$\uparrow$ & FPR95$\downarrow$ & AUROC$\uparrow$ & FPR95$\downarrow$ \\ 
\midrule
\makecell[l]{hidden classifier \\ confidence} & $\max_k \overline{\psi}_k^{(L)}(\mathbf{x})$ & 95.06 & 33.35 & 94.54 & 35.92 & 92.17 & 45.10 & 94.66 & 39.91 & 94.66 & 30.15 & 94.22 & 36.89 \\ 
\bottomrule
\end{tabular}
}
\caption{
Results on CIFAR-10 (ID) with ResNet-18. \textbf{The hidden classifier confidence} is evaluated as a score function for OOD detection. The results shows that the hidden classifier confidence is capable of OOD detection.
}
\label{table:supp_ablation_hidden}
\end{table*}

We evaluate the \textbf{hidden classifier for OOD detection}.
NAN's numerator is the $l_1$-norm of the activation vector, which we proved is a confidence value of the hidden classifier. We test this numerator component by testing the OOD detection capability of this hidden classifier confidence.
Table \ref{table:supp_ablation_hidden} shows the hidden classifier confidence is capable of OOD detection.

\subsection{Evaluation of NAN on CIDER}

\begin{table}[!t]
\centering
\resizebox{.95\linewidth}{!}{
\begin{tabular}{l cc cc cc cc cc cc}
\toprule
~ & \multicolumn{2}{c}{SVHN} & \multicolumn{2}{c}{Places365} & \multicolumn{2}{c}{iSUN} & \multicolumn{2}{c}{Texture} & \multicolumn{2}{c}{LSUN} & \multicolumn{2}{c}{Average}\\ 
~ & FPR95 & AUROC & FPR95 & AUROC & FPR95 & AUROC & FPR95 & AUROC & FPR95 & AUROC & FPR95 & AUROC \\ 
\hline
NAN & 73.82 & 90.46 & 26.33 & 94.65 & 25.47 & 96.46 & 25.35 & 95.21 & 1.17 & 99.45 & 30.43 & 95.25 \\ 
KNN & 4.44 & 99.36 & 37.88 & 92.97 & 22.94 & 96.16 & 17.27 & 97.15 & 9.85 & 98.21 & 18.48 & 96.77 \\ 
NAN+KNN & 5.70 & 98.62 & 21.79 & 95.32 & 14.01 & 97.64 & 16.21 & 96.61 & 0.95 & 99.68 & \textbf{11.73} & \textbf{97.57} \\ 
\bottomrule
\end{tabular}
}
\caption{
The results of the OOD detection scores (KNN, NAN, NAN+KNN) on the model trained by CIDER on CIFAR-10 (ID).
}
\label{table:cider}
\end{table}

CIDER \cite{ming2022exploit} is a training framework that is particularly effective for the KNN score.
We evaluate NAN's compatibility to the KNN score from the model trained by CIDER. The results shown in Table \ref{table:cider} indicates that NAN can effectively enhance the KNN score of CIDER.

\subsection{Comparison of NAN to various forms of vector norms}

\begin{table}[!t]
\centering
\resizebox{.995\linewidth}{!}{
\begin{tabular}{lllllllllllllllll}
\toprule
~ & \multicolumn{2}{c}{iNaturalist} & \multicolumn{2}{c}{SUN} & \multicolumn{2}{c}{Places} & \multicolumn{2}{c}{Texture} & \multicolumn{2}{c}{ImageNet-O} & \multicolumn{2}{c}{OpenImage-O} & \multicolumn{2}{c}{Species} & \multicolumn{2}{c}{Average} \\ 
~ & FPR95 & AUROC & FPR95 & AUROC & FPR95 & AUROC & FPR95 & AUROC & FPR95 & AUROC & FPR95 & AUROC & FPR95 & AUROC & FPR95 & AUROC \\ 
\hline
$l_1$-norm & 97.52 & 52.06 & 95.58 & 59.40 & 95.65 & 61.30 & 92.11 & 59.21 & 88.20 & 67.97 & 92.43 & 63.10 & 95.83 & 59.42 & 93.90 & 60.35 \\ 
$1/l_0$-norm & 15.66 & 96.58 & 33.38 & 91.83 & 39.10 & 90.37 & 44.36 & 87.41 & 88.60 & 56.76 & 41.29 & 88.58 & 64.04 & 79.55 & 46.63 & 84.44 \\ 
Residual & 28.74 & 95.09 & 46.88 & 89.76 & 58.91 & 85.77 & 11.28 & 96.45 & 63.50 & 84.24 & 34.96 & 93.34 & 74.43 & 73.72 & 45.53 & \textbf{88.34} \\ 
NAN & 15.86 & 96.94 & 29.81 & 92.77 & 37.21 & 91.46 & 43.46 & 88.09 & \textbf{87.95} & 69.74 & 38.12 & 92.44 & 64.56 & 80.09 & \textbf{45.28} & 87.36 \\ 
\hline 
\multicolumn{3}{l}{\textit{with ReAct:}} \\
$l_1$-norm & 98.07 & 37.19 & 96.37 & 46.97 & 96.90 & 45.47 & 85.44 & 61.21 & 84.95 & 74.80 & 93.54 & 54.48 & 98.81 & 41.25 & 93.44 & 51.62 \\ 
$1/l_0$-norm & 21.19 & 95.60 & 36.56 & 90.81 & 41.28 & 89.63 & 52.16 & 82.23 & 90.35 & 53.37 & 49.25 & 85.85 & 61.45 & 81.89 & 50.32 & 82.77 \\ 
Residual & 28.59 & 95.06 & 39.40 & 91.95 & 51.02 & 88.18 & 12.11 & 96.87 & 68.30 & 83.01 & 36.67 & 92.62 & 72.27 & 75.03 & 44.05 & \textbf{88.96} \\ 
NAN & 13.86 & 97.37 & 24.90 & 94.69 & 33.31 & 92.52 & 34.02 & 91.44 & 84.10 & 71.72 & 37.27 & 92.02 & 63.68 & 81.10 & \textbf{41.59} & 88.69 \\ 
\bottomrule
\end{tabular}
}
\caption{
The comparison of NAN with various forms of vector norms on ImageNet-1k (ID).
}
\label{table:various_norm}
\end{table}

To further highlight the effectiveness of NAN, we compare NAN with various forms of vectors norms; namely, $l_1$-norm, the reciprocal of $l_0$-norm, and the residual of ViM which is the $l_2$-norm of the orthogonal projection. 
The experiment protocol follows \cite{yang2022openood}, and the OOD datasets can be downloaded from its GitHub repository.

The results in Table \ref{table:various_norm} indicate that NAN is significantly better than the $l_1$-norm and the reciprocal of $l_0$-norm. We note that $l_1$-norm does not capture deactivation, while the reciprocal of $l_0$-norm captures only deactivation. Hence, the superiority of NAN over these vector norms indicate that capturing both activation and deactivation is crucial.

Compared to the residual of ViM, on the other hand, NAN is notably superior with respect to the FPR95 metric when ReAct is applied on the model, while NAN is comparable to the residual when without ReAct. We note, however, that the residual of ViM requires eigen decomposition of the bankset features, while the computation of NAN is done by a single forward pass of the network.

\section{Limitation of NAN}
\label{asec:limitation}

Based on our theoretical observations, NAN is intrinsically a classifier output and hence may inherit the weaknesses of classifier-based OOD detectors that have been recently found in \cite{dietterich2022familiarity,fang2022out}. In addition, as observed in Sec.~\ref{asec:more_ablation_proj_head}, the optimal usage of NAN requires networks that involve the MLP projection head.

\clearpage

\end{document}